\documentclass[sigconf, nonacm]{acmart}
\usepackage{titlesec}

\usepackage{array}
\usepackage{subcaption}
\usepackage{tikz}
\usepackage{pgf-pie}
\usepackage{pgfplots, pgfplotstable}
\usepackage{lipsum}
\usepackage{censor}
\usepackage[export]{adjustbox}
\usepackage{afterpage}
\usepackage{ifthen}
\usepackage{xstring}
\usepackage{multirow}

\newboolean{fullsource}
\setboolean{fullsource}{true}

\newcommand{\opensource}[1]{\ifbool{fullsource}{}{#1}}
\newcommand{\fullsource}[1]{\ifbool{fullsource}{#1}{}}

\newcommand{\eq}{\mkern1.5mu{=}\mkern1.5mu}
\newcommand{\ignore}[1]{{}}

\newcommand{\mymkr}[1]{
    \begin{tikzpicture}
    \draw (0,0) node[rectangle,
          rounded corners,
          very thin,
          inner sep=1.75pt,
          draw=black!20!red,
          ](A){#1};
    \end{tikzpicture}
}

\newcommand{\mymkg}[1]{
    \begin{tikzpicture}
    \draw (0,0) node[rectangle,
          rounded corners,
          very thin,
          inner sep=2pt,
          draw=black!50!green,
          ](A){#1};
    \end{tikzpicture} 
}

\newcommand{\mymkb}[1]{
    \begin{tikzpicture}
    \draw (0,0) node[rectangle,
          rounded corners,
          very thin,
          inner sep=2pt,
          draw=white!30!blue,
          ](A){#1};
    \end{tikzpicture} 
}

\xdefinecolor{coolyellow}{RGB}{255, 246, 100}
\definecolor{darkpastelgreen}{rgb}{0.01, 0.75, 0.24}

\usetikzlibrary{arrows}
\usetikzlibrary{shapes}
\pgfplotsset{compat=1.9}

\newcommand{\etal}{et al.}
\newcommand{\teaser}[1]{\begin{teaserfigure}#1\end{teaserfigure}}
\titleformat{\paragraph}[runin]{\bfseries}{\thesection}{0em}{}[\\]
\titlespacing{\paragraph}{0em}{1em}{0em}
\AtBeginDocument{%
  \providecommand\BibTeX{{%
    \normalfont B\kern-0.5em{\scshape i\kern-0.25em b}\kern-0.8em\TeX}}}

\begin{document}
    \title[Re:Draw, arXiv Version]{Re:Draw - Context Aware Translation as a Controllable Method for Artistic Production}
    
    \author{\href{http://www.jaliborc.com/}{Jo\~{a}o Lib\'{o}rio Cardoso}}
    \affiliation{
      \institution{TU Wien}
      \country{Austria}
    }
    
    \author{\href{http://www.banterle.com/francesco/}{Francesco Banterle}}
    \affiliation{
      \institution{CNR-ISTI}
      \country{Italy}
    }
    
    \author{\href{https://vcg.isti.cnr.it/~cignoni/}{Paolo Cignoni}}
    \affiliation{
      \institution{CNR-ISTI}
      \country{Italy}
    }
    
    \author{\href{https://www.cg.tuwien.ac.at/staff/MichaelWimmer}{Michael Wimmer}}
    \affiliation{
      \institution{TU Wien}
      \country{Austria}
    }

    \begin{abstract}
        \noindent We introduce context-aware translation, a novel method that combines the benefits of inpainting and image-to-image translation, respecting simultaneously the original input and contextual relevance -- where existing methods fall short. By doing so, our method opens new avenues for the controllable use of AI within artistic creation, from animation to digital art.

As an use case, we apply our method to redraw any hand-drawn animated character eyes based on any design specifications -- eyes serve as a focal point that captures viewer attention and conveys a range of emotions, however, the labor-intensive nature of traditional animation often leads to compromises in the complexity and consistency of eye design. Furthermore, we remove the need for production data for training and introduce a new character recognition method that surpasses existing work by not requiring fine-tuning to specific productions. This proposed use case could help maintain consistency throughout production and unlock bolder and more detailed design choices without the production cost drawbacks. A user study shows context-aware translation is preferred over existing work 95.16\% of the time.
    \end{abstract}
    \teaser{
    \centering
    \setlength{\tabcolsep}{1pt}
    \footnotesize{
    \begin{tabular}{cccccc}
        \multirow{2}{*}[20ex]{\raisebox{1.3cm}{\rotatebox{90}{Input}} \includegraphics[width=0.157\textwidth]{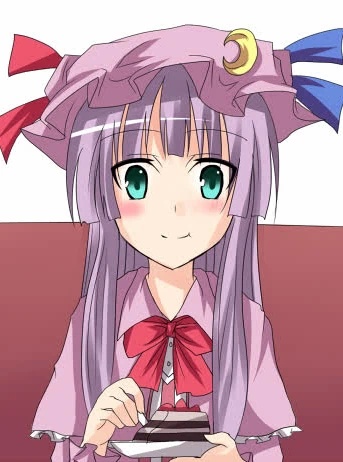}}
        & \raisebox{.7cm}{\rotatebox{90}{Ours}} \includegraphics[trim=0 175 0 0, clip, width=0.157\textwidth]{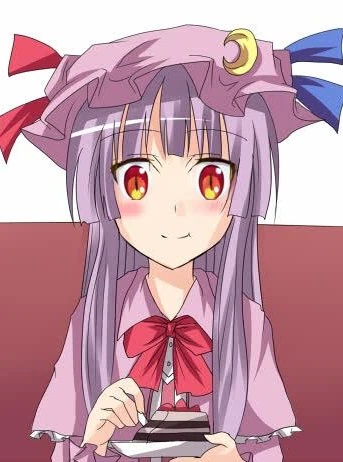}
        & \includegraphics[trim=0 175 0 0, clip, width=0.157\textwidth]{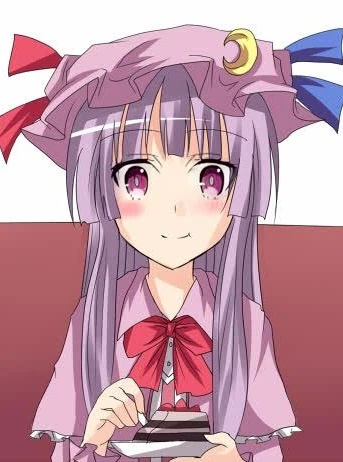}
        & \includegraphics[trim=0 175 0 0, clip, width=0.157\textwidth]{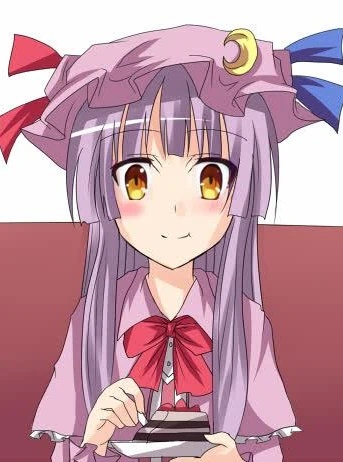}
        & \includegraphics[trim=0 175 0 0, clip, width=0.157\textwidth]{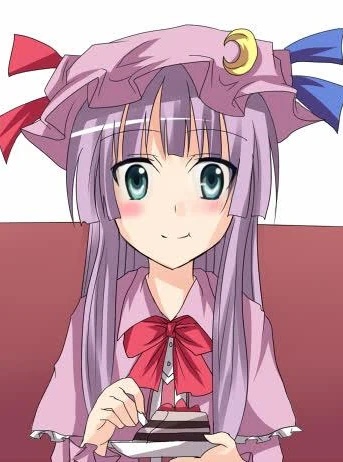}
        & \includegraphics[trim=0 175 0 0, clip, width=0.157\textwidth]{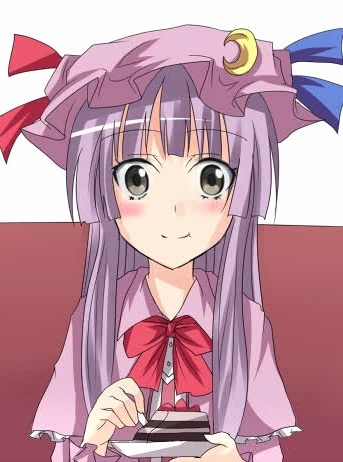}\\
        {}
        & \raisebox{0.15cm}{\rotatebox{90}{\tiny{Existing Work}}} \includegraphics[width=0.077\textwidth]{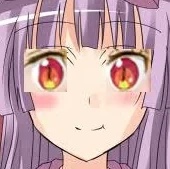} \includegraphics[width=0.077\textwidth]{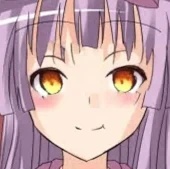}
        & \includegraphics[width=0.077\textwidth]{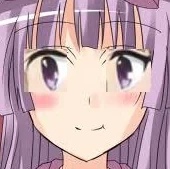} \includegraphics[width=0.077\textwidth]{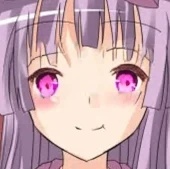}
        & \includegraphics[width=0.077\textwidth]{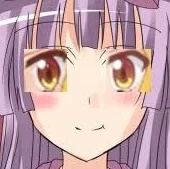} \includegraphics[width=0.077\textwidth]{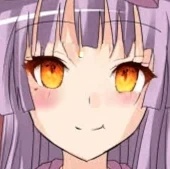}
        & \includegraphics[width=0.077\textwidth]{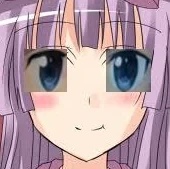} \includegraphics[width=0.077\textwidth]{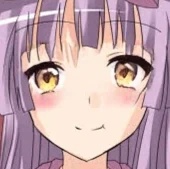}
        & \includegraphics[width=0.077\textwidth]{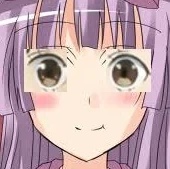} \includegraphics[width=0.077\textwidth]{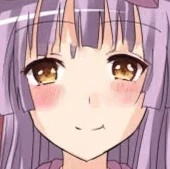}
    \end{tabular}
    }
    \vspace{-.8em}
    \caption{\textbf{Teaser.} As shown, the proposed context-aware translation is capable of automatically redrawing parts of images according to any provided design, without the need for fine-tuning. Unlike image-to-image translation \cite{liu2019few}, which neglects surrounding context, our approach considers the entire frame. Unlike inpainting \cite{sdlx}, which lacks artistic control by ignoring the original content, our method honors the artist's input. This facilitates the production of more consistent artwork and allows for more complex design choices.}
    \label{fig:teaser}
    \vspace{1em}
}
    
    \renewcommand{\shortauthors}{Lib\'{o}rio Cardoso \etal}
    \maketitle
    
	\section{Introduction} \label{sec:intro}
\noindent Alfred Yarbus’s influential work \cite{yarbus1967} quantified a long-standing intuition: using an eye tracker, he noted that observers spend a surprisingly large fraction of time fixated on the eyes in a picture. The eyes of others are important to humans because they convey subtle information about a person’s mental state (e.g., attention, intention, emotion) and physical state (e.g., age, health, fatigue). This significance has translated into the realm of hand-drawn animation, where the eye designs have often become increasingly complex and expressive to capture these nuances. However, this complexity comes at a cost. Despite its massive resurgence in the last decade \cite{Hiromichi2019}, traditional animation has struggled to benefit from advances in computer graphics: techniques used in production remain largely the same, with productions relying on repetitive manual labor from a large workforce. As a result, the eyes, being the most time-consuming and intricate to draw, are often the first elements to be simplified, leading to compromises in both expression and artistic consistency. Our aim is to introduce a computational method that can alleviate some of these challenges without sacrificing the artistic integrity of the medium. 

\begin{figure}
    \centering
    \begin{tikzpicture}
        \pie[radius=2]{
            41/1st and 2nd Keys,
            24.8/Clean-Up,
            19/Inbetweening,
            7.4/Coloring,
            7.8/Compositing}
    \end{tikzpicture}
    \caption[Industry Survey]{\textbf{Industry Survey}. Average reported time spent in the different stages of production of a cut, in percentage.} \label{survey-stages}
\end{figure}

\begin{figure}
    \pgfplotstableread{ 
        Label       1st         2nd         3rd         4th     5th
        Objects     8           8           8           10      66
        Clothes     7           0           23          61      9
        Hair        14          21          35          21      9
        Body        28          28          28          0       16
        Face        50          35          8           7       0
    }\testdata
    
    \begin{tikzpicture}
      \tikzstyle{every node}=[font=\small]
        \begin{axis}[
            xbar stacked, xmin=0, xmax=110,
            y=8mm, width=\linewidth,
            ytick=data, yticklabels from table={\testdata}{Label}]
        
        \addplot [fill=blue!60] table [x=1st, meta=Label, y expr=\coordindex] {\testdata};
        \addplot [fill=cyan!60] table [x=2nd, meta=Label, y expr=\coordindex] {\testdata};
        \addplot [fill=coolyellow!60] table [x=3rd, meta=Label, y expr=\coordindex] {\testdata};
        \addplot [fill=orange!60] table [x=4th, meta=Label, y expr=\coordindex] {\testdata};
        \addplot [fill=red!60] table [x=5th, meta=Label, y expr=\coordindex] {\testdata};
        \legend{1\textsuperscript{st},2\textsuperscript{nd},3\textsuperscript{rd},4\textsuperscript{th},5\textsuperscript{th}}
        
        \end{axis}
    \end{tikzpicture}
    
    \caption[Industry Survey]{\textbf{Industry Survey}. Reported ranking of common elements in animation from most to least time consuming, in percentage.}
    \label{survey-elements}
\end{figure}

We further conducted a survey among 17 professional animators, of which 29.4\% (5/17) work at established studios, and the rest either freelanced or worked for smaller studios. We asked them multiple questions about time consumption, (answering optional), of which the full breakdown is shown in Figs.~\ref{survey-stages} ,\ref{survey-elements}.  Character faces were reported to be the most complex part of animation, with 50\% (8/16) reporting it as the element they spend the most time on. Of the remaining animators, 75\% (6/8) voted for either anatomy or hair. Drawing was estimated to constitute the vast majority of the work (84.8\%), and doing it at the highest level of detail was estimated to take 1.7 times the amount of work than on average, for a total of 66m of additional human effort per key frame from 1st key through coloring. Sadly, the fact that 52.9\% (9/17) still use paper drawings in their studios, despite 100\% preferring to draw digitally, indicates that using computational tools during the early drawing stages might not be possible yet in practical terms.

\subsection{Problem}
\noindent Existing deep learning methods present significant limitations within artistic applications. Inpainting, while capable of generating detailed art that fits within existing content, offers little control over the generated content, making it unsuitable for most precise artistic endeavors \cite{Kenta+2020}. Image-to-image translation, while being able to take artistic input, is constrained by only being applicable to entire images, as it does not take into account the context surrounding target regions.

We propose context-aware translation as the solution to these limitations. We then apply it in a novel pipeline that  automates increasing consistency and amount of detail in the eyes of hand-drawn animation characters. It effectively mimics the work of cleanup animators, who redraw frames to fix mistakes and better match the character color guides -- despite the misleading name, color guides, also known as model sheets, depict all the information an artist would need to draw a character while remaining true to its intended design and the art style of the production\fullsource{(see Fig.~\ref{color-guide} for an example)}. We also tackle an additional problem this use-case raises: the lack of training datasets of anime production, which we address by proposing methods to negate the need for production data entirely, including a novel character recognition method.

\fullsource{
    \begin{figure}
        \centering
        \includegraphics[width=\linewidth]{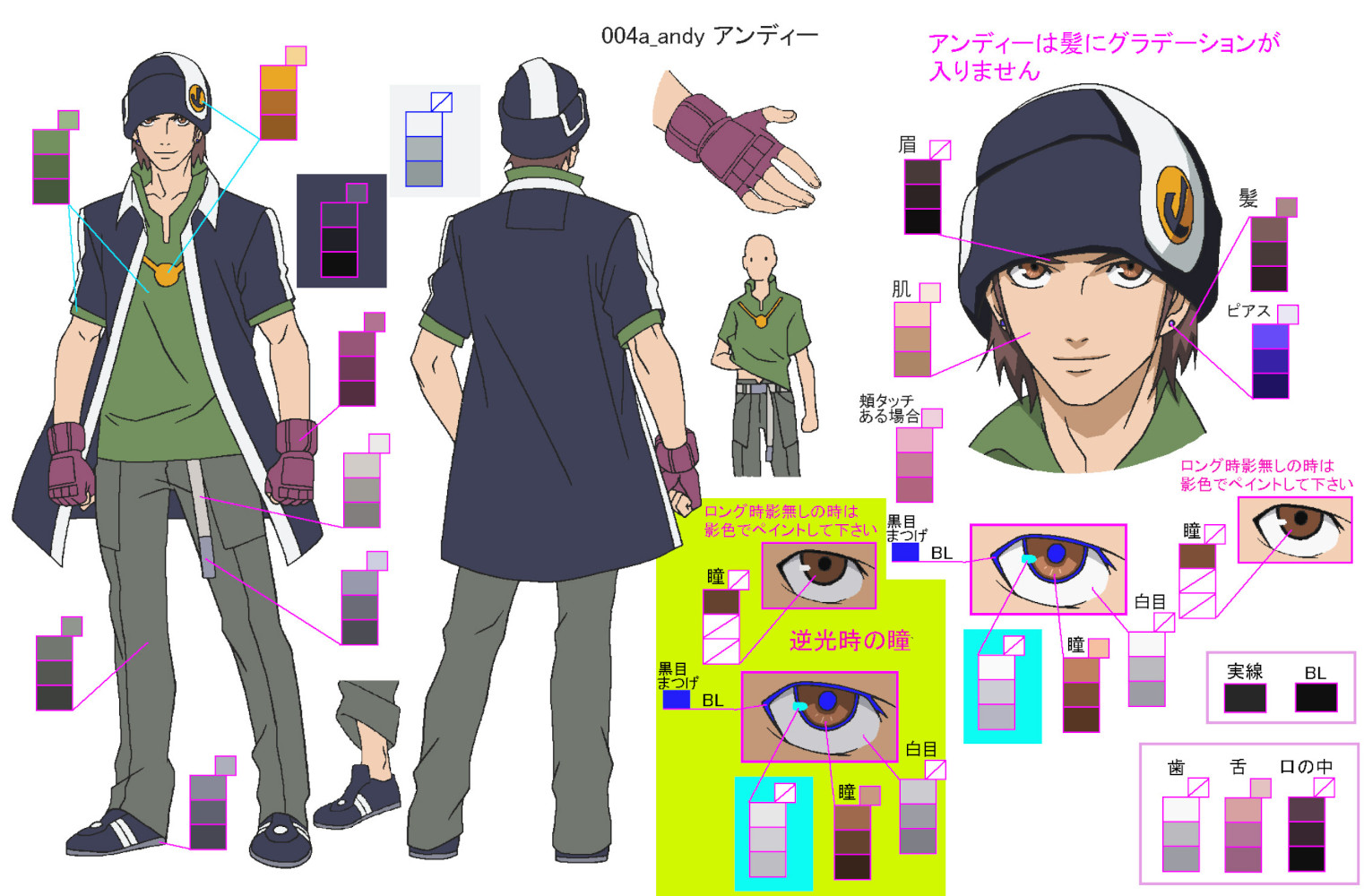}
        \caption[Color Guide]{\textbf{Color Guide}. Also known as a model sheet, this type of document depicts all the information an artist is expected to need to know how to draw a character in accordance with the production direction and style. Example from \textit{Aquarion Evol} \cite{aquarion}}.
        \label{color-guide}
    \end{figure}
}

In summary, these are our key contributions:
\begin{enumerate}
    \item Context-aware translation, a novel general deep-learning method that avoids the limitations of both inpainting and image-to-image translation.
    \begin{enumerate}
        \item A dual discriminator structure and novel adversarial losses that enforce simultaneous respect for input content, translation requirements, and context constraints.
        \item A triple-reconstruction loss that yields greater generation capabilities than traditional loss.
    \end{enumerate}
    \item A character design recognition network that outperforms existing work by using a production-style-aware latent space.
    \item A novel pipeline that takes advantage of the aforementioned contributions to automatically increase the consistency and amount of detail in the eye region of characters, and without the need of production data during training.
\end{enumerate}
Furthermore, we present an ablation study in Section \ref{sec:ablation} that scrutinizes the benefits of each of our novel components, contribution of each loss used, compares both our context-aware translation and style-aware clustering against existing work, and assesses the robustness and temporal coherence of our method. We also present a user study with 63 participants in Section \ref{user-study} that tests three key properties: the absence of detectable artifacts, the enhancement of artwork detail, and the overall aesthetic preference when compared to existing methods, all of which our pipeline successfully validated. 
	\fullsource{\begin{figure*}
    \centering
    \begin{subfigure}{.121\linewidth}
         \centering
         \includegraphics[width=\textwidth]{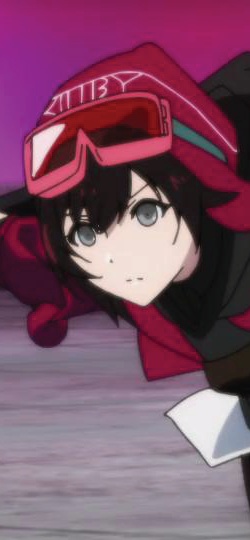}
         \caption{Input}
    \end{subfigure}
    \begin{subfigure}{.121\linewidth}
         \centering
         \includegraphics[width=\textwidth]{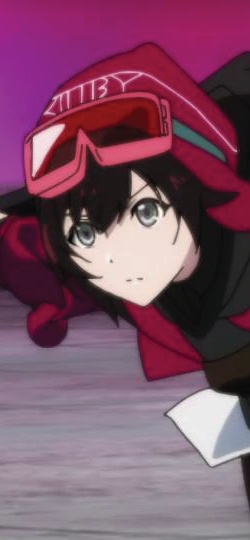}
         \caption{Ours}
    \end{subfigure}
    \begin{subfigure}{.121\linewidth}
         \centering
         \includegraphics[width=\textwidth]{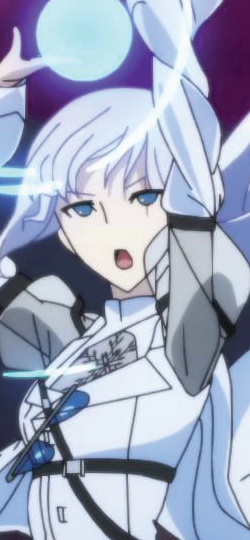}
         \caption{Input}
    \end{subfigure}
    \begin{subfigure}{.121\linewidth}
         \centering
         \includegraphics[width=\textwidth]{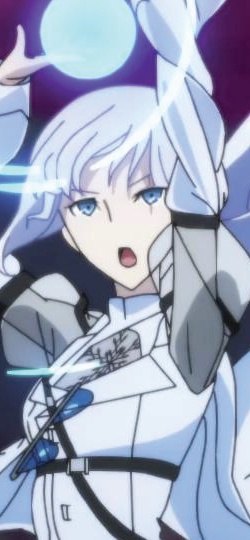}
         \caption{Ours}
    \end{subfigure}
    \begin{subfigure}{.121\linewidth}
         \centering
         \includegraphics[width=\textwidth]{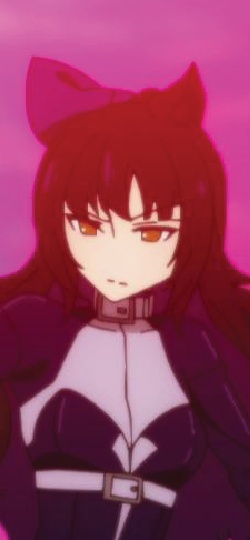}
         \caption{Input}
    \end{subfigure}
    \begin{subfigure}{.121\linewidth}
         \centering
         \includegraphics[width=\textwidth]{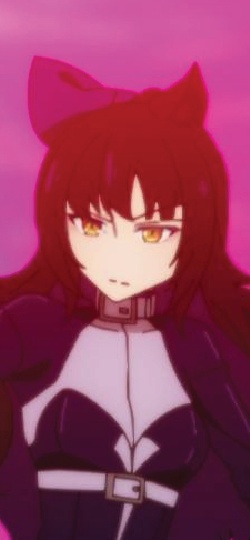}
         \caption{Ours}
    \end{subfigure}
    \begin{subfigure}{.121\linewidth}
         \centering
         \includegraphics[width=\textwidth]{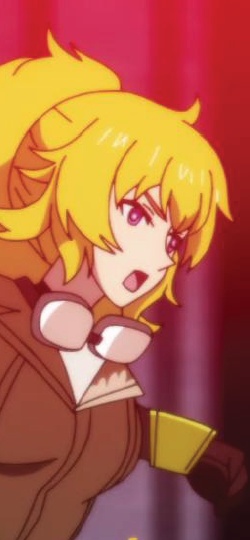}
         \caption{Input}
    \end{subfigure}
    \begin{subfigure}{.121\linewidth}
         \centering
         \includegraphics[width=\textwidth]{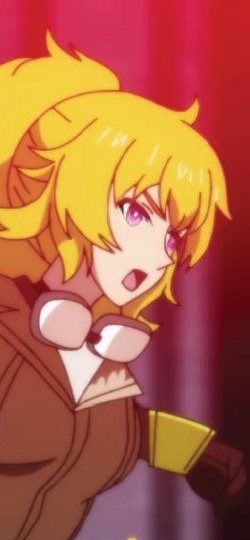}
         \caption{Ours}
    \end{subfigure}
    \vspace{-.8em}
    \caption{\textbf{Proposed Use-Case.} Context-aware translation can be used to automatically redraw the eyes of any character according to a provided color-guide, without fine-tuning. These characters (from \textit{RWBY: Ice Queendom} \cite{rwby}) were not part of the training set.}
    \label{fig:rwby}
\end{figure*}}

\section{Related Work} \label{sec:related}
\noindent In this Section we describe the minimum animation production background necessary to frame our use-case, and analyze relevant deep-learning existing work.

\paragraph{Anime Production Background}
The production of limited animation is complex and requires the combined efforts of many professional technicians and artists. This results in the need for a pipeline with precisely defined steps, which can be roughly divided into drawing, finishing, and compositing \cite{furansujin}. In the first step, frame sequences are planned out and frames are drawn, going through multiple revisions and artists. As per our survey in Section \ref{sec:intro}, paper is still prevalent. Finishing involves cleaning and coloring. Each frame is redrawn and in-between frames are added, both in a strict digital format that contains info for the coloring team. Starting with coloring, all operations are digital. Finally, compositing assembles the drawn animations with other elements, such as background or digital effects.

Thus, for a computational method to be usable in this setting, it must: 1) fit within the existing pipeline, preferably after cleaning; 2) not require manual input for each frame in a sequence, defeating its purpose; 3) have a high level of artistic control. 

\paragraph{Editing Content with Style}
The seminal work by Gatys \etal \cite{Gatys+2016} showed how to edit an image by varying its overall style by transferring the style of a target image using deep learning optimizations. This work has generated a prolific field \cite{Huang+2017, Karras+2019, Karras+2020, Karras+2021b, Karras+2021}. 

Compared to early image-to-image translation methods\cite{Zhu+2017, Isola+2017}, Liu \etal \cite{liu2019few} introduced FUNIT a few-shot unsupervised image-to-image translation framework based on MUNIT \cite{Xun+2018}. 
This approach has the ability to translate from an unseen domain with unpaired data. FUNIT achieves this by coupling an adversarial training scheme with a novel multi-task adversarial discriminator. This work has already stemmed variants and improvements \cite{Kim+2020, Saito+2020, Nizan+2020, Li+2021}, but maintain the main drawback of previously mentioned works to modify globally the entire image without guaranteeing that characters in the image will maintain the same pose or expression and other elements won't be modified. In our work, we want to edit/modify the drawing of a character, but only the design should be varied locally to certain parts; e.g., the eyes. A crucial aspect is to maintain the original pose, facial expressions (e.g., eyes' looking direction), and non-targeted local elements of the original image.

\paragraph{Artistic Methods}
Deep learning techniques have started to be applied to comics \cite{Augereau+2018} and illustration editing; most literature focused on the colorization of either sketches or shaded manga drawings and main lines extraction.
Regarding this latter topic, Simo-Serra \etal \shortcite{Serra+2016} proposed one of the first deep learning methods based on a simple encoder-decoder CNN architecture for sketches. Li \etal \shortcite{Li+2017lines} generalized it for patterns in drawings using a U-Net with skip connections. Lines extraction methods can be improved when they are paired with user inputs \cite{Serra+2018b} or adversarial training \cite{Serra+2018} or unpaired data with the synthesis of paired ones \cite{Lee+2019}.

Concerning colorization, Yuanzheng \etal \shortcite{Yuanzheng+2018} employed a conditional GAN to color illustration line art using scribbles from the user. This approach was improved by Zhang \etal \cite{Zhang+2018} who proposed a two-stage sketch colorization for illustration: First the user splashes colors on a sketch, then, the system generates a draft using a GAN that the user can correct mistakes with edits propagated using a refinement GAN. The illustration dataset by Zhang \etal \shortcite{danbooru2021} was used for training, and it has also become a staple for many other methods.
In a different approach for manga colorization was proposed in \cite{Silva+2019,Shimizu+2021}, where the user, instead of scribbling or splashing colors, provides an input image with basic colors. In particular, Shimizu \etal \cite{Shimizu+2021} showed that providing a flat colored image of the sketch image can generate high-quality colorization with few training data. Note that this flat filling colorization can also be automated using deep learning \cite{Zhang+2021} and user inputs for complex line art.
In this domain, researchers have also focused on specific parts to colorize using user inputs \cite{Kenta+2020}, or combined with text tags \cite{Kim+2019}. While at first, this might appear appropriate for animation, as they do not require input per frame, they suffer from limitations when it comes to artistic control. Both remove control of lighting conditions, and the first removes control over pose and expression by virtue of being an inpainting method, while the second entirely removes control over shading. Akinobu \etal \shortcite{Akinobu+2021} \ignore{proposed one of the few methods that} tackle anime colorization using a few-shot strategy and an ad-hoc sampling method for patches tailored for anime\ignore{ exploiting their peculiar drawing lines}. 

An emerging topic is manga generation\ignore{starting from noise}  using adversarial training and some character parameters \cite{Jin+2017}, photos \cite{Wang+2020, Su+2021}, and sketches \cite{You+2019}. Although they can generate high-quality results, they lack precise artistic control.

Recently researchers have applied classic topics such as segmentation \cite{Zhang+2020b} and clustering \cite{Nir+2022} to illustration and cartoon art. Nir \etal \shortcite{Nir+2022} proposed a method to learn a style-specific semantic representation \ignore{more suitable }for animated content using self-supervision.
However, it needs to be trained separately on each production and is more suited for tracking characters and not their designs.
In our work, we introduce a more flexible semantic clustering that decouples the style of the anime from the content. The content is clusterized in an art-style normalized Euclidean space, where distances between these portraits correspond to a measure of character design similarity in a production. 
	\begin{figure*}
    \centering
    \includegraphics[width=\textwidth]{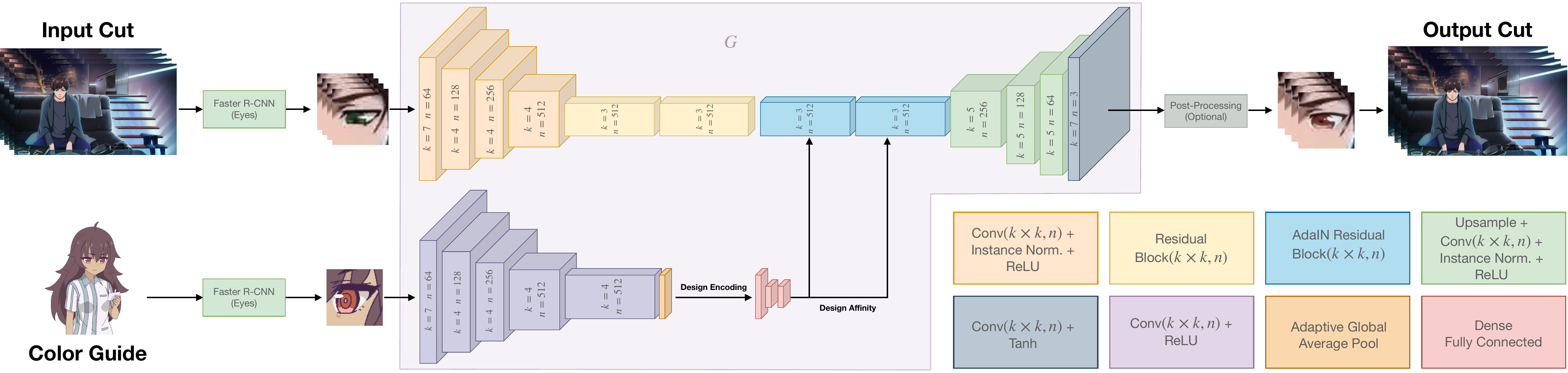}
    \caption[Context-Aware Inference]{\textbf{Context-Aware Inference}. Eyes are detected in a sequence of frames and a color guide, then fed to our context-aware redrawer $G$, and the resulting styled eyes are post-processed into the original art. The whole process can run in real-time.}
    \label{fig:redraw-usage}
\end{figure*}

\section{Method} \label{sec:method}
\begin{figure*}[t]
    \centering
    \includegraphics[width=\textwidth]{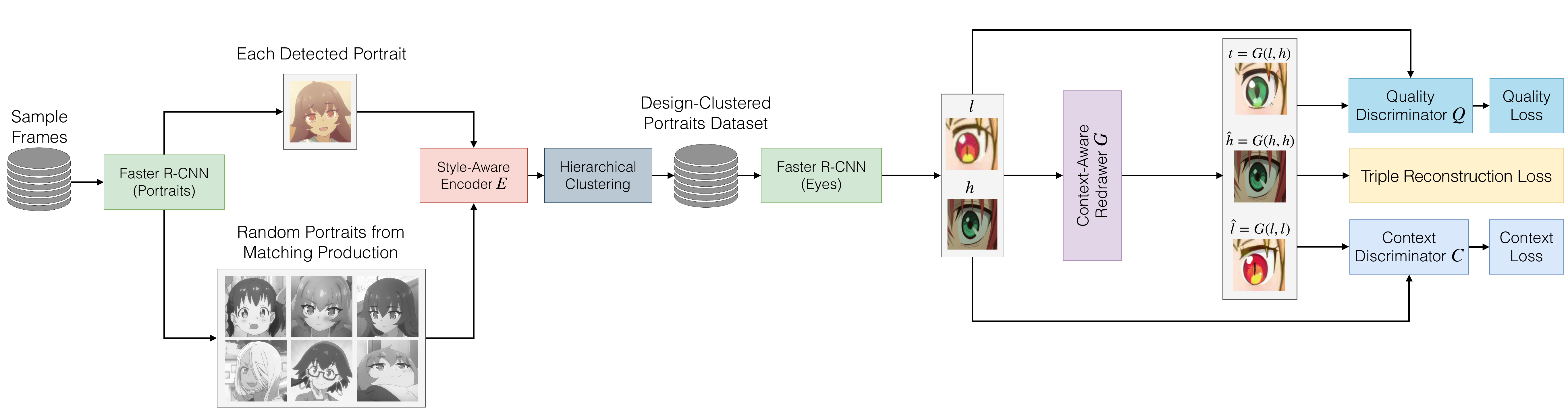}
    \vspace{-5mm}
    \caption[Context-Aware Training]{\textbf{Context-Aware Training}. Detected portraits are clustered using our style-aware encoder (left), with additional portraits as style guides. Content and style images are extracted from portraits of different designs (center). These are used to generate 3 redrawings from different combinations of these input pairs, which are judged by multiple losses, including two multi-class discriminators (right).}
    \label{fig:redraw-training}
\end{figure*}

\noindent Our approach for enhancing animated eyes takes as input the animation frames to be improved and a character color guide; see Fig.~\ref{fig:redraw-usage}. We use an unsupervised convolutional network trained alongside classification networks, capable of telling designs apart, as its adversaries. Using such a model requires artists to manually associate regions to redraw and color guides manually, which is not practical. Even more problematic, to train this type of adversarial structure, one normally uses pairs of these images, labeled into different classes (character designs). In particular, to ensure our model is capable of generalizing to new designs, we need to train on a large enough variety of them. Yet, art direction is not easily available and generally not created in high enough quantities that would be needed for a robust training. Moreover, manually tagging and cropping this data would be extremely labor intensive and hard to replicate. To address these issues we propose a novel character design clustering method, and use it to automatically infer training data from random frames, thus solving the association problem. As such, Re:Draw does not require internal production data for training, instead it only only requires a set of random sampled frames from different productions; see Fig.~\ref{fig:redraw-training}.

Image in-painting has shown to be capable of completing missing regions, yet predictions based only on the surrounding of the area to be redrawn do not allow artists to finely control the output results using art or style direction examples. Image-to-image translation and style transfer are capable of using both of these inputs, yet existing work is incapable of generating art that fits and correctly matches within the actual context of the drawing: they can be very unreliable in preserving the  artwork pose. For these reasons, we introduce context-aware translation. We make use of two adversarial discriminators built using partial convolutions, allowing them to weight images differently and independently, and a novel triple reconstruction loss based on the concept of the generation of image triplets.

\subsection{Dataset Generation} \label{sec:data}
\noindent We will now describe how we avoid the need for production data, by automatically clustering art by character design and then further splitting it into low- and high-levels of detail. This categorized dataset 
is required during the training phase of our context-aware translation model, which involves solving multiple adversarial classification tasks simultaneously.

\paragraph{Object Detection} 
We first train an object-detection network -- we use the well-established Faster R-CNN network \cite{ren2015faster} -- to identify character faces and details in them (such as eyes) and run it on the randomly sampled frames. This results in a dataset of character faces in a variety of poses, split by the sources they were sampled from.

\paragraph{Style-Aware Clustering} \label{sec:clustering} 
Although re-identification of human faces is a long studied topic \cite{balaban2015deep}, we found existing work to be ineffective at automatically identifying animated characters not seen during training. We attribute this to the fact that, while human faces have a consistent and predictable structure, animated characters are not restrained by the laws of reality and thus present a much higher variance: in a given production, characters might have a very similar look and feel, while in another production they might vary widely in structure and shape \fullsource{(see Fig.~\ref{figure:character-structure})}.

\fullsource{
    \begin{figure}
        \centering
        \centering
        \begin{subfigure}{0.12\columnwidth}
            \centering
            \includegraphics[width=\textwidth]{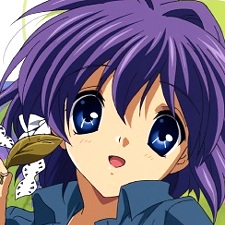}
        \end{subfigure}
        \begin{subfigure}{0.12\columnwidth}
            \centering
            \includegraphics[width=\textwidth]{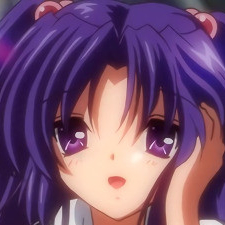}
        \end{subfigure}
        \begin{subfigure}{0.12\columnwidth}
            \centering
            \includegraphics[width=\textwidth]{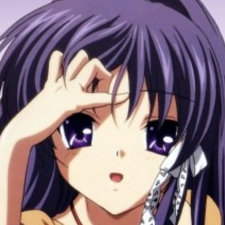}
        \end{subfigure}
        \hspace{2em}
        \begin{subfigure}{0.12\columnwidth}
            \centering
            \includegraphics[width=\textwidth]{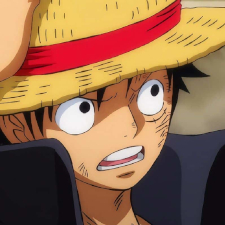}
        \end{subfigure}
        \begin{subfigure}{0.12\columnwidth}
            \centering
            \includegraphics[width=\textwidth]{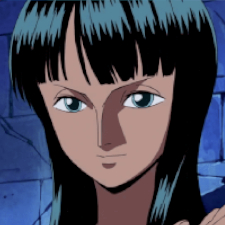}
        \end{subfigure}
        \begin{subfigure}{0.12\columnwidth}
            \centering
            \includegraphics[width=\textwidth]{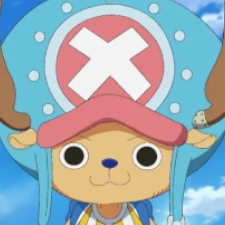}
        \end{subfigure}
        \caption{\textbf{Design Variety.} Characters in some productions have very similar structures (left \shortcite{clannad}), and yet others may present a high variety of designs (right \shortcite{onepiece}). This hinders traditional facial recognition.}
        \label{figure:character-structure}
    \end{figure}
}

\begin{figure}
    \centering
    \includegraphics[width=\columnwidth]{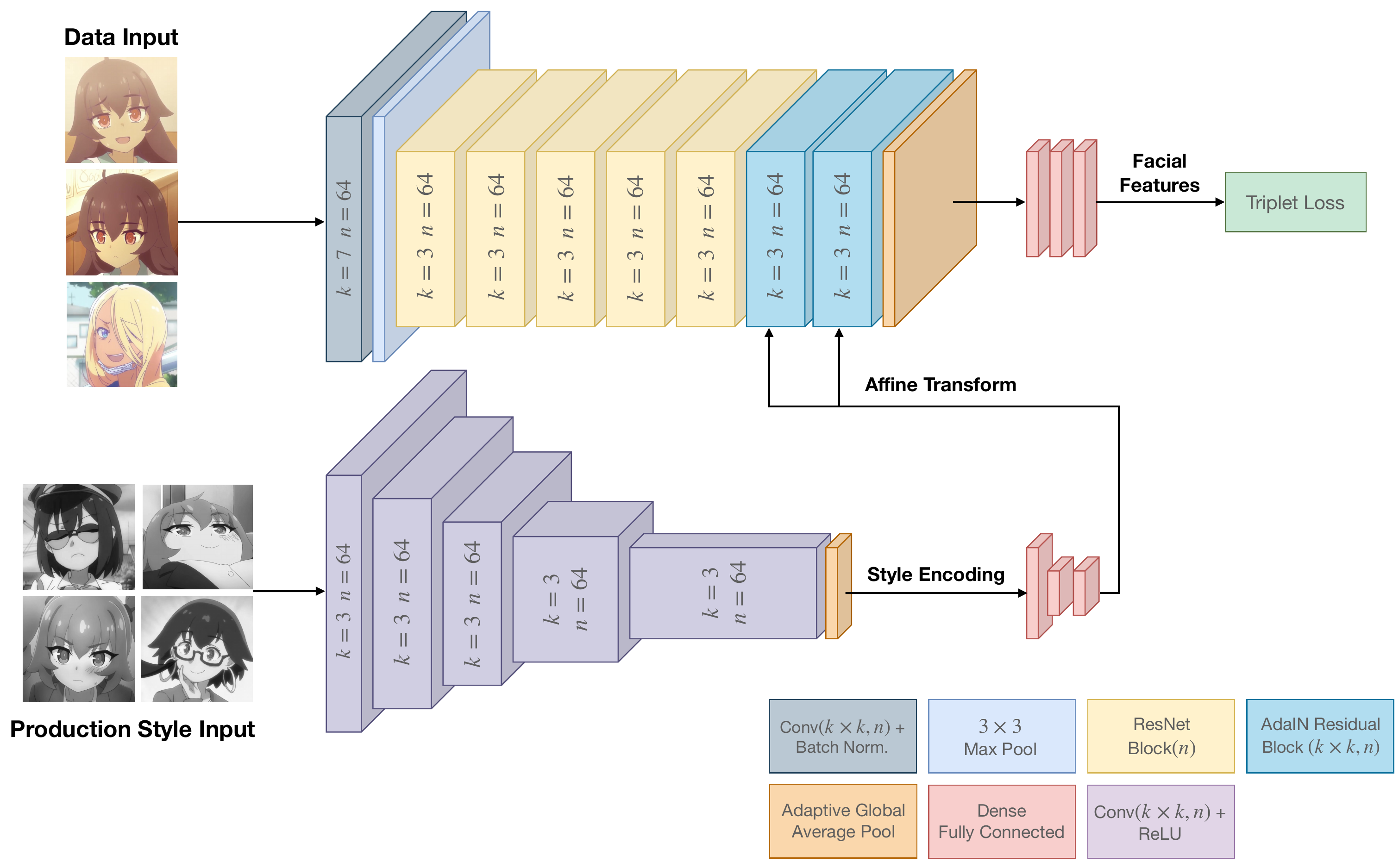}
    \caption{\textbf{Style-Aware Encoder}. We train a network capable of generalized character design recognition using triplet loss. Unlike traditional facial recognition, where only a set of labeled portraits is seen, we additionally input random unlabeled portraits to account for production style.}
    \label{fig:cluster-training}
\end{figure}

To solve this issue, we improve upon the state of the art of character recognition by combining ideas from facial recognition and image-to-image translation. We propose a supervised network that, unlike existing work, maps character portraits to an art-style normalized Euclidean space, where distances between these portraits correspond to a measure of character design similarity within its production. It takes as input character portraits to be mapped and a collection of random portraits from the same production for normalization estimation.

As shown in Fig.~\ref{fig:cluster-training}, latent representations of both inputs are estimated: we compute the content representation using a ResNet \cite{He+2016} encoder -- which is well established for object recognition -- and the production representation using a convolutional encoder. The latter is done using only the lightness in $l\alpha\beta$ color space \cite{reinhard2001color}, as we found that the normalization input works better if it only contains the main shape information, so we use a color space to decorrelate it from color variation. This style-latent representation is then used to compute a set of affine transformations, with the goal of mapping the encoding from an absolute Euclidean-space representation of portraits to the style-normalized one. This mapping is done using Adaptive Instance normalization \cite{Huang+2017} on the content-input latent representation. This finally results in $32$ parameters per portrait thanks to the linear layers, which are then clustered using traditional hierarchical clustering, with unweighted pair group method, arithmetic mean and Euclidean distance. These methods and parameters were chosen by testing the rate of correct clustering across a validation dataset.

To train this network $E$ and ensure the content-encoding output respects the desired intra and inter-class proprieties of the normalized Euclidean space, we use the option of Triplet Margin Loss \cite{balntas2016learning, hermans2017defense} -- that is, given a pair of portraits from the same design $\{P_1, P_2\}$ and one from another $P_3$ but from the same production $\mathbb{P}$, we minimize the distance from the first two, while maximizing the distance of the third:
{ \small
\vspace{-0.4mm}
\begin{align}
    \underset{E}{\mathrm{argmin}}\,\, max
    \bigl\{
    &\|E(P_1, \mathbb{P}) - E(P_2, \mathbb{P})\|^2 - \\
    &\|E(P_1, \mathbb{P}) - E(P_3, \mathbb{P})\|^2 + 1, 0\notag
    \bigr\}
\end{align}
}%
This means that, during training, character portraits must be provided in sets of three. We also ensure that the total training weight of each production style and of each character design within each style is the same, to further help with generalization. While it is technically possible to train $E$ in conjunction with the context-aware translation, it is more computationally efficient to train $E$ first and then freeze it to train the remainder networks.

\paragraph{Level-of-Detail Split}
Having the portraits organized by character design, we extract the intended art details (eyes in our case study) from them using a Faster R-CNN network again, but now with the knowledge of their corresponding designs. We standardize all extracted art to the same size. Then, we exploit the fact that characters are often drawn with different levels of detail, depending on their prominence on screen, to discriminate between low and high details regions. So, after empirical observation, regions with less than $0.31\%$ pixels were assumed to be low-detail and to be redrawn, while regions with more than $0.48\%$ pixels were used as art direction examples (see Tab.~\ref{tab:data}). 

\begin{table}
    \caption[Generated Dataset]{Statistics of our animation frames dataset generation.} \label{tab:data}
    \centering
    \resizebox{\columnwidth}{!}{
        \begin{tabular}{|rl|rl|}
            \hline
            Selected Productions & 48    & Predicted Elements & 35884 \\
            Marked Designs       & 23    & Element Mean Size  & 14395 \small{$pixels^2$}\\
            Sampled Frames       & 14338 & Content Images     & 20374 \\
            Predicted Designs    & 476   & Style Images       & 15510 \\
            \hline
        \end{tabular}
    }
\end{table}

\subsection{Context-Aware Translation} \label{sec:losses}
\noindent Having generated the dataset, we can now train the redrawing model. We want to find a function $x^{+} = g(x, \mathbb{S})$ that, given a low-detail content image $x$ and color guide $\mathbb{S}$ (equivalent to style images in a style transfer context), is capable of outputting a higher detail version $x^{+}$ of $x$. This leaves us with two conflicting goals: we want the translated artwork $x^{+}$ to match the provided design $\mathbb{S}$ and its level of detail, but to still fit within the original drawing of $x$.

We define an image-to-image network $G$, to be trained as a context-aware redrawer, with the purpose of approximating $g$. As illustrated in Fig.~\ref{fig:redraw-usage}, it is composed of a convolutional encoder-decoder structure with an additional style encoder. The latter matches exactly the encoder described in Section \ref{sec:clustering} and is used to compute a set of affine transformations that control the Adaptive Instance normalization in the decoder.

\begin{figure}[b]
    \centering
    \footnotesize{\begin{tabular}{
    >{\raggedleft\arraybackslash}m{0.15cm}@{\hspace{2pt}} 
    >{\centering\arraybackslash}m{1.3cm}@{\hspace{2pt}} 
    >{\centering\arraybackslash}m{1.3cm}@{\hspace{2pt}} 
    >{\centering\arraybackslash}m{1.3cm}@{\hspace{2pt}}
    }
    $l$
     & \includegraphics[width=0.06\textwidth]{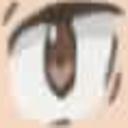}
     & \includegraphics[width=0.06\textwidth]{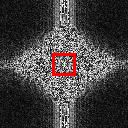}
     & \includegraphics[width=0.06\textwidth]{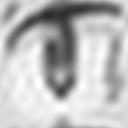} \\
    $\hat{l}$
     & \includegraphics[width=0.06\textwidth]{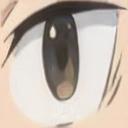}
     & \includegraphics[width=0.06\textwidth]{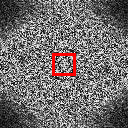}
     & \includegraphics[width=0.06\textwidth]{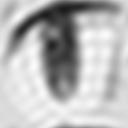} \\
    & Example & Coefficients & Filtered \\
    \end{tabular}}
    \caption[Low-Pass Filter]{\textbf{Low-Pass Filter}. $\hat{l}$ is computed during training from enhancing $l$. Right shows the result of our low pass filter.}
    \label{fig:fourier}
\end{figure}

\paragraph{Triple Reconstruction Loss}
Let $l$ be a low-detail image and $h$ a high-detail one, each sampled from different designs $\mathbb{L}$ and $\mathbb{H}$, respectively. Our approach is to train the redrawer as an image translation problem such that $t = G(l, h)$ outputs the result of applying design $\mathbb{H}$ to $l$.

To help $G$ learn a translation model and ensure it maintains the local structure of $l$, a second output $\hat{l} = G(l, l)$ is frequently used as part of a reconstruction loss \cite{Xun+2018}. However, this is not appropriate for our problem, as we are not interested in the network producing low detail images. We propose a novel reconstruction loss that analyses a total of three generated images:
\vspace{-0.4mm}
{ \small 
\begin{equation}
    \mathcal{L}_{R} = [ \|h - G(h,h)\| + \|F(l) - F(t)\| + \|F(l) - F(\hat{l})\| ]^1_1
    \text{,}
\end{equation}
}%
\noindent where $F(x)$ is a low-pass image filter applied on the lightness of the image $x$, implemented by converting to frequency space using fast Fourier transform and remove any frequencies above a set threshold (0.06, as shown in Fig~\ref{fig:fourier}). The basis is that, by removing high frequencies and color changes, differences between low-detail and high-detail images are ignored as well, allowing us to create a reconstruction loss in low-detail images.

\begin{figure}
    \centering
    \includegraphics[width=\linewidth]{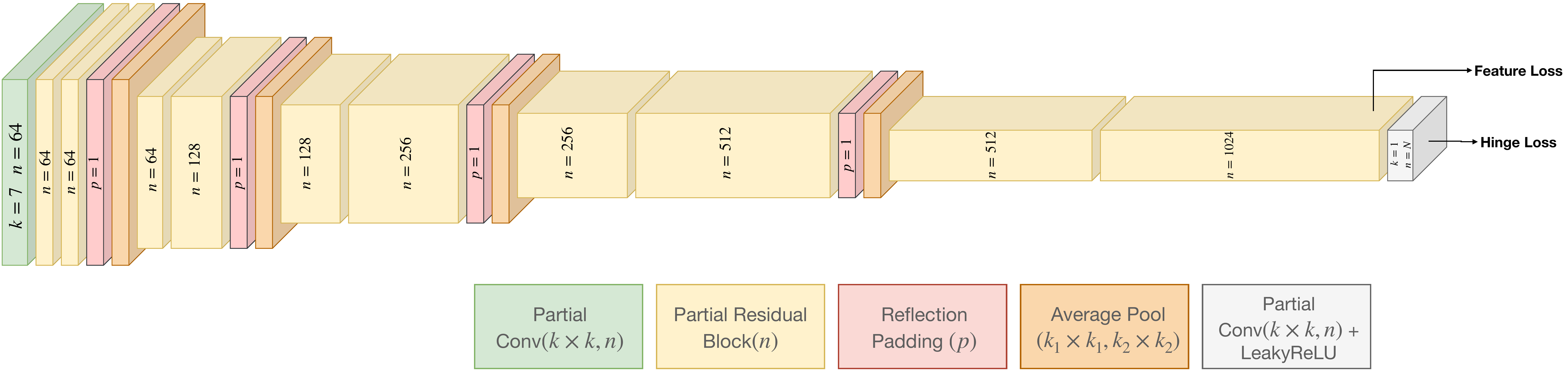}
    \caption{\textbf{Discriminator Architecture}. Both our quality and context discriminators share this structure.}
    \label{fig:discriminator}
\end{figure}

\paragraph{Adversarial Discriminators}
We address our aforementioned conflicting goals by using two independent image multi-task classifiers instead: a {\em quality discriminator} $Q$ judging whether the output is high detail and matches the intended design, and a {\em context discriminator} $C$  judging whether it fits within the original artwork and its own design, irrespective of detail-level.

To achieve this purpose, as we show in Section \ref{sec:ablation}, we need to train each discriminator differently, despite sharing many commonalities: both have the same partial convolutional structure and are trained using hinge loss with R1 regularization \cite{mescheder2018training}, to prevent over-fitting and mode collapse. This results in the following losses for penalizing wrong classifications of positive $\mathcal{L}_{P}$ and negative $\mathcal{L}_{N}$ examples:
{\small 
\vspace{-0.4mm}
\begin{equation}
    \begin{gathered}
        \mathcal{L}_{P}(x, \mathbb{S}) = [ \max(0, 1 - D(x)_\mathbb{S}) + \gamma \|\nabla D(x)_\mathbb{S}\|^2]_1 \\
        \mathcal{L}_{N}(x, \mathbb{S}) = [ \max(0, 1 + D(x)_\mathbb{S}) ]_1
    \end{gathered}
    \text{,}
\end{equation}
}%
\noindent where $D \in \{Q,C\}$ can be any one of the discriminators, $D(x)_\mathbb{S}$ is the discriminator's score of image input $x$ for design class $\mathbb{S}$, $\nabla D(x)_\mathbb{S}$ its derivative used for R1 regularization and $\gamma = 10$ (R1 standard weight). That is, $D$ should converge to $[0,1]$ for positive entries and to $[-1,0]$ otherwise. The discriminators are then given different input masks to weight disparate regions of images differently: the quality discriminator $Q$ focuses on the interior of the redrawn region, while the context discriminator $C$ focuses on the opposite, including an outer border that is not redrawn. They meet and oppose each other in the intersection of their two regions. Finally, they are trained to judge the training image pairs $\{l,h\}$ and the generated triplets $\{t, \hat{l}, \hat{h}\}$ such that $Q$ looks for high detail output, while $C$ tries to tell real and generated art apart:
{\small
\begin{equation}
    \begin{gathered}
        \underset{Q}{\mathrm{argmin}}\, \mathcal{L}_{P}(h,\mathbb{H}) + \frac{\mathcal{L}_{N}(l,\mathbb{H}) + \mathcal{L}_{N}(t,\mathbb{H})}{2} \\
        \underset{C}{\mathrm{argmin}}\, \mathcal{L}_{P}(h,\mathbb{H}) +  \mathcal{L}_{N}(t,\mathbb{H}) + \mathcal{L}_{P}(l,\mathbb{L}) + \mathcal{L}_{N}(\hat{l},\mathbb{L})
    \end{gathered}
\end{equation}
}%
\noindent That is, $Q$ attempts to learn to identify real high-detail images as positive examples, and low-detail or generated art as negative ones; while $C$ attempts to identify real as positive and generated as negative, independently of detail. Images are always judged for the design class they are supposed to belong to. Then, to train the redrawer network $G$ using these discriminators, we use hinge loss with a latent feature loss to regularize the adversarial training. Let $D^F$ be the latent features computed by a discriminator $D$ in a hidden layer, and $s$ a sampled image (either $l$ or $h$) from the given design class $\mathbb{S}$. The adversarial loss function of each discriminator $D$ becomes:
{\small
\begin{equation} \label{eq:L_D}
    \mathcal{L}_{D}(x, \mathbb{S}) = [1 - D(x)_\mathbb{S}]^1_1 + [D^F(x)_\mathbb{S} - D^F(s)_\mathbb{S}]^1_1
\end{equation}
}%
\noindent The use of hinge loss, R1 regularization and feature matching loss have been used in different forms in image-to-image translation problems \cite{liu2019few, Saito+2020}. Just as with training the discriminators themselves, our contribution is how these are then used to train a redrawer capable of addressing our problem. We combine our novel reconstruction loss with two adversarial losses to verify discriminator conditions and another two to ensure reconstruction persistence, where $\mathcal{L}_Q$ and $\mathcal{L}_C$ are the adversarial functions of each discriminator, defined in Equation \ref{eq:L_D}:
\vspace{-.4em}
{\small 
\begin{equation}
    \underset{G}{\mathrm{argmin}}\, \mathcal{L}_{R} + \mathcal{L}_{Q}(t, \mathbb{H}) + \mathcal{L}_{Q}(\hat{l}, \mathbb{L}) + \mathcal{L}_{C}(t, \mathbb{L}) + \mathcal{L}_{C}(\hat{h}, \mathbb{H})
\end{equation}
}%
\paragraph{Post-Processing} \label{sec:post}
To further ensure image regions generated fit within the original image, we apply a few image processing operations: after re-sampling the network output to the original resolution, we apply color transfer \cite{reinhard2001color} and place it into the original image using Poisson image editing \cite{perez2003poisson}.

	\begin{figure*}
    \centering
    \begin{subfigure}{.46\linewidth}
         \centering
         \fullsource{\includegraphics[width=\textwidth]{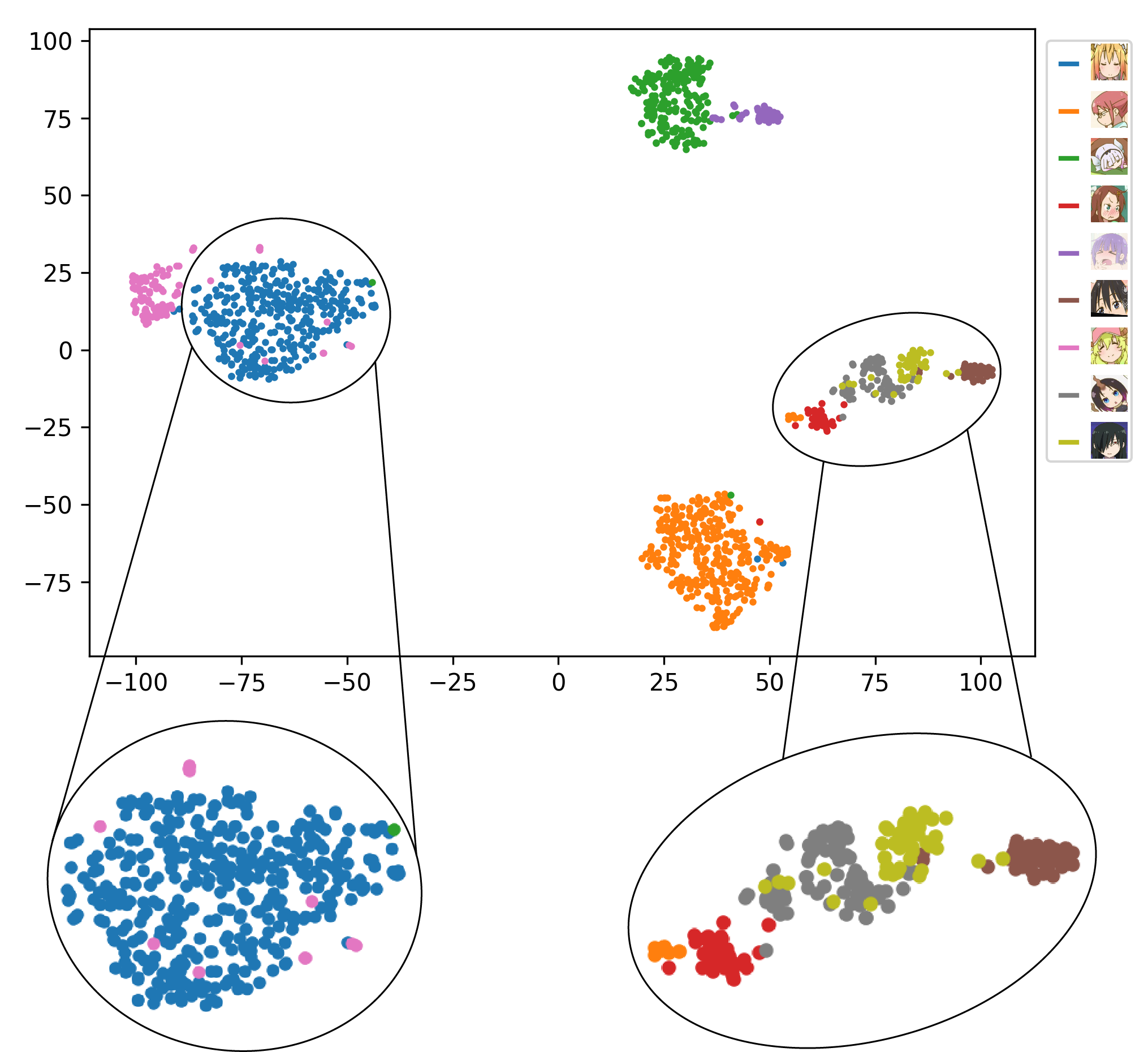}}
         \opensource{\includegraphics[width=\textwidth]{img/graphs/clustering/theirs-censored.png}}
         \vspace{-2em}
         \caption{FaceNet}
     \end{subfigure}
     \begin{subfigure}{.46\linewidth}
         \centering
         \fullsource{\includegraphics[width=\textwidth]{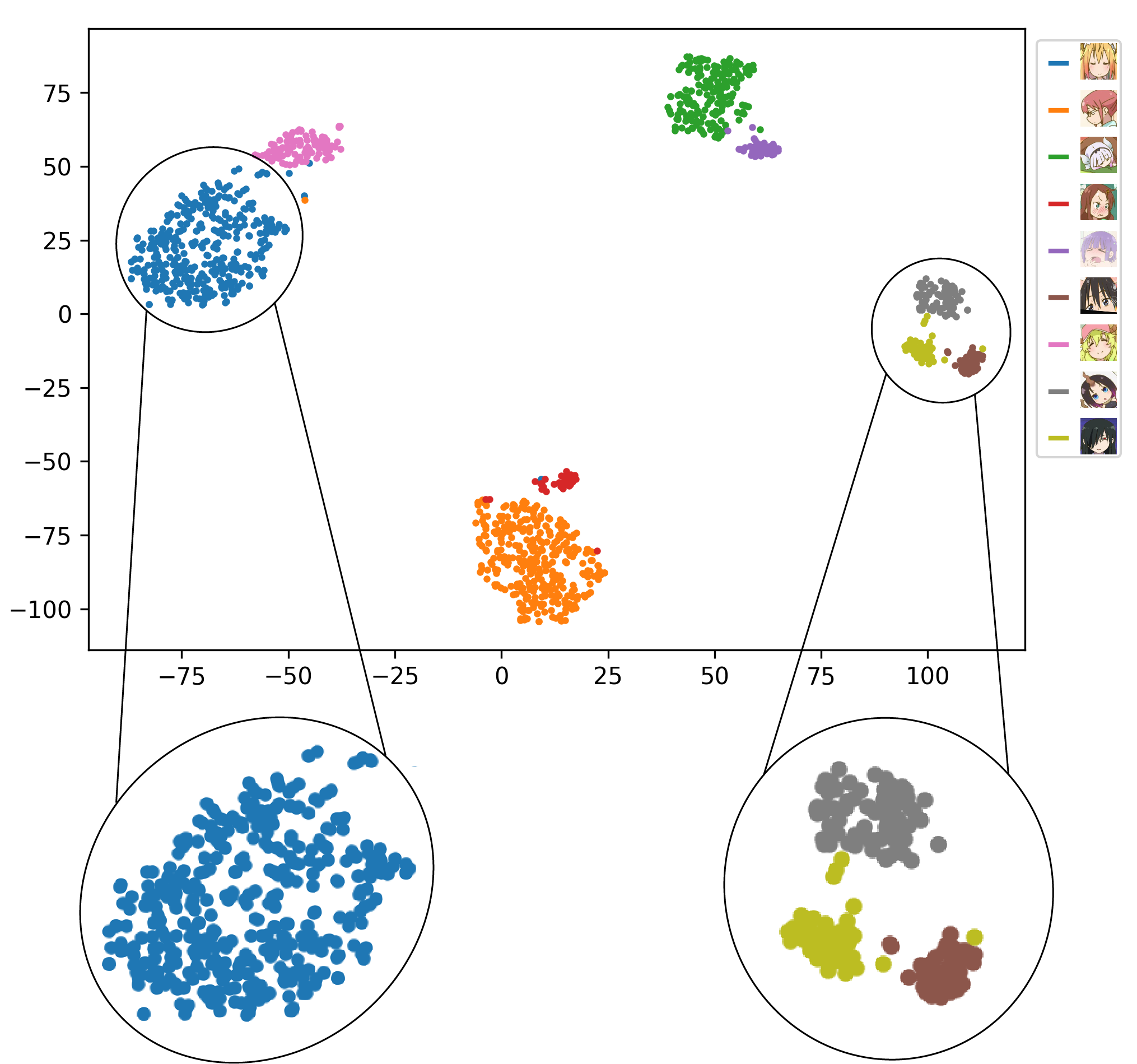}}
         \opensource{\includegraphics[width=\textwidth]{img/graphs/clustering/ours-censored.png}}
         \vspace{-2em}
         \caption{Ours}
     \end{subfigure}
    \caption[Clustering Validation]{\textbf{Clustering Validation} 2D visualization of characters faces from a production not seen during training \cite{dragonmaid}, generated using T-Distributed Stochastic Neighbor Embedding \cite{van2008visualizing} on the latent spaces learned using FaceNet \cite{schroff2015facenet} and our proposed encoder network. Colors correspond to ground truth labels. As shown, character differentiation is clearer in our learned space.}
    \label{fig:tsne}
\end{figure*}

\begin{figure*}
     \centering
     \footnotesize{\begin{tabular}{>{\raggedleft\arraybackslash}m{1.5cm} >{\centering\arraybackslash}m{1.2cm}@{\hspace{6pt}} >{\centering\arraybackslash}m{1.2cm}@{\hspace{6pt}} >{\centering\arraybackslash}m{1.2cm}@{\hspace{6pt}} >{\centering\arraybackslash}m{1.2cm}@{\hspace{6pt}} >{\centering\arraybackslash}m{1.2cm}@{\hspace{6pt}} >{\centering\arraybackslash}m{1.2cm}@{\hspace{6pt}} >{\centering\arraybackslash}m{1.2cm}@{\hspace{6pt}} >{\centering\arraybackslash}m{1.2cm}@{\hspace{6pt}} >{\centering\arraybackslash}m{1.2cm}@{}}
         Input
          & \includegraphics[width=0.075\textwidth]{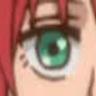}
          & \includegraphics[width=0.075\textwidth]{img/ablation/eye-crops/30/1.jpg}
          & \includegraphics[width=0.075\textwidth]{img/ablation/eye-crops/30/1.jpg}
          & \includegraphics[width=0.075\textwidth]{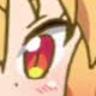}
          & \includegraphics[width=0.075\textwidth]{img/ablation/eye-crops/5/0.jpg}
          & \includegraphics[width=0.075\textwidth]{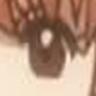}
          & \includegraphics[width=0.075\textwidth]{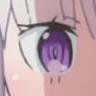}
          & \includegraphics[width=0.075\textwidth]{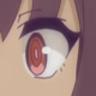}
          & \includegraphics[width=0.075\textwidth]{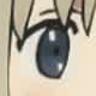} \\
         Design
          & \includegraphics[width=0.075\textwidth]{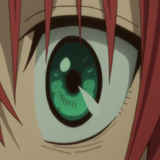}
          & \includegraphics[width=0.075\textwidth]{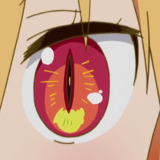}
          & \includegraphics[width=0.075\textwidth]{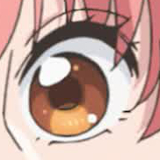}
          & \includegraphics[width=0.075\textwidth]{img/ablation/eye-crops/dragonMaid.jpg}
          & \includegraphics[width=0.075\textwidth]{img/ablation/eye-crops/magusBride.jpg}
          & \includegraphics[width=0.075\textwidth]{img/ablation/eye-crops/wotakoi.jpg}
          & \includegraphics[width=0.075\textwidth]{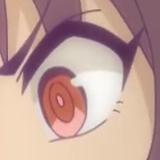}
          & \includegraphics[width=0.075\textwidth]{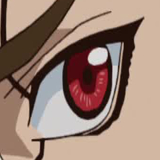}
          & \includegraphics[width=0.075\textwidth]{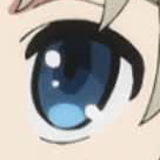} \\ \\
         FUNIT \cite{liu2019few}
          & \includegraphics[width=0.075\textwidth]{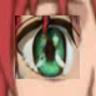}
          & \includegraphics[width=0.075\textwidth]{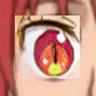}
          & \includegraphics[width=0.075\textwidth]{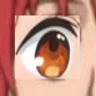}
          & \includegraphics[width=0.075\textwidth]{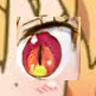}
          & \includegraphics[width=0.075\textwidth]{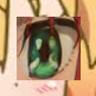}
          & \includegraphics[width=0.075\textwidth]{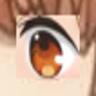}
          & \includegraphics[width=0.075\textwidth]{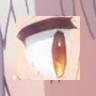}
          & \includegraphics[width=0.075\textwidth]{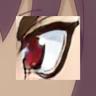}
          & \includegraphics[width=0.075\textwidth]{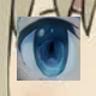} \\
         + Our\linebreak Clustering
          & \includegraphics[width=0.075\textwidth]{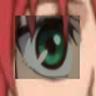}
          & \includegraphics[width=0.075\textwidth]{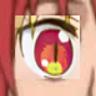}
          & \includegraphics[width=0.075\textwidth]{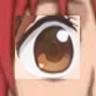}
          & \includegraphics[width=0.075\textwidth]{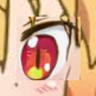}
          & \includegraphics[width=0.075\textwidth]{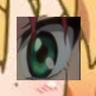}
          & \includegraphics[width=0.075\textwidth]{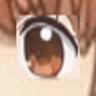}
          & \includegraphics[width=0.075\textwidth]{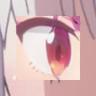}
          & \includegraphics[width=0.075\textwidth]{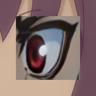}
          & \includegraphics[width=0.075\textwidth]{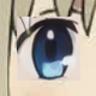} \\
         + Quality \& Context
          & \includegraphics[width=0.075\textwidth]{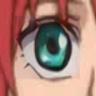}
          & \includegraphics[width=0.075\textwidth]{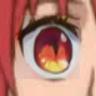}
          & \includegraphics[width=0.075\textwidth]{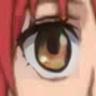}
          & \includegraphics[width=0.075\textwidth]{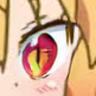}
          & \includegraphics[width=0.075\textwidth]{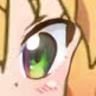}
          & \includegraphics[width=0.075\textwidth]{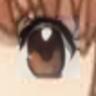}
          & \includegraphics[width=0.075\textwidth]{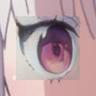}
          & \includegraphics[width=0.075\textwidth]{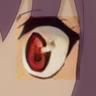}
          & \includegraphics[width=0.075\textwidth]{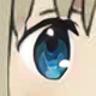} \\
         Ours
          & \includegraphics[width=0.075\textwidth]{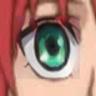}
          & \includegraphics[width=0.075\textwidth]{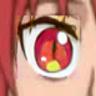}
          & \includegraphics[width=0.075\textwidth]{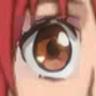}
          & \includegraphics[width=0.075\textwidth]{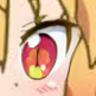}
          & \includegraphics[width=0.075\textwidth]{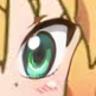}
          & \includegraphics[width=0.075\textwidth]{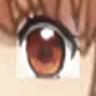}
          & \includegraphics[width=0.075\textwidth]{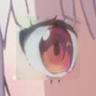} 
          & \includegraphics[width=0.075\textwidth]{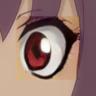}
          & \includegraphics[width=0.075\textwidth]{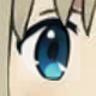} \\
         Ours + Post
          & \includegraphics[width=0.075\textwidth]{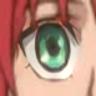}
          & \includegraphics[width=0.075\textwidth]{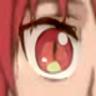}
          & \includegraphics[width=0.075\textwidth]{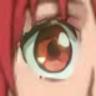}
          & \includegraphics[width=0.075\textwidth]{img/ablation/eye-crops/5/0-dragonMaid-4_ours_.jpg}
          & \includegraphics[width=0.075\textwidth]{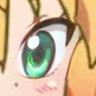}
          & \includegraphics[width=0.075\textwidth]{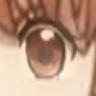}
          & \includegraphics[width=0.075\textwidth]{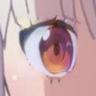}
          & \includegraphics[width=0.075\textwidth]{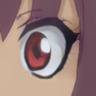}
          & \includegraphics[width=0.075\textwidth]{img/ablation/eye-crops/82/1-franxx-4_ours_.jpg} \\
         & (a) & (b) & (c) & (d) & (e) & (f) & (g) & (h) & (i) \\
     \end{tabular}}
     \caption[Ablation of Contributions]{\textbf{Ablation of Contributions}. We compared our method with traditional image translation by progressively introducing our contributions (rows 4 to 6). The synthetic data extrapolated using our style-aware clustering prevents over-fitting (a,e) and allows generalization to unseen designs (g,h,i). Further introducing our dual discriminators allows redrawn areas to maintain artwork fit, but sacrifices detail (a,b,c,e,i) and ability to redesign shape or color (g,h). Finally, introducing the triplet reconstruction loss brings that expressiveness back.}
     \label{fig:comparison}
 \end{figure*}
 \begin{figure}
     \centering
     \footnotesize{\begin{tabular}{>{\centering\arraybackslash}m{.8cm} >{\centering\arraybackslash}m{1.2cm}@{\hspace{6pt}} >{\centering\arraybackslash}m{1.2cm}@{\hspace{6pt}} >{\centering\arraybackslash}m{1.2cm}@{\hspace{6pt}} >{\centering\arraybackslash}m{1.2cm}}
         No $\Delta Q$/$\Delta C$
         & \includegraphics[width=0.075\textwidth]{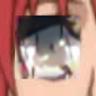}
         & \includegraphics[width=0.075\textwidth]{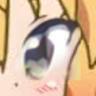}
         & \includegraphics[width=0.075\textwidth]{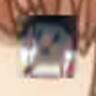}
         & \includegraphics[width=0.075\textwidth]{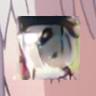}\\
         No $\mathcal{L}_Q$
         & \includegraphics[width=0.075\textwidth]{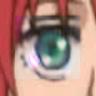}
         & \includegraphics[width=0.075\textwidth]{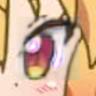}
         & \includegraphics[width=0.075\textwidth]{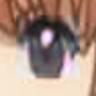}
         & \includegraphics[width=0.075\textwidth]{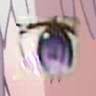}\\
         No $\mathcal{L}_R$
         & \includegraphics[width=0.075\textwidth]{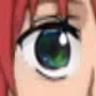}
         & \includegraphics[width=0.075\textwidth]{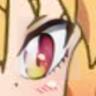}
         & \includegraphics[width=0.075\textwidth]{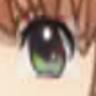}
         & \includegraphics[width=0.075\textwidth]{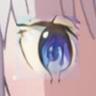} \\
         No $\mathcal{L}_C$
         & \includegraphics[width=0.075\textwidth]{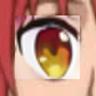}
         & \includegraphics[width=0.075\textwidth]{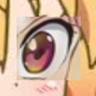}
         & \includegraphics[width=0.075\textwidth]{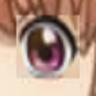}
         & \includegraphics[width=0.075\textwidth]{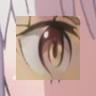}\\
         & (b) & (d) & (f) & (g) \\
     \end{tabular}}
     \caption[Ablation of Losses]{\textbf{Ablation of Losses}. We tested removing each loss during training of redrawer $G$ on examples from Fig.~\ref{fig:comparison}. It shows $\mathcal{L}_Q$ and $\mathcal{L}_R$ are crucial for useful results. Most notably, it reveals $\mathcal{L}_C$ makes output much more predictable than otherwise expected, and this stability is what enables the use of our less explicit $\mathcal{L}_R$.}
     \label{fig:losses}
\end{figure}
\begin{figure}
     \centering
     \footnotesize{\begin{tabular}{>{\raggedleft\arraybackslash}m{1cm} >{\centering\arraybackslash}m{1.2cm}@{\hspace{6pt}} >{\centering\arraybackslash}m{1.2cm}@{\hspace{6pt}} >{\centering\arraybackslash}m{1.2cm}@{\hspace{6pt}} >{\centering\arraybackslash}m{1.2cm}}
         Style Transfer
         & \includegraphics[width=0.075\textwidth]{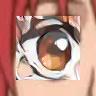}
         & \includegraphics[width=0.075\textwidth]{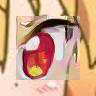}
         & \includegraphics[width=0.075\textwidth]{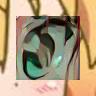}
         & \includegraphics[width=0.075\textwidth]{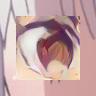}\\
         S.Transfer\linebreak+ Post
         & \includegraphics[width=0.075\textwidth]{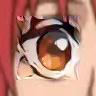}
         & \includegraphics[width=0.075\textwidth]{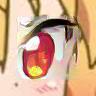}
         & \includegraphics[width=0.075\textwidth]{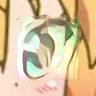}
         & \includegraphics[width=0.075\textwidth]{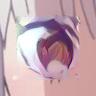}\\
         FUNIT\linebreak+ Post
         & \includegraphics[width=0.075\textwidth]{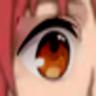}
         & \includegraphics[width=0.075\textwidth]{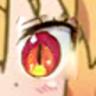}
         & \includegraphics[width=0.075\textwidth]{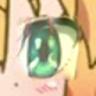}
         & \includegraphics[width=0.075\textwidth]{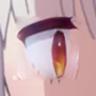}\\
         Ours
         & \includegraphics[width=0.075\textwidth]{img/ablation/eye-crops/30/1-wotakoi-5_ours+post_.jpg}
         & \includegraphics[width=0.075\textwidth]{img/ablation/eye-crops/5/0-dragonMaid-4_ours_.jpg}
         & \includegraphics[width=0.075\textwidth]{img/ablation/eye-crops/5/0-magusBride-5_ours+post_.jpg}
         & \includegraphics[width=0.075\textwidth]{img/ablation/eye-crops/11/1-lawman-5_ours+post_.jpg}\\
         & (c) & (d) & (e) & (g) \\
     \end{tabular}}
     \caption[Ablation of Related Work]{\textbf{Ablation of Related Work}. We compared our method with style transfer and image-to-image translation with and without post-processing.}
     \label{fig:style-transfer}
     \vspace{4em}
\end{figure}

\begin{figure}
     \centering
     \begin{subfigure}{.16\columnwidth}
         \centering
         \includegraphics[width=\textwidth]{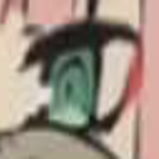}
         \raisebox{1mm}[0mm][0mm]{\footnotesize \textbf{(a)} Input}
     \end{subfigure}
     \begin{subfigure}{.16\columnwidth}
         \centering
         \includegraphics[width=\textwidth]{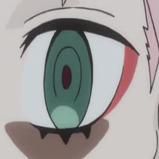}
         \raisebox{1mm}[0mm][0mm]{\footnotesize \textbf{(b)} Design}
     \end{subfigure}
     \begin{subfigure}{.16\columnwidth}
         \centering
         \includegraphics[width=\textwidth]{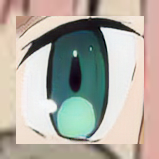}
         \raisebox{1mm}[0mm][0mm]{\footnotesize \textbf{(c)} FUNIT}
     \end{subfigure}
     \begin{subfigure}{.16\columnwidth}
         \centering
         \includegraphics[width=\textwidth]{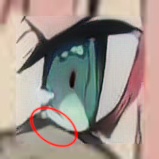}
         \raisebox{1mm}[0mm][0mm]{\footnotesize \textbf{(d)} Ours}
     \end{subfigure}
     \\
     \begin{subfigure}{.32\columnwidth}
         \centering
         \includegraphics[width=\textwidth]{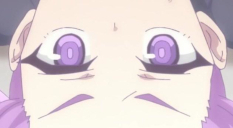}
         \raisebox{1mm}[0mm][0mm]{\footnotesize \textbf{(e)} Input}
     \end{subfigure}
     \begin{subfigure}{.32\columnwidth}
         \centering
         \includegraphics[width=\textwidth]{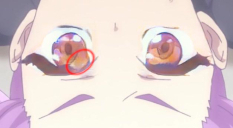}
         \raisebox{1mm}[0mm][0mm]{\footnotesize \textbf{(f)} Ours}
     \end{subfigure}
     \caption[Redrawing Limitations]{\textbf{Redrawing Limitations}. Our method vastly improves upon existing work on handling occlusions; however, it still can struggle with uncommon scenarios (top). As it was trained mostly with upright eyes, fails if input is not pre-rotated to upright position (bottom).}
     \label{fig:limitations}
 \end{figure}

\section{Ablation Experiments} \label{sec:ablation}
\paragraph*{Clustering} We compared our style-aware clustering approach with FaceNet \cite{schroff2015facenet}, trained on the same labeled animated character faces. We statistically analyzed how effective the latent representations learned from either method work on a validation dataset of productions styles not seen during training: we measured the ratio of the average squared norm distance between each point of the same character, and the average squared norm distance between the mean points of each character. The lower this value, the better is the latent space representation in principle. Our method measures a ratio of $1.212e^{-4}$ in the validation data, which outperforms FaceNet's $1.494e^{-4}$ ratio. Interestingly, our method performs similarly to a FaceNet network trained on the validation dataset, whose ratio was $1.209e^{-4}$. This shows our method presents better generalization to unseen data. Furthermore, we show in Fig.~\ref{fig:tsne} how well split the character faces from validation production styles are. The colors of the points represent their ground-truth labels, and there is a visible improvement with our method.

\paragraph*{Redrawing} We evaluated our redrawing approach against neural cross-domain image translation with content and style inputs. While multiple variations of these networks exist \cite{Saito+2020, ojha2021few}, they mostly compete in translation ability and do not address our problems. Thus, we chose as a baseline for comparison Liu \etal\shortcite{liu2019few} method (FUNIT). We progressively introduced each of our novelties to show the importance of each one (Fig~\ref{fig:comparison}): we introduced our style-aware clustering by training FUNIT only on the few manually labeled faces versus the generated dataset. We then added our double-discriminator method, while maintaining FUNIT's traditional reconstruction loss. Finally, we introduced the proposed triplet reconstruction, followed by the post-processing step. FUNIT fails to respect pose/expression and context. Without our large-scale dataset, it often fails to generate realistic art and cannot generalize to designs not seen during training. Our dataset and double discriminator approach solve all of these issues, but reduce the ability of the network to generate highly detailed art. Our novel triplet reconstruction fixes that.

Our method is also very stable and demonstrates temporal coherence, \fullsource{shown in Fig.~\ref{temporal-coherence}}\opensource{see add.materials}, despite not specifically incorporating it in the loss. We believe the reason for this consistency is its emphasis on preserving the intended pose and context of the drawing: as shown in Fig.~\ref{fig:losses}, removing the regularizer from the adversarial losses results in mode-collapse, and removing the quality loss results in a poor auto-encoder as expected. Yet, removing the reconstruction loss does not result in unpredictable output as expected, just an inability to learn useful transformations, and removing the context loss does not result in output reminiscent of FUNIT. Our explanation is that the network, by following the context loss, is being incentivized to exhibit spatial and temporal consistency in its output. This stability is what enables our less explicit form of reconstruction loss to direct the training toward useful results.

We critically evaluate the performance of existing techniques -- namely style-transfer, image-to-image translation, and text-to-image diffusion—against our proposed context-aware translation. Figs.~\ref{fig:teaser} and ~\ref{fig:style-transfer} clearly illustrate the limitations of these existing approaches, thereby reinforcing why our method is a more apt solution for this particular type of problem. Furthermore, \fullsource{Figs. \ref{fig:rwby}, \ref{detail}, \ref{redesign1} and \ref{redesign2}}\opensource{Fig. \ref{redesign-otachan} and add.materials} provide compelling evidence of our method's robustness and versatility in handling a variety of challenging scenarios, further substantiating its suitability for this application. 

The limitations of the network we found boiled down to two cases: uncommon occlusions and strong rotations. As most occlusion to anime eyes is hair and the vast majority of are drawn up-right, the network can generate artifacts outside of those expected conditions. As shown in Fig.~\ref{fig:limitations}, the shape of occluders below the eyeline might be distorted as if was a skin tattoo. Eyes will be drawn upright if a character is reversed. Adding a network capable of estimating eye rotation to the pipeline would automate this process and further improve the generated dataset. But outside of these two unusual spaces, artifacts we did find were created by Poisson-blending, not our models, which leads us to conclude post-processing is the current main limitation of the method.

\fullsource{
    \begin{figure*}
      \setlength{\tabcolsep}{1pt}
      \footnotesize{\begin{tabular}{lcccccc}
       & \textit{Dragon Maid} \cite{dragonmaid} & \textit{Magus Bride} \cite{magusbride} & \textit{Franxx} \cite{franxx} & \textit{Xenoglossia} \cite{xenoglossia} & \textit{Magus Bride} \cite{magusbride} & \textit{Eden of the East} \cite{eden2009} \\
        \raisebox{-.75cm}{\rotatebox{90}{Original}}
            & \includegraphics[valign=c, width=0.157\textwidth]{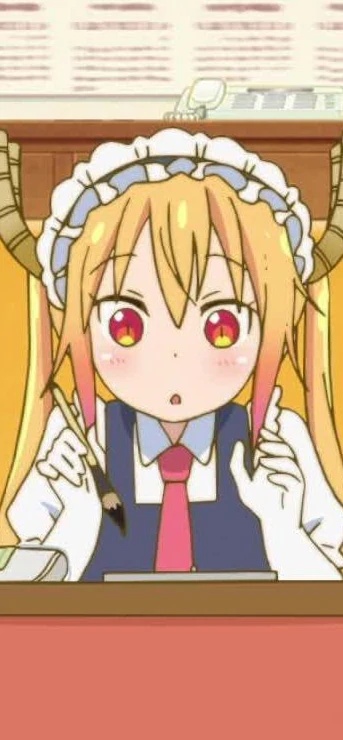}
            & \includegraphics[valign=c, width=0.157\textwidth]{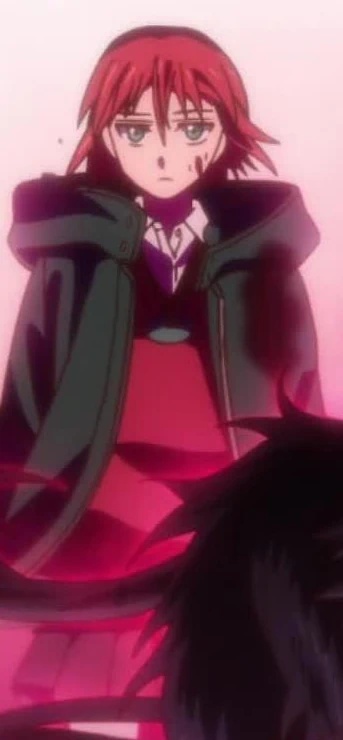}
            & \includegraphics[valign=c, width=0.157\textwidth]{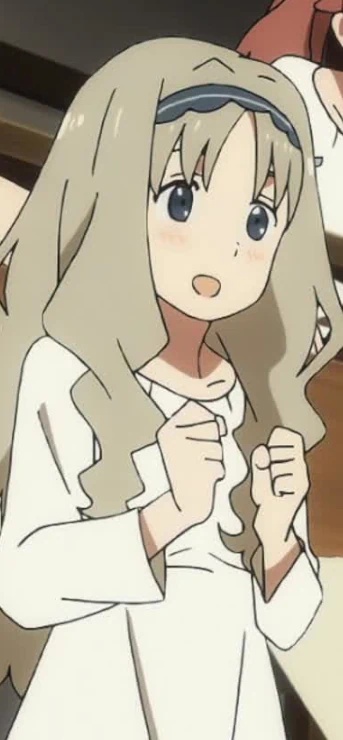}
            & \includegraphics[valign=c, width=0.157\textwidth]{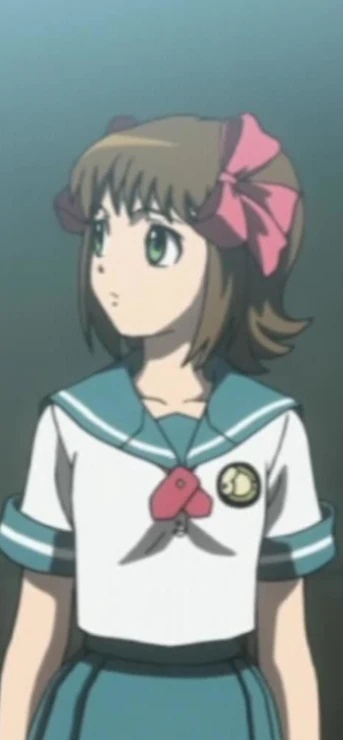}
            & \includegraphics[valign=c, width=0.157\textwidth]{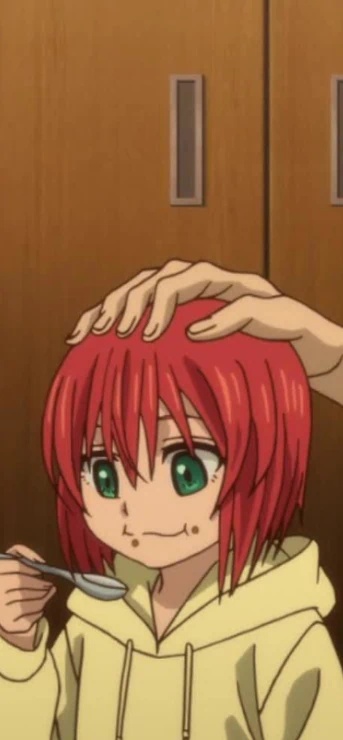}
            & \includegraphics[valign=c, width=0.157\textwidth]{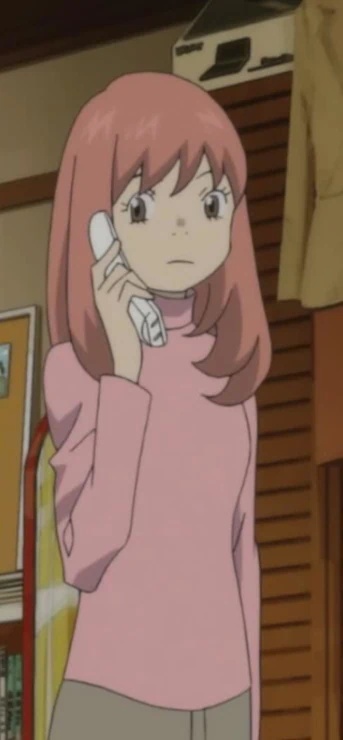} \\
        \raisebox{-.75cm}{\rotatebox{90}{Ours}}
            & \includegraphics[valign=c, width=0.157\textwidth]{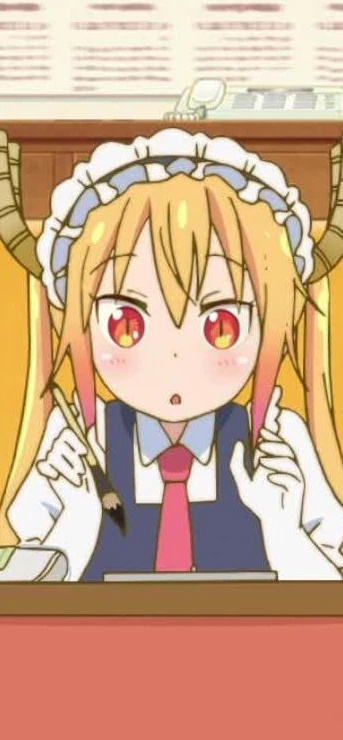}
            & \includegraphics[valign=c, width=0.157\textwidth]{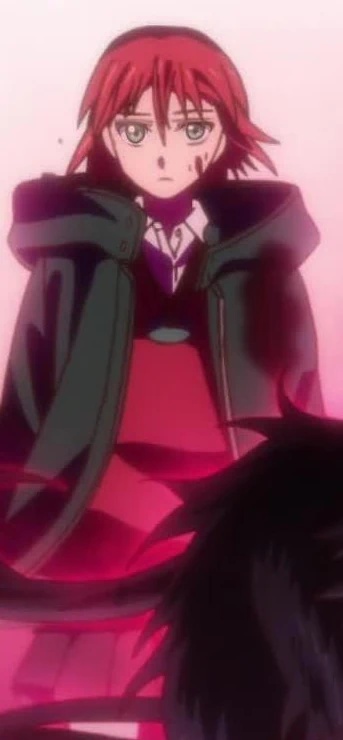}
            & \includegraphics[valign=c, width=0.157\textwidth]{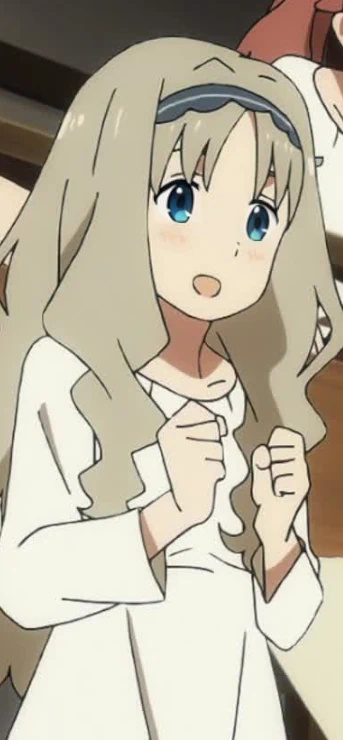}
            & \includegraphics[valign=c, width=0.157\textwidth]{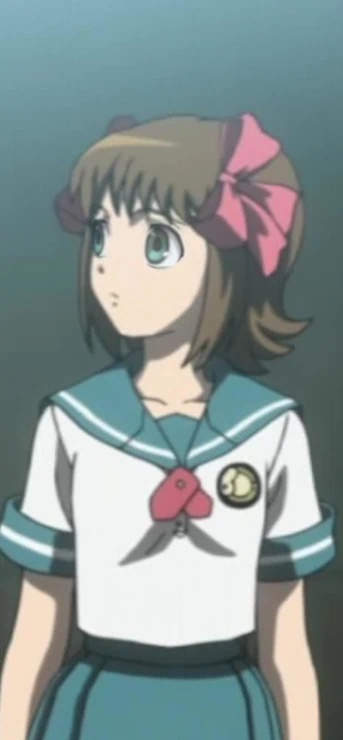}
            & \includegraphics[valign=c, width=0.157\textwidth]{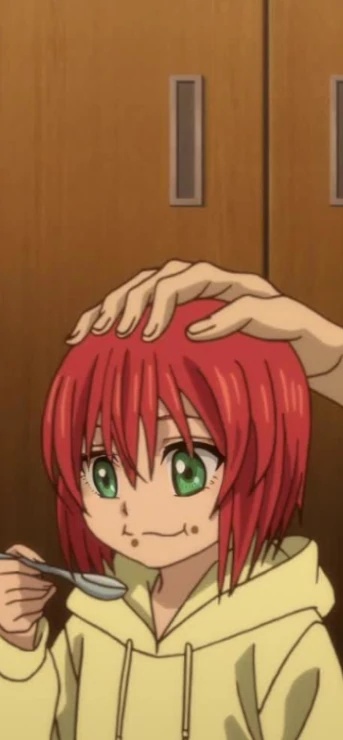}
            & \includegraphics[valign=c, width=0.157\textwidth]{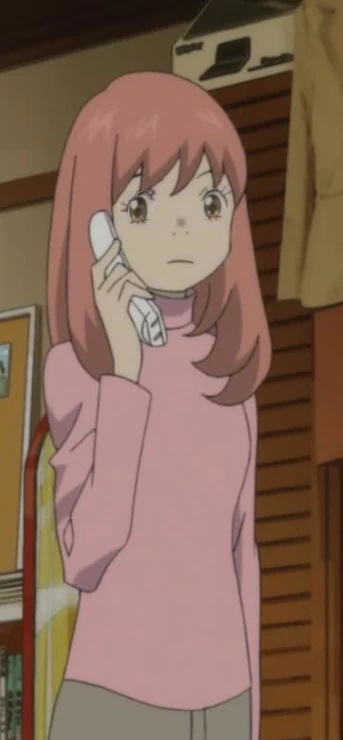}
      \end{tabular}}
      \caption[Increasing Detail]{\textbf{Increasing Detail}. The method behaves as intended when prompted to increase the amount of detail in character eyes, despite challenges such as unusual face proportions, adverse color balance due to lighting effects, head tilts, minor hair occlusions and character designs from productions that have not been seen during training.}
      \label{detail}
    \end{figure*}
    
    \begin{figure*}
      \setlength{\tabcolsep}{1pt}
      \footnotesize{\begin{tabular}{lcccccc}
         & Original & & $\longleftarrow$  & Redesigned & $\longrightarrow$ & \\
        \raisebox{-.75cm}{\rotatebox{90}{\textit{Dragon Maid} \cite{dragonmaid}}}
            & \includegraphics[valign=c, width=0.157\textwidth]{img/results/G.jpg}
            & \includegraphics[valign=c, width=0.157\textwidth]{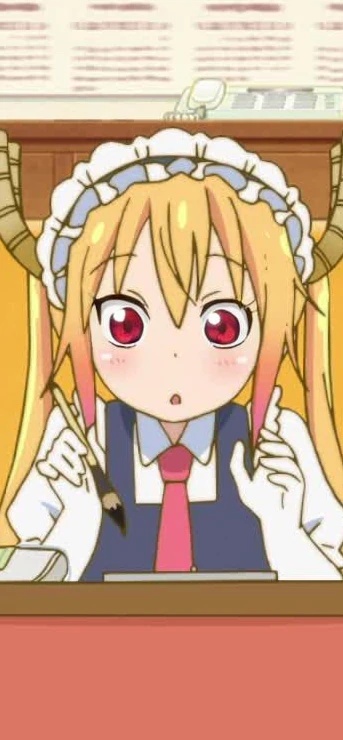}
            & \includegraphics[valign=c, width=0.157\textwidth]{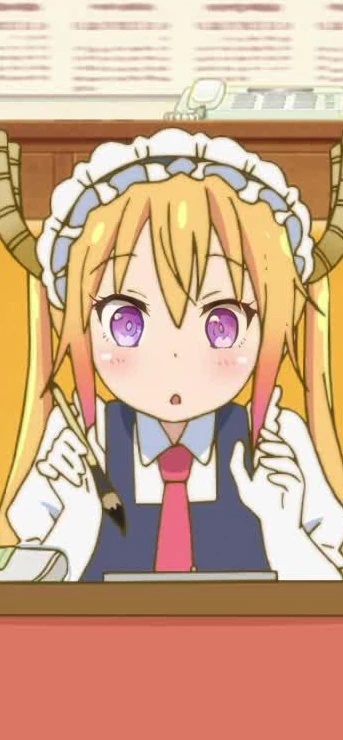}
            & \includegraphics[valign=c, width=0.157\textwidth]{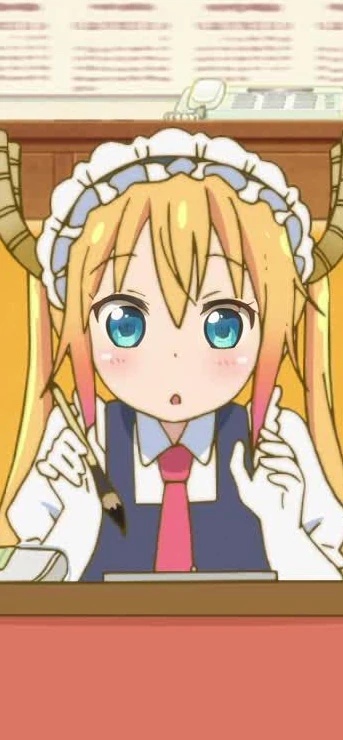}
            & \includegraphics[valign=c, width=0.157\textwidth]{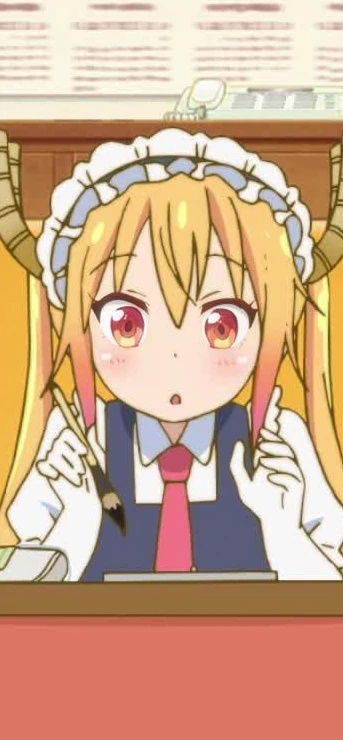}
            & \includegraphics[valign=c, width=0.157\textwidth]{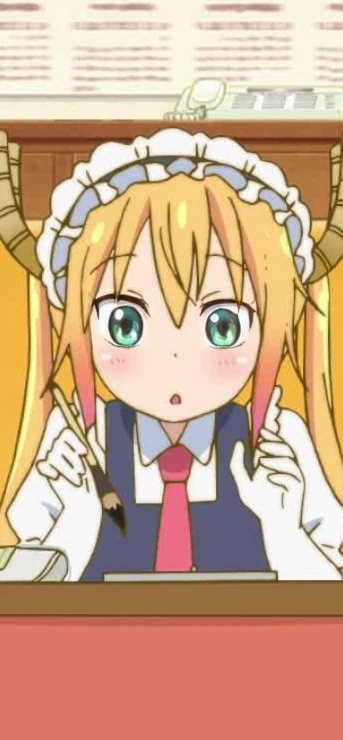} \\
        \raisebox{-.75cm}{\rotatebox{90}{\textit{Re:Zero} \cite{re-zero}}}
            & \includegraphics[valign=c, width=0.157\textwidth]{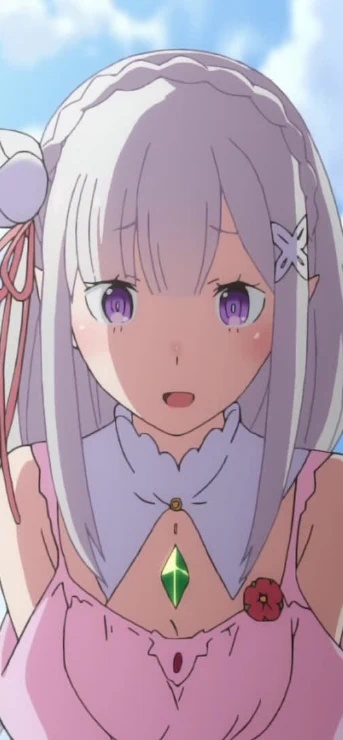}
            & \includegraphics[valign=c, width=0.157\textwidth]{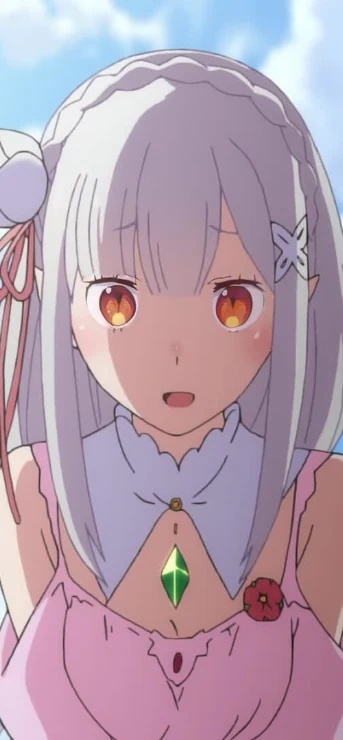}
            & \includegraphics[valign=c, width=0.157\textwidth]{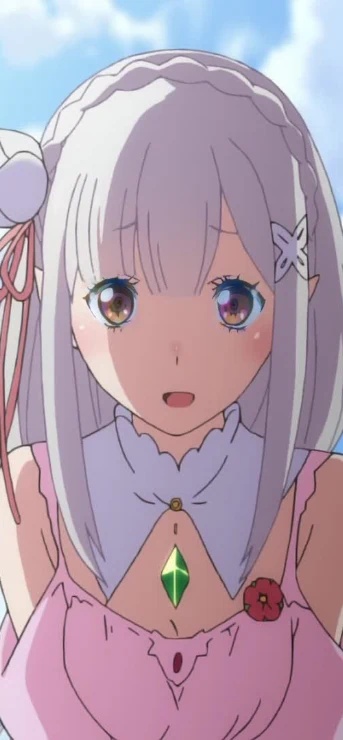}
            & \includegraphics[valign=c, width=0.157\textwidth]{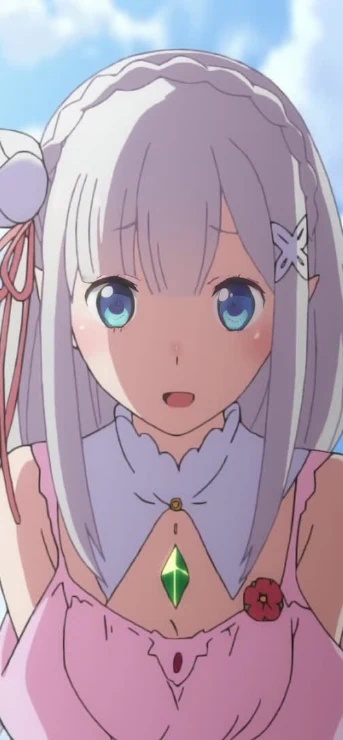}
            & \includegraphics[valign=c, width=0.157\textwidth]{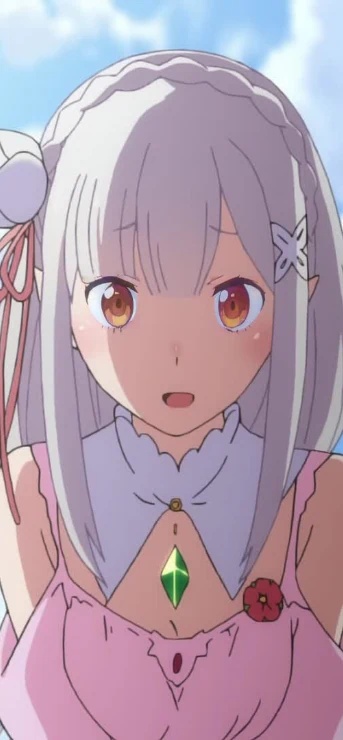}
            & \includegraphics[valign=c, width=0.157\textwidth]{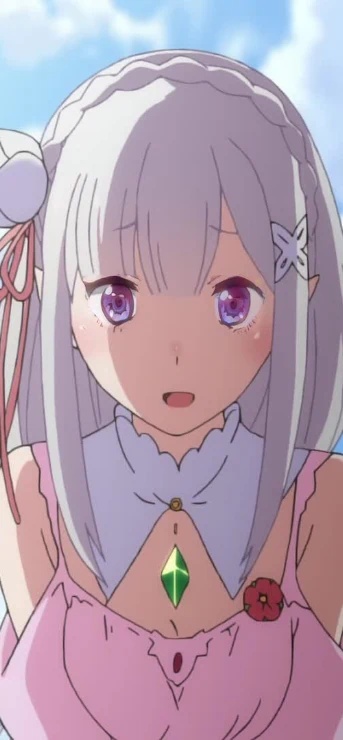} \\
        \raisebox{-.75cm}{\rotatebox{90}{\textit{Magus Bride} \cite{magusbride}}}
            & \includegraphics[valign=c, width=0.157\textwidth]{img/results/E.jpg}
            & \includegraphics[valign=c, width=0.157\textwidth]{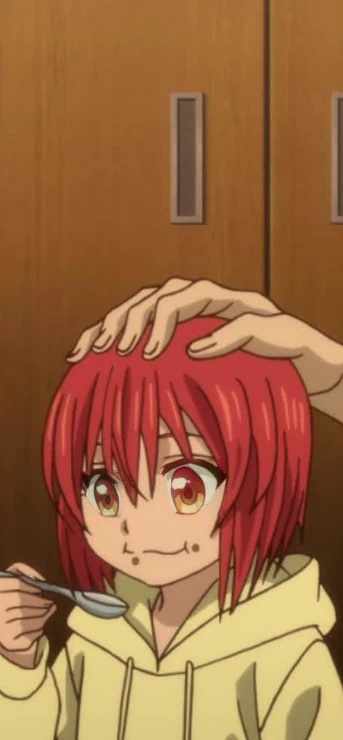}
            & \includegraphics[valign=c, width=0.157\textwidth]{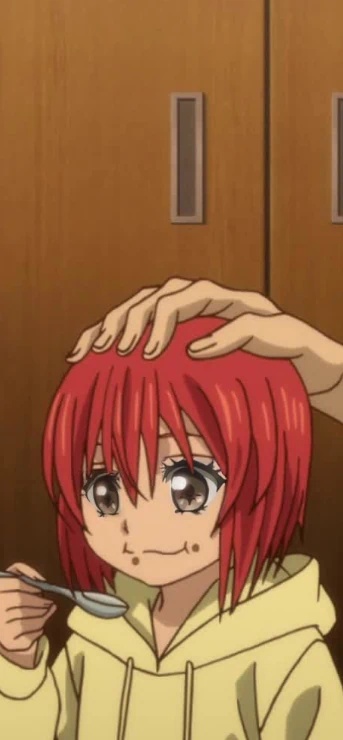}
            & \includegraphics[valign=c, width=0.157\textwidth]{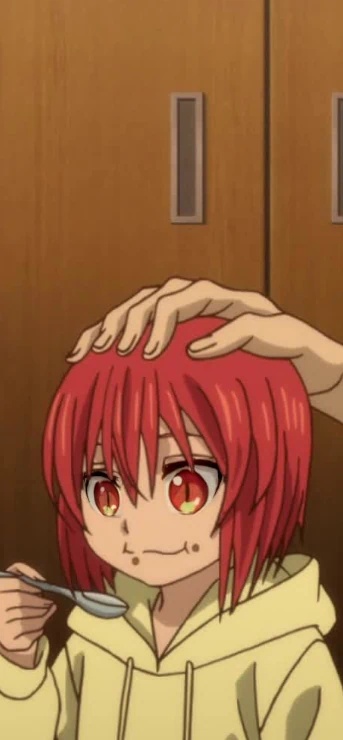}
            & \includegraphics[valign=c, width=0.157\textwidth]{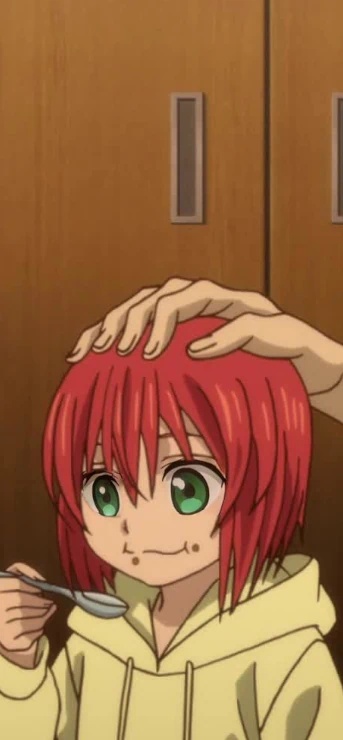}
            & \includegraphics[valign=c, width=0.157\textwidth]{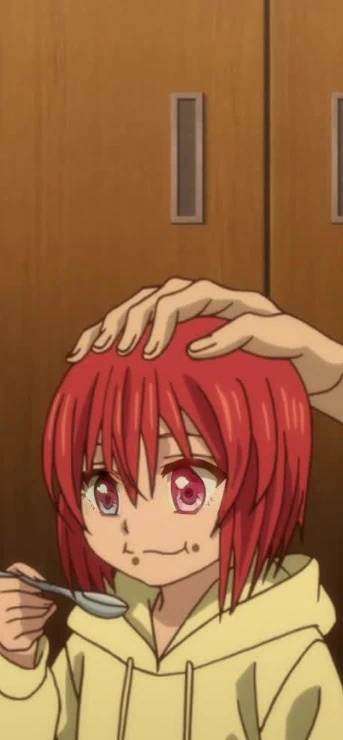}
      \end{tabular}}
      \caption[Redesign Results]{\textbf{Redesign Results}. A side-effect of how our proposed networks are trained is that our method is also capable of applying entirely different designs to characters. While not the focus of our work, it demonstrates the versatility of our method and the robustness of the learned model, even when applied to high-resolution imagery.}
      \label{redesign1}
    \end{figure*}
    
    \begin{figure*}
      \setlength{\tabcolsep}{1pt}
      \footnotesize{\begin{tabular}{lcccccc}
       & Original & & $\longleftarrow$ & Redesigned & $\longrightarrow$ & \\
        \raisebox{-.75cm}{\rotatebox{90}{\textit{Angels of Death} \cite{angelsofdeath}}}
            & \includegraphics[valign=c, width=0.157\textwidth]{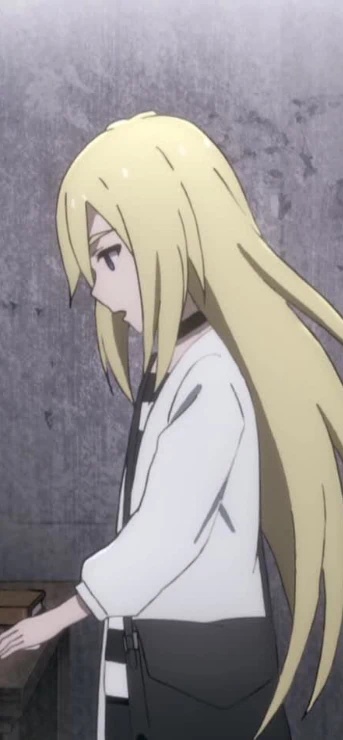}
            & \includegraphics[valign=c, width=0.157\textwidth]{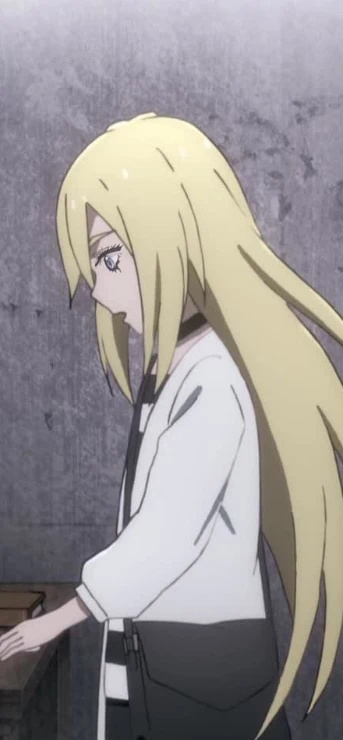}
            & \includegraphics[valign=c, width=0.157\textwidth]{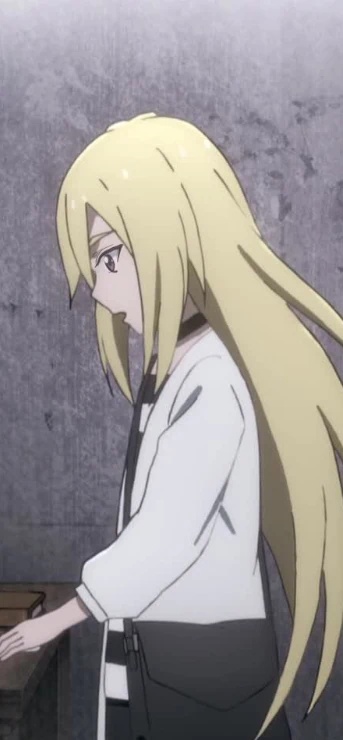}
            & \includegraphics[valign=c, width=0.157\textwidth]{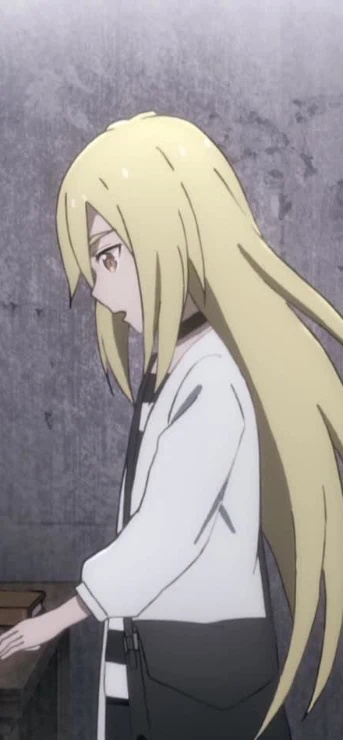}
            & \includegraphics[valign=c, width=0.157\textwidth]{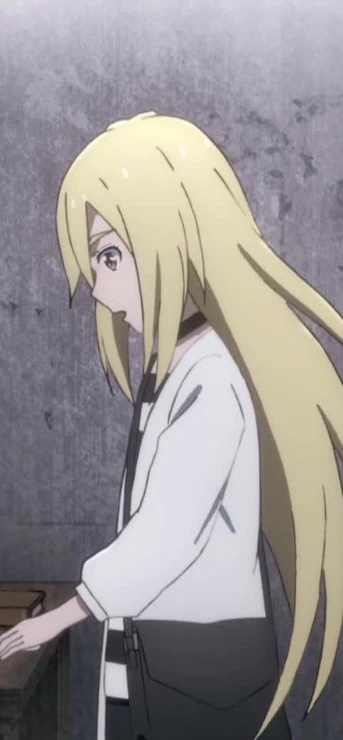}
            & \includegraphics[valign=c, width=0.157\textwidth]{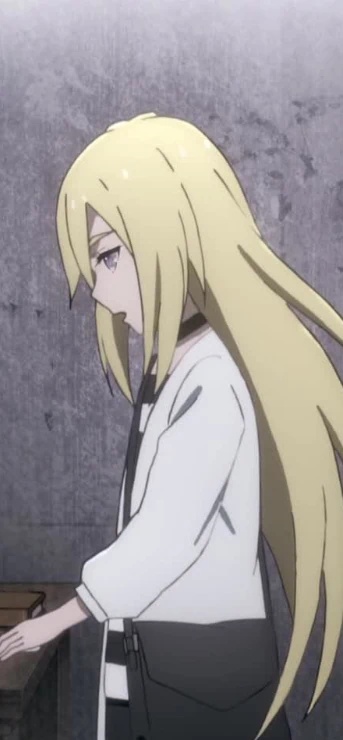} \\
        \raisebox{-.75cm}{\rotatebox{90}{\textit{Magus Bride} \cite{magusbride}}}
            & \includegraphics[valign=c, width=0.157\textwidth]{img/results/I.jpg}
            & \includegraphics[valign=c, width=0.157\textwidth]{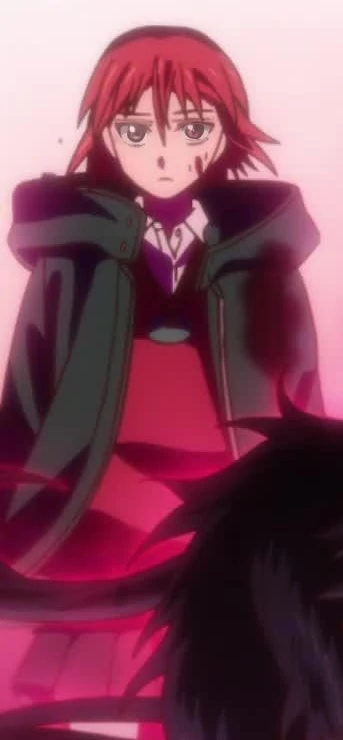}
            & \includegraphics[valign=c, width=0.157\textwidth]{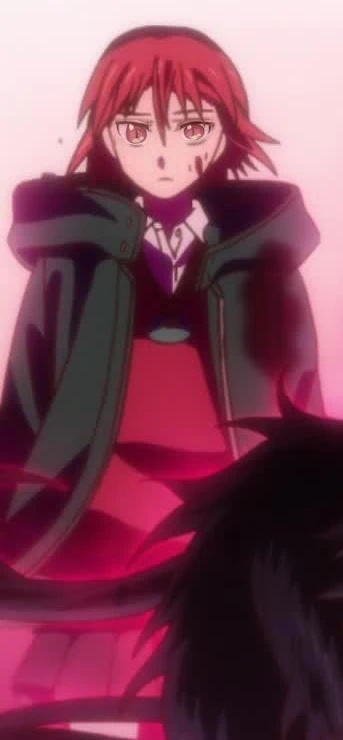}
            & \includegraphics[valign=c, width=0.157\textwidth]{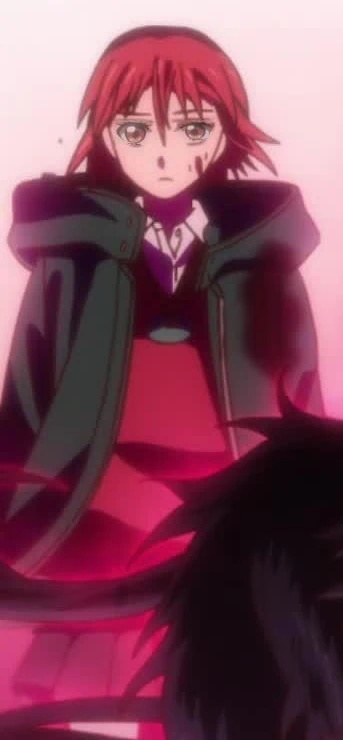}
            & \includegraphics[valign=c, width=0.157\textwidth]{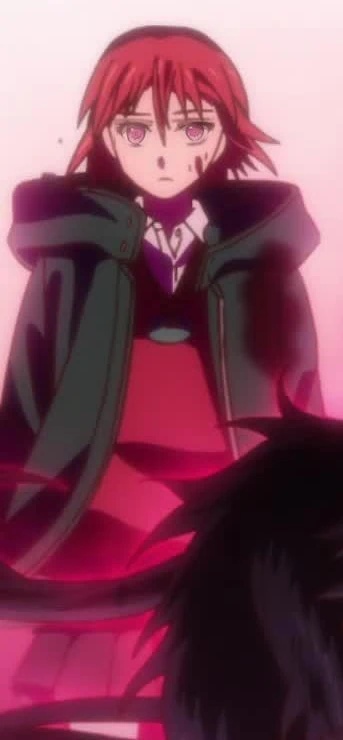}
            & \includegraphics[valign=c, width=0.157\textwidth]{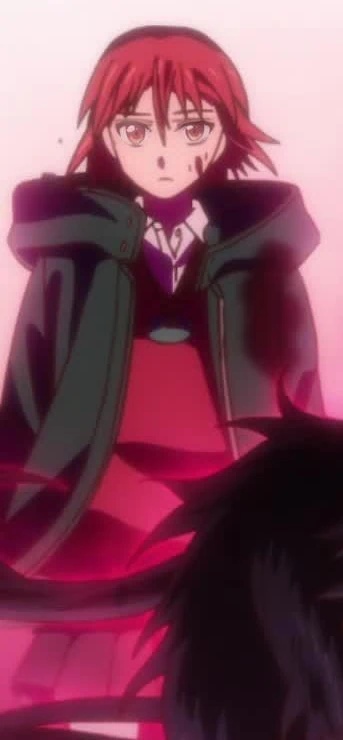} \\
        \raisebox{-.75cm}{\rotatebox{90}{\textit{Dr.Stone} \cite{dr-stone}}}
            & \includegraphics[valign=c, width=0.157\textwidth]{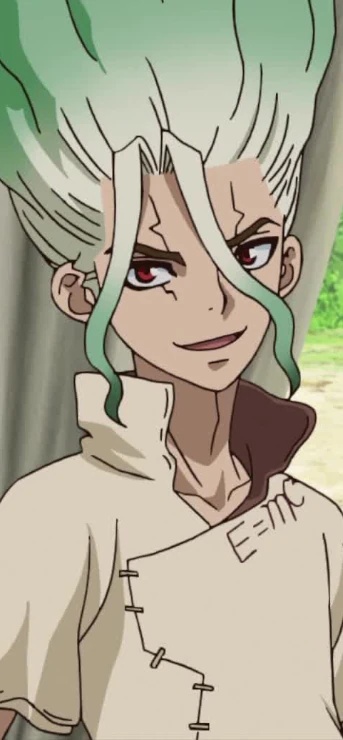}
            & \includegraphics[valign=c, width=0.157\textwidth]{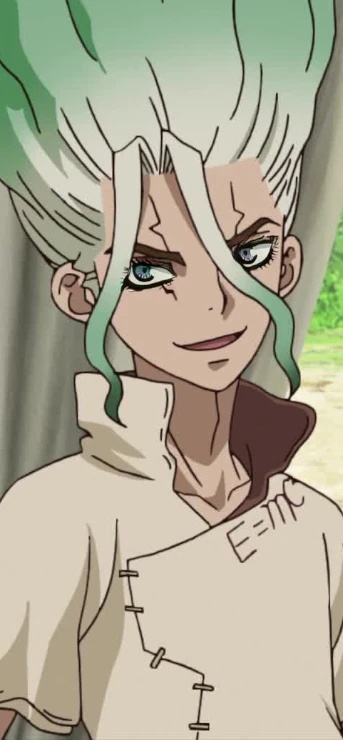}
            & \includegraphics[valign=c, width=0.157\textwidth]{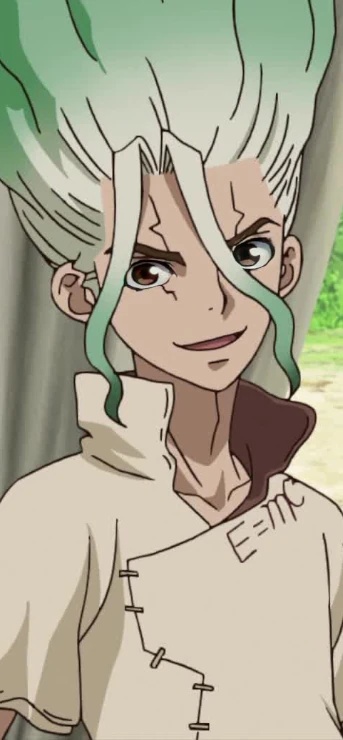}
            & \includegraphics[valign=c, width=0.157\textwidth]{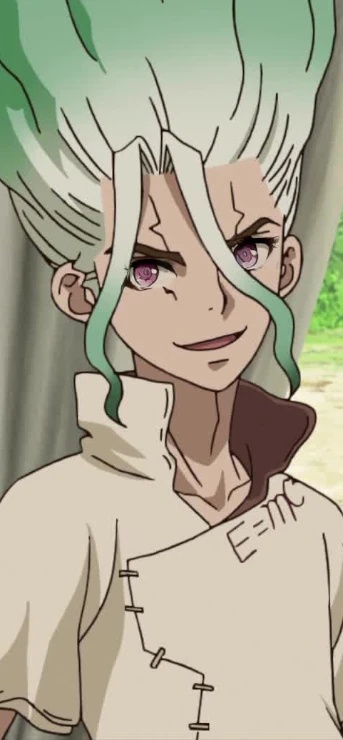}
            & \includegraphics[valign=c, width=0.157\textwidth]{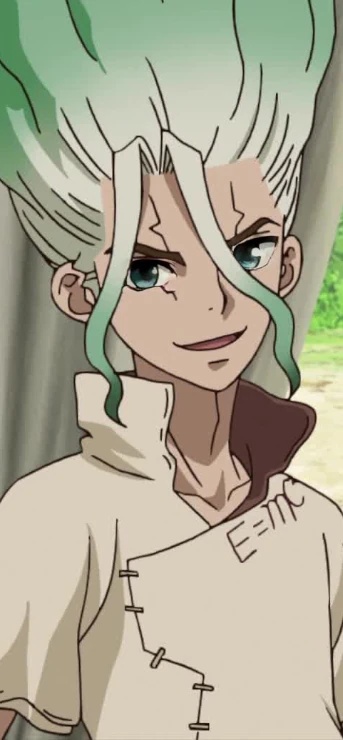}
            & \includegraphics[valign=c, width=0.157\textwidth]{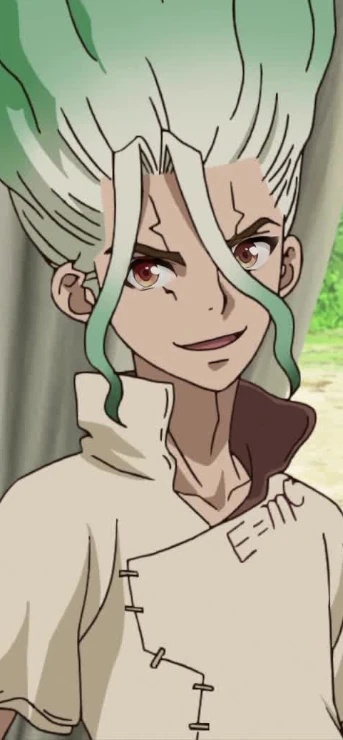} \\
      \end{tabular}}
      \caption[Redesign Results]{\textbf{Redesign Results}. Further examples of character redesign which demonstrate that our method is robust enough to handle characters in a variety of challenging scenarios. This includes dealing with oblique camera angles, irregular lighting conditions, head tilts and occlusions caused by hair.}
      \label{redesign2}
    \end{figure*}
    
    \begin{figure*}
        \centering
        \begin{subfigure}{0.49\textwidth}
            \centering
            \includegraphics[width=\textwidth]{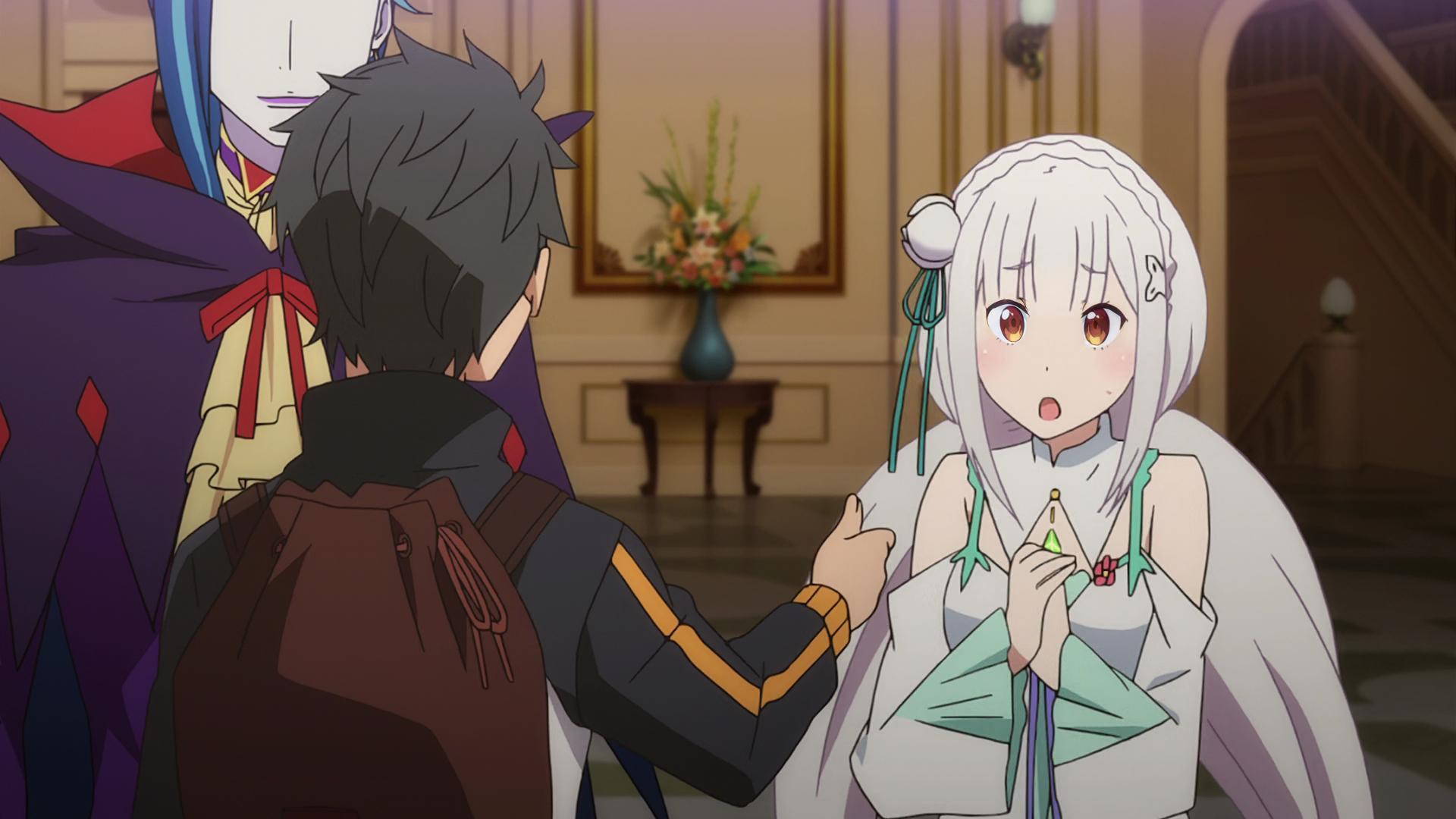}
        \end{subfigure}
        \begin{subfigure}{0.49\textwidth}
            \centering
            \includegraphics[width=\textwidth]{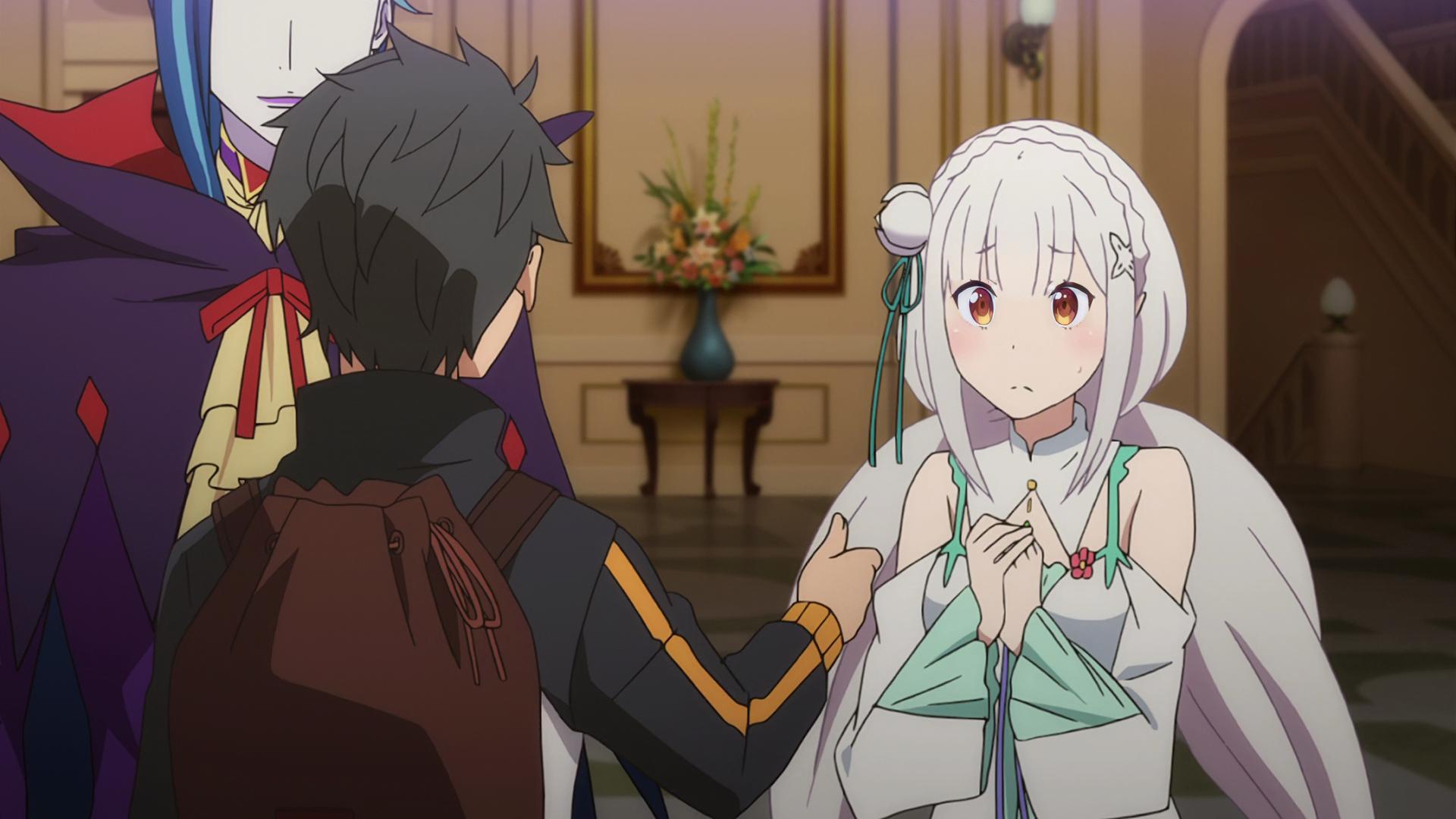}
        \end{subfigure}
        \begin{subfigure}{0.49\textwidth}
            \centering
            \includegraphics[width=\textwidth]{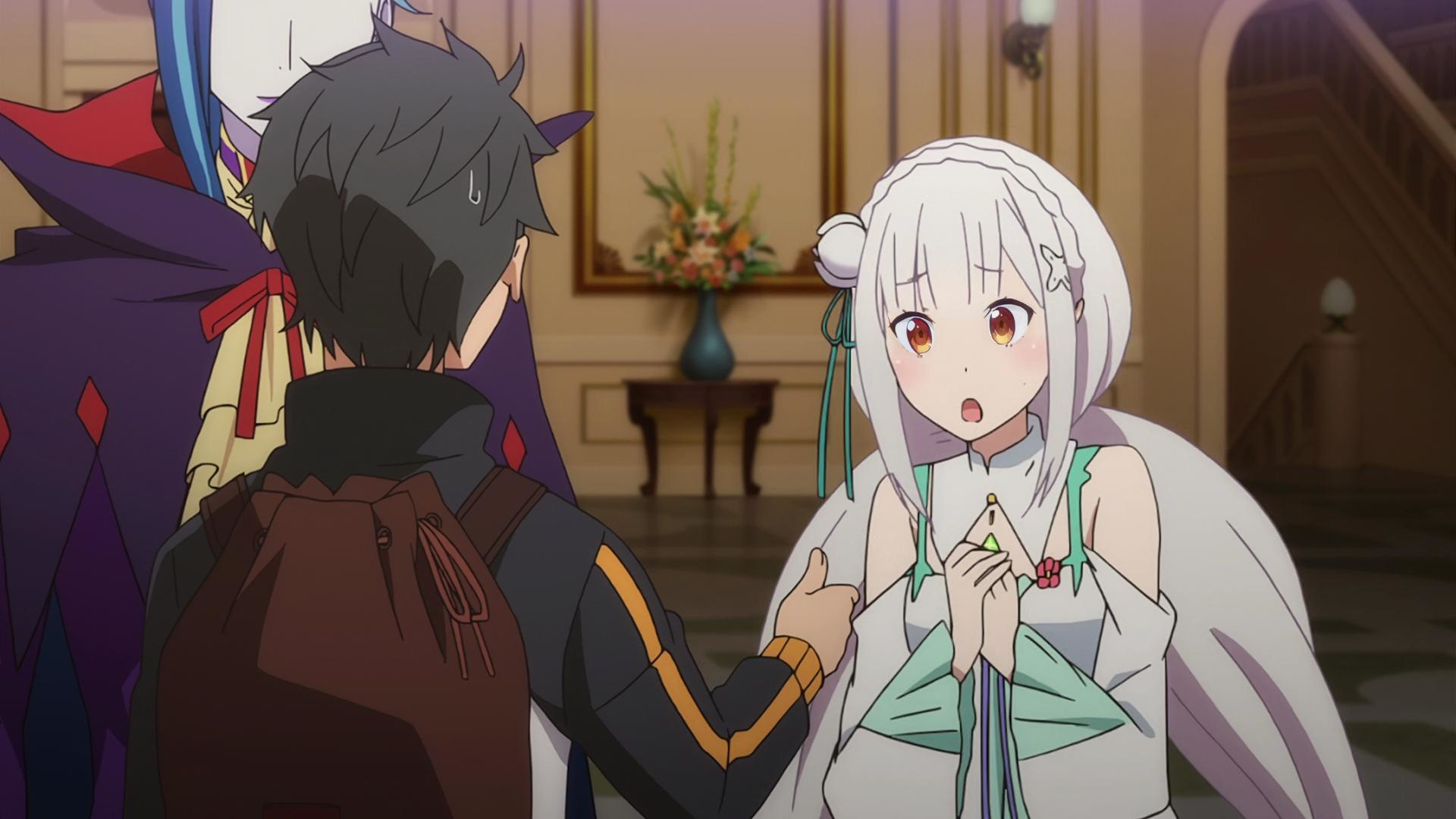}
        \end{subfigure}
        \begin{subfigure}{0.49\textwidth}
            \centering
            \includegraphics[width=\textwidth]{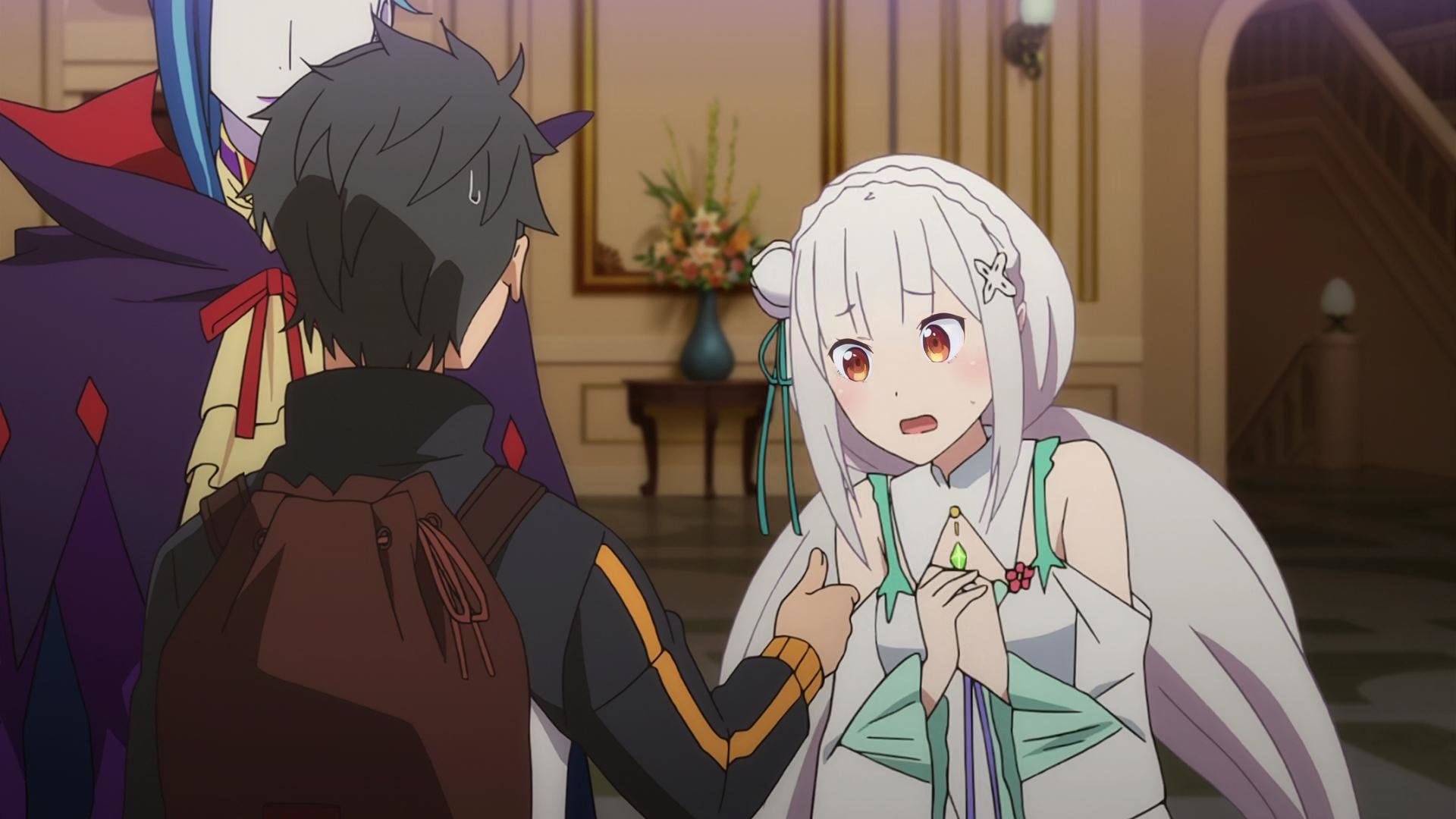}
        \end{subfigure}
        \begin{subfigure}{0.49\textwidth}
            \centering
            \includegraphics[width=\textwidth]{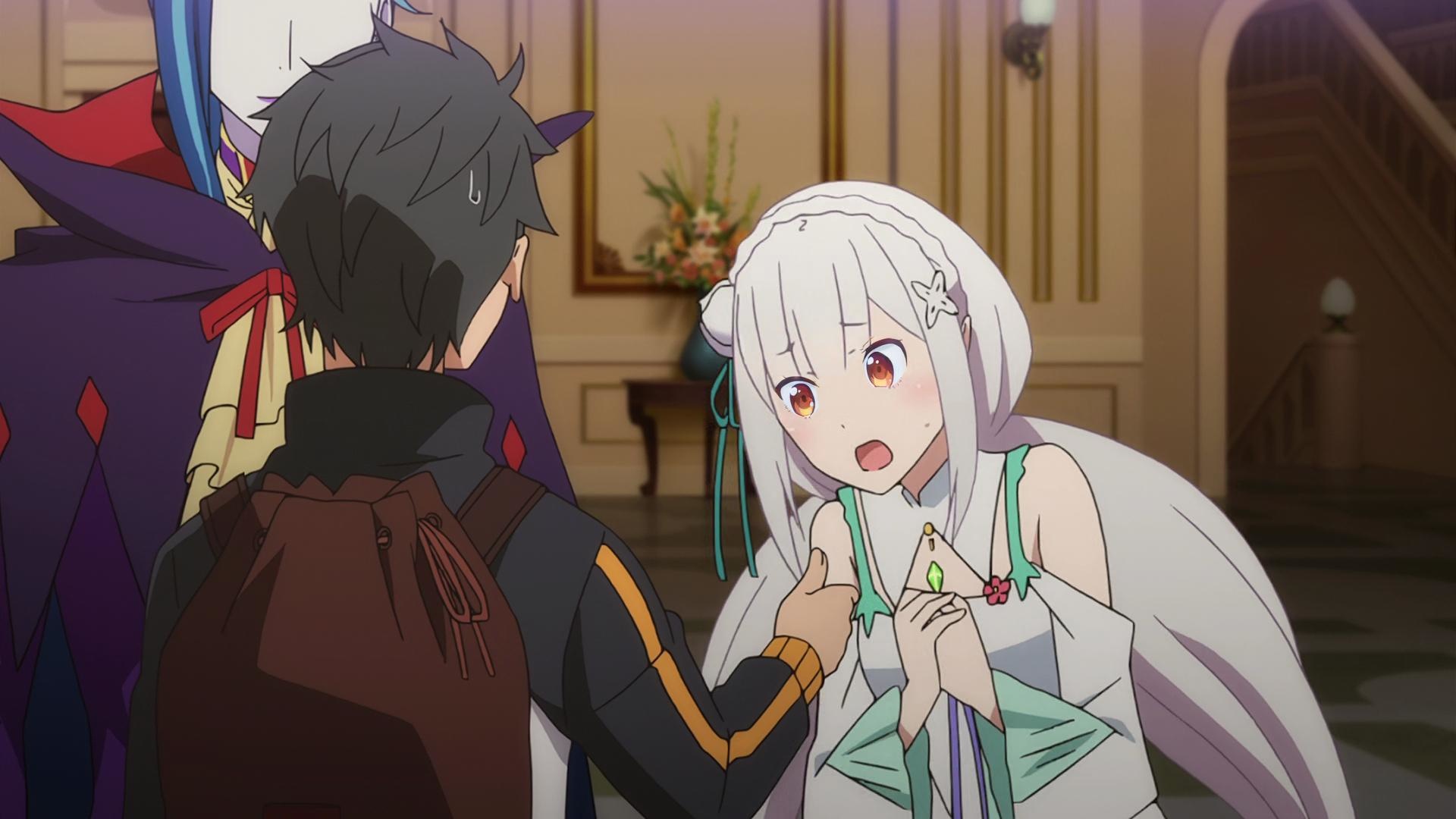}
        \end{subfigure}
        \begin{subfigure}{0.49\textwidth}
            \centering
            \includegraphics[width=\textwidth]{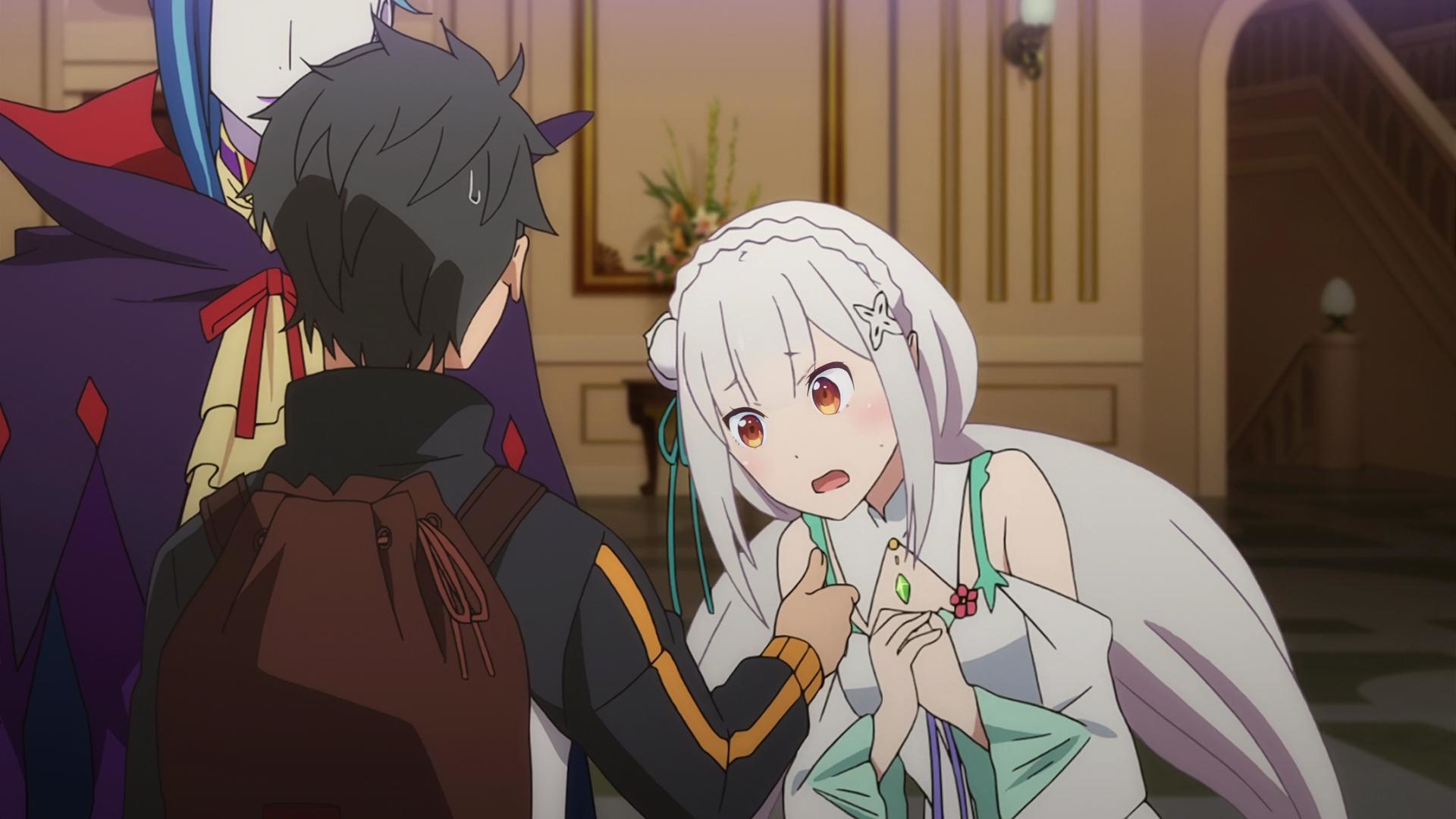}
        \end{subfigure}
        \begin{subfigure}{0.49\textwidth}
            \centering
            \includegraphics[width=\textwidth]{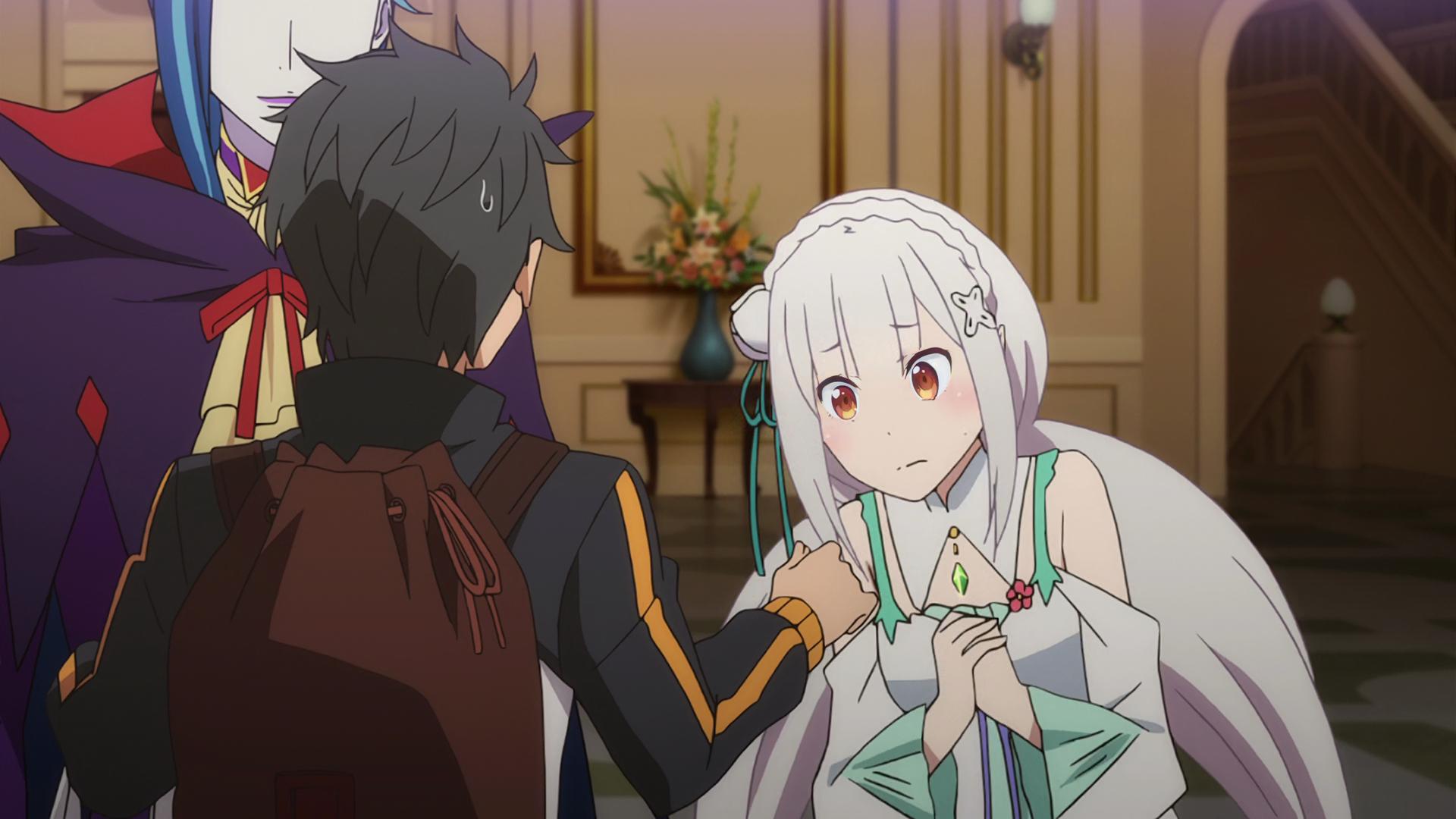}
        \end{subfigure}
        \begin{subfigure}{0.49\textwidth}
            \centering
            \includegraphics[width=\textwidth]{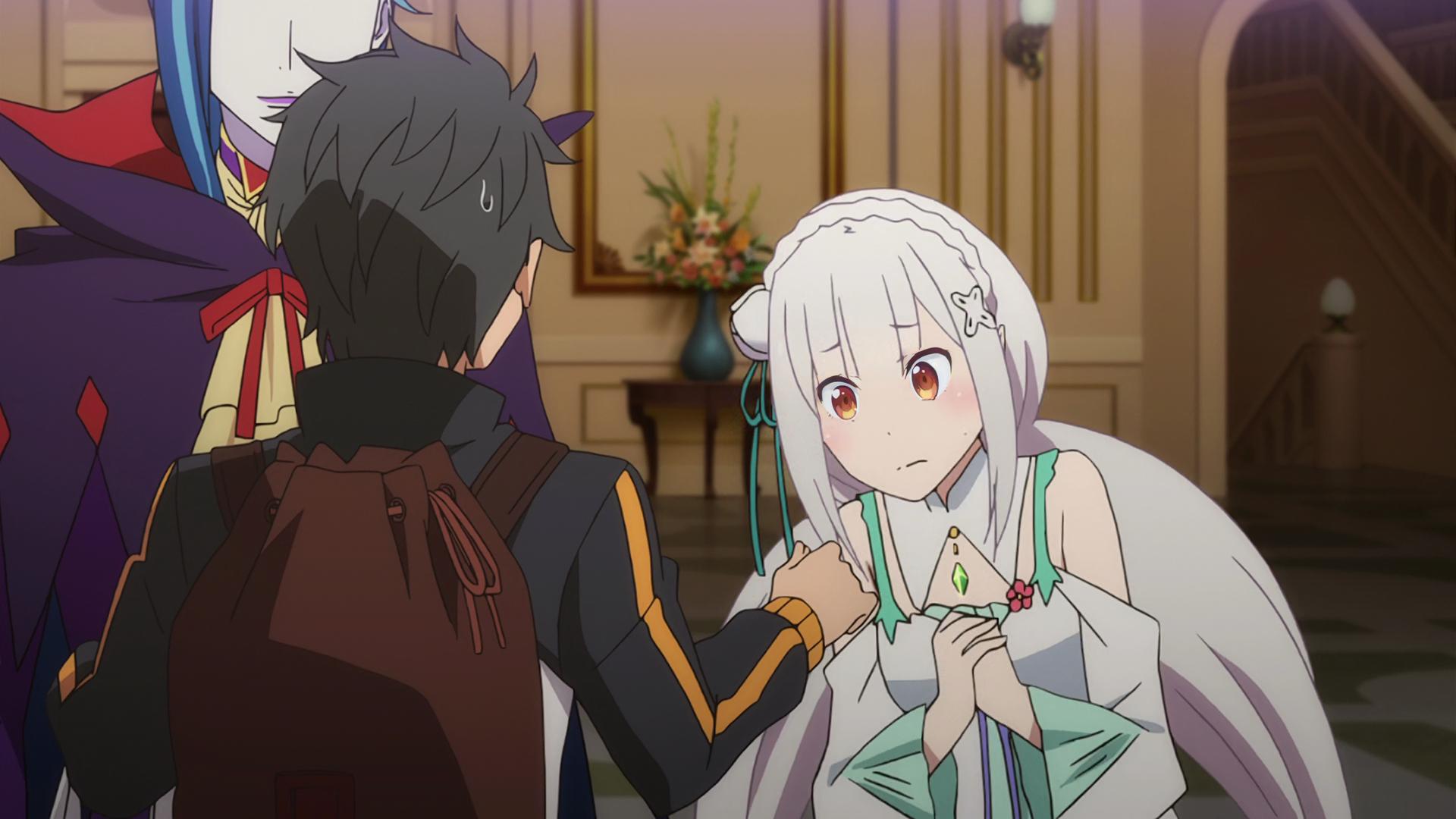}
        \end{subfigure}
        \caption[Temporal Coherence]{\textbf{Temporal Coherence}. Example of redesigning the eyes of a character from \textit{Re:Zero} \cite{re-zero} in a particularly lengthy shot. Output is temporally consistent.}
        \label{temporal-coherence}
    \end{figure*}
}

\opensource{
    \begin{figure*}
      \setlength{\tabcolsep}{1pt}
      \footnotesize{\begin{tabular}{cccccc}
        Original & {} & $\longleftarrow$  & Redesigned & $\longrightarrow$ & {} \\
        \includegraphics[valign=c, width=0.157\textwidth]{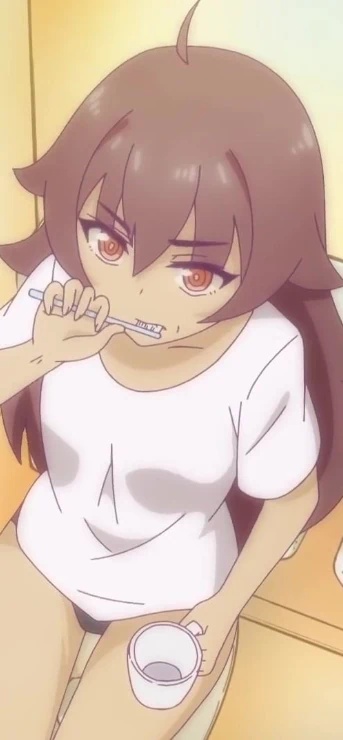} &
        \includegraphics[valign=c, width=0.157\textwidth]{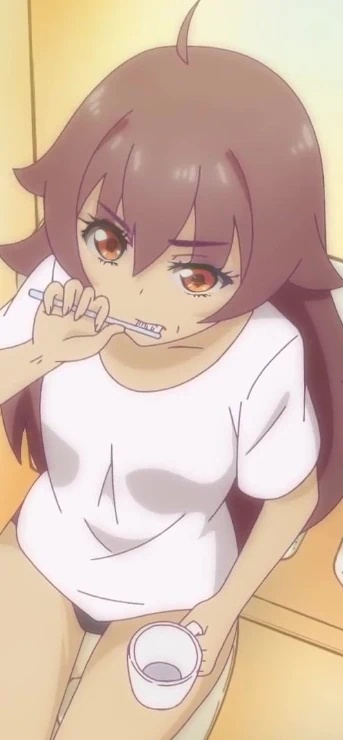} &
        \includegraphics[valign=c, width=0.157\textwidth]{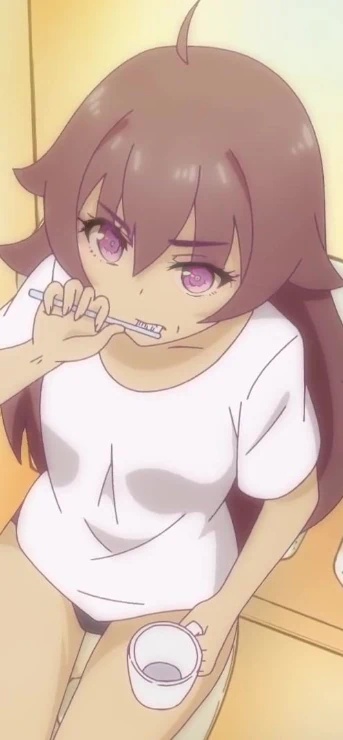} &
        \includegraphics[valign=c, width=0.157\textwidth]{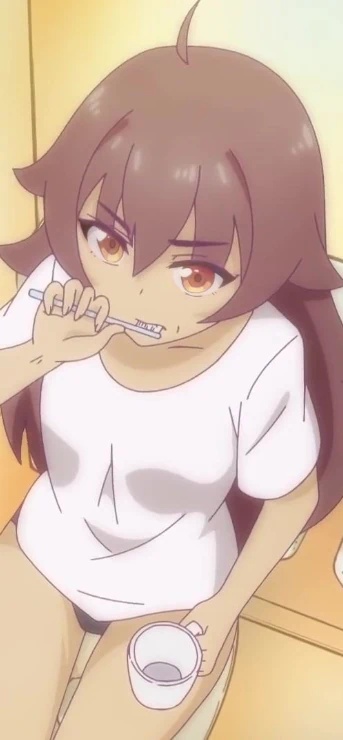} &
        \includegraphics[valign=c, width=0.157\textwidth]{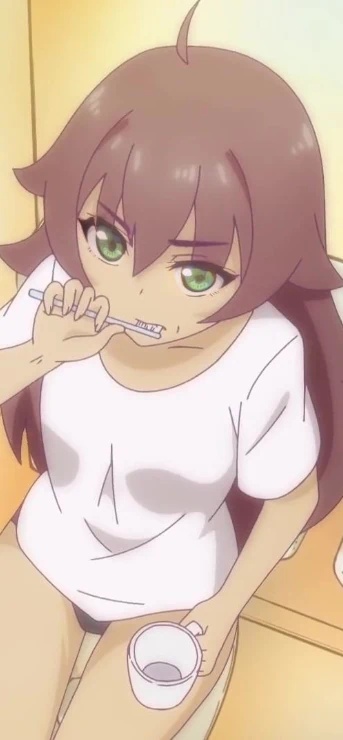} &
        \includegraphics[valign=c, width=0.157\textwidth]{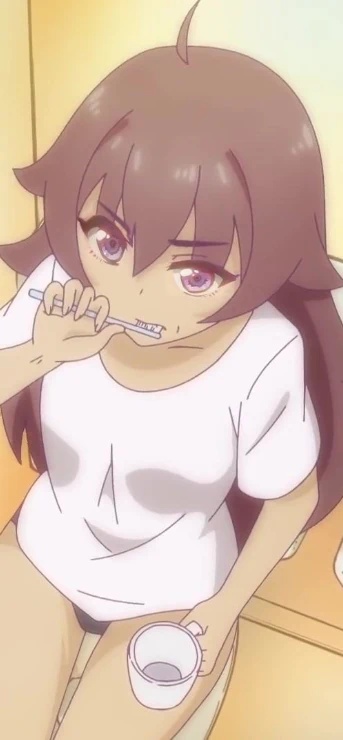}
      \end{tabular}}
      \caption[Redesign Results]{\textbf{Redesign Results}. Despite not being the intended use case we wanted to solve, a side-effect of how our networks are trained is that our method is also capable of applying entirely different designs to characters. This demonstrates the versatility of our method and the robustness of the learned models.}
      \label{redesign-otachan}
 \end{figure*}
}

    \section{User Study}\label{user-study}
\noindent To validate our work, we conducted a user study with 63 participants and three different tasks with images coming from an anime production. Participants applied voluntarily to the study and the majority of them classified themselves as anime experts. Three visual tasks were presented:

\paragraph*{Realness (T1)}
The user had to watch eight images and pick the ones with drawing problems. In this test, four were original un-retouched images ({\small \em Market, Tent, Cold Night, Field of Leaves}), and four had the eyes replaced by our approach ({\small \em Dormitory, Bar, Bathroom, Crying}). Images were presented in random order, and the user did not know how many could have problems. This study aimed to determine if our method generates visual artifacts or if its result is not distinguishable from a production.
To analyze these results, we used the $\chi^2$ test with Yates correction \cite{SIEGEL88} and null hypothesis $H_0$:\emph{``Original and retouched images are equally probable to be considered as wrong''}. The number of times users decided that an image was wrong has a very similar distribution for images generated by our approach and original ones (see Tabs.~\ref{fig:T1-barchart} and ~\ref{tab:realness}). $\chi^2(1,N\eq504)\eq0.0405, p\eq0.840$, therefore, with a $p$\-value larger than 0.05, we can conclude  independence between groups (\emph{Our vs Real}) and categories (\emph{Wrong vs Right}), or in simpler words, our generated eyes do not have evident artifacts and are indistinguishable from real production.

\paragraph*{Level of Detail (T2)}
Participants were asked to choose between two images: one unedited (U) and one with eyes enhanced by our approach (O), focusing on which image displayed more detailed eyes.   Each user evaluated eight pairs of images. This study aimed to determine if we effectively produce images with higher detail than the original production.
To understand if the expressed preferences were statistically significant, we employed the formula for the multiple comparison test \cite{DAVID88}: if difference between the score of one method against another exceeded a critical value $R$, the results were deemed significant. $R$ was calculated according to specified alpha significance level of $0.05$. We also tested if the agreement among observers $u \in [-1,1]$ (where $u\eq-1$ strong disagreement and $u\eq1$  strong agreement) with the following null hypothesis $H_0$: \emph{``There is no agreement among participants on one category being better''}. The failure of this hypothesis implies that categories are perceptually equivalent causing difficulties in judging (e.g. we check if users alloted their preferences at random).
The overall result (Table~\ref{tab:lod}) is that our method effectively produces images with higher detail than the original production artwork with statistical significance.

\paragraph*{Preference (T3)}
The user had to select the better looking image from a pair. For each of the 8 scenes, there were three pairs cross-comparing 3 methods: FUNIT\cite{liu2019few} (F), FUNIT with our clustering (FC), and our method (O); in total, 24 pairs for each user. This study aims to determine if our method is more effective in producing results better looking than F and FC. In short, our method was preferred to (F) 95.16\% of times. In Tab.~\ref{tab:preference} we used the same analysis of \textbf{T2}, showing that images generated by our method are more preferred than the ones generated by the previous work (F), even when it is enhanced with our dataset (FC).

\begin{table}
    \caption[Domain Knowledge]{\textbf{Domain Knowledge}. More than half of the users classified themselves on the high end of an anime (viewer) knowledge spectrum.}
    \label{fig:userstudy-animeknowledge}
    
    \begin{tikzpicture}
      \tikzstyle{every node}=[font=\small]
      \begin{axis}[
        xbar stacked, xmin=0,
        y=6mm, width=1.1\linewidth,
        symbolic y coords={0-1,2-3,4-5,6-7,8-10}, ytick=data]
        
        \addplot [fill=blue!60] coordinates {(0,0-1) (0,2-3) (0,4-5) (0,6-7) (32,8-10)};
        \addplot [fill=cyan!60] coordinates {(0,0-1) (0,2-3) (0,4-5) (11,6-7) (0,8-10)};
        \addplot [fill=coolyellow!60] coordinates {(0,0-1) (0,2-3) (10,4-5) (0,6-7) (0,8-10)};
        \addplot [fill=orange!60] coordinates {(0,0-1) (8,2-3) (0,4-5) (0,6-7) (0,8-10)};
        \addplot [fill=red!60] coordinates {(1,0-1) (0,2-3) (0,4-5) (0,6-7) (0,8-10)};
      \end{axis}
    \end{tikzpicture}
\end{table}

\begin{table}
    \caption[Realness Test (T1)]{\textbf{Realness Test (T1)}: The number of times that our redrawn art frames $\color{blue} \blacksquare$ and original unmodified ones $\color{cyan} \blacksquare$ have been picked as fake is not significantly different.}
    \label{fig:T1-barchart}
    
    \pgfplotstableread{ 
        Scene           Ours    Original
        {Cold Night}    0       21
        {Leaf Field}    0       20 
        {Market}        0       14
        {Tent}          0       13
        {Bar}           10      0
        {Bathroom}      20      0 
        {Crying}        24      0 
        {Dormitory}     12      0
    }\testdata

    \centering
    \begin{tikzpicture}
    \tikzstyle{every node}=[font=\small]
    \begin{axis}[
        xbar stacked,
        y=6mm, width=0.9\linewidth,
        ytick=data, yticklabels from table={\testdata}{Scene}]
        
        \addplot [fill=blue!60] table [x=Ours, meta=Scene, y expr=\coordindex] {\testdata};
        \addplot [fill=cyan!60] table [x=Original, meta=Scene, y expr=\coordindex] {\testdata};
      \end{axis}
    \end{tikzpicture}
\end{table}

\begin{table}
    \caption[Realness Test (T1)]{\textbf{Realness Test (T1)}. We tested the independence between groups (Our vs Real) and categories (Wrong vs Right) and our generated eyes were indistinguishable from real artwork.}
    \label{tab:realness}
    
    \centering
    \begin{tabular}{|c|c|c||c|}
        \hline
             & \textbf{Wrong} & \textbf{Right} & \textbf{Marginal Row Totals}\\
        \hline
        \textbf{Ours}  & 66 & 186 & 252\\ 
        \hline
        \textbf{Real}  & 69 & 183 & 252\\ 
        \hline
    \end{tabular}
\end{table}

\begin{table}
    \caption[Level-of-Detail Test (T2)]{\textbf{Level-of-Detail Test (T2)}. U and O refer to unedited and our generated images, respectively. For this test, the minimum score difference $R$ to determine statistical difference is $R=45$ for the overall comparison and $R=16$ for each single image \cite{DAVID88}, with results not significantly different being circled together. We also performed the $\chi^2$ test with Yates correction \cite{SIEGEL88}, setting a lower bound of $\chi^2=3.84$ for this test.}
    \label{tab:lod}
    
    \centering
    \resizebox{\columnwidth}{!}{
        \begin{tabular}{|c|c|c|c|c|}
            \hline
            Image &  Agree. Coeff. $u$ & $\chi^2$ & Sign. $u$ & Groups\\
            \hline
            1 & 0.230 & 15.25 & OK & \mymkg{U - 16} \mymkr{O - 47}\\
            2 & 0.041 &  3.57 & NO & \mymkr{U - 24; O - 39}\\
            3 & 0.119 &  8.40 & OK & \mymkg{U - 20} \mymkr{O - 43}\\
            4 & 0.334 & 21.73 & OK & \mymkg{U - 13} \mymkr{O - 50}\\
            5 & 0.373 & 24.14 & OK & \mymkg{U - 12} \mymkr{O - 51}\\
            6 & 0.297 & 19.44 & OK & \mymkg{U - 14} \mymkr{O - 49}\\
            7 & 0.058 &  4.59 & OK & \mymkg{U - 23} \mymkr{O - 40}\\
            8 & 0.144 &  9.92 & OK & \mymkg{U - 19} \mymkr{O - 44}\\
            \hline
            Overall & 0.192 & 97.79 & OK & \mymkg{U - 141} \mymkr{O - 363}\\
            \hline
        \end{tabular}
    }
\end{table}

\begin{table}
    \caption[Preference Test (T3)]{\textbf{Preference Test (T3)}. F, FC and O refer to images generated by FUNIT, FUNIT with our clustering and our method, respectively. For this test, the minimum score difference $R$ to determine statistical difference is $R=65$ for the overall comparison and $R=24$ for each single image \cite{DAVID88}. We also performed the $\chi^2$ test with Yates correction \cite{SIEGEL88}, setting a lower bound of $\chi^2=7.82$ for this test.}
    \label{tab:preference}
    
    \centering
    \resizebox{\columnwidth}{!}{
        \begin{tabular}{|c|c|c|c|c|}
            \hline
            Image & Agree. Coeff. $u$ & $\chi^2$ & Sign. $u$ & Groups\\
            \hline
            1 & 0.460 &  73.38 & OK & \mymkg{FC - 32; F - 40} \mymkr{O - 117}\\
            2 & 0.808 & 126.61 & OK & \mymkb{F - 5} \mymkg{FC - 63} \mymkr{O - 121}\\
            3 & 0.856 & 134.00 & OK & \mymkb{F - 2} \mymkg{FC - 68} \mymkr{O - 119}\\
            4 & 0.830 & 129.92 & OK & \mymkb{F - 4} \mymkg{FC - 63} \mymkr{O - 122}\\
            5 & 0.396 &  63.53 & OK & \mymkb{F - 20} \mymkg{FC - 67} \mymkr{O - 102}\\
            6 & 0.307 &  49.92 & OK & \mymkb{F - 32} \mymkg{FC - 59} \mymkr{O - 98}\\
            7 & 0.526 &  83.53 & OK & \mymkb{F - 17} \mymkg{FC - 56} \mymkr{O - 116}\\
            8 & 0.413 &  66.23 & OK & \mymkg{F - 23; FC - 52} \mymkr{O - 114}\\
            \hline
            All & 0.530 & 802.93 & OK & \mymkb{F - 143} \mymkg{FC - 460} \mymkr{O - 909}\\
            \hline
        \end{tabular}
    }
\end{table}
	\section{Conclusions}
\noindent We have presented context-aware translation: a novel unsupervised image-to-image network, trained with two adversarial classification networks. We built these classifiers using partial convolutions, allowing them to weight generated images differently and independently. We introduced novel loss functions for these discriminators and a novel reconstruction loss based on image triplets to achieve context-aware translation. We proposed a deep-learning approach with a novel character design clustering to automatically collect training data from animation frames and input data during inference.

We have shown the method is capable of automatically redrawing the eyes of an anime character according to a provided character design direction. In this way, the shape and color of the iris can be changed, features, like reflexes and shades, can be added and the level of detail can be increased under the precise artist's control. We have made the method easily replicable, by removing the need for internal production data or labeled data. Given the general nature of our method, we expect it to be usable or extendable to other elements in and outside of animation. Only our frequency threshold value and the quality-split criteria could be specific to our use case of animation, but such is the case of many meta parameters in deep learning methods.

The obtained results also indicate the model might be presenting emergent behavior of precise image segmentation, despite never having seen segmented data, which we plan to explore in future work. As the models run nearly in real-time, Re:Draw could be used interactively as artists draw or color line-art. Only post-processing, which is significantly more computationally intensive, prevents the entire method from being run interactively. Given blending is also the main culprit behind artifacts, the focus of future work will be to further improve the method, likely within the reconstruction loss, to remove the need for post-processing altogether.

We have substantiated the need for such a style-driven enhancement with a professional user survey that reported the impact in the time of high quality drawing of the face details. Finally, we have validated our approach and results with ablation and a user study showing that our style-normalized latent space pushes the state of the art regarding the identification of non-photorealistic imagery, our approach is preferred over traditional image-to-image translation 95.16\% of the time and the images generated are not discernible with respect to images drawn by artists with traditional techniques. Please look at the additional materials for videos showing frame coherence.

\section{Acknowledgments}
We would like to especially thank Jarret Martin for teaching us about the animation pipeline needs, his continuous feedback throughout the course of our research and help running the industry survey. We would also like to thank Tonari Animation and OtakuVS for offering us the production data and other elements from their works Otachan and Second Self for illustration of the research, which we did whenever possible.

All copyrighted content belongs to its original owners. Characters, music, and all other forms of creative content are owned by their original creators, and we make no claim to them. This content was made in compliance with Fair Use law for the purpose of illustration of non-profit scientific research.
    \bibliographystyle{ACM-Reference-Format}
    \bibliography{main}


\begin{thebibliography}{64}


\ifx \showCODEN    \undefined \def \showCODEN     #1{\unskip}     \fi
\ifx \showDOI      \undefined \def \showDOI       #1{#1}\fi
\ifx \showISBNx    \undefined \def \showISBNx     #1{\unskip}     \fi
\ifx \showISBNxiii \undefined \def \showISBNxiii  #1{\unskip}     \fi
\ifx \showISSN     \undefined \def \showISSN      #1{\unskip}     \fi
\ifx \showLCCN     \undefined \def \showLCCN      #1{\unskip}     \fi
\ifx \shownote     \undefined \def \shownote      #1{#1}          \fi
\ifx \showarticletitle \undefined \def \showarticletitle #1{#1}   \fi
\ifx \showURL      \undefined \def \showURL       {\relax}        \fi
\providecommand\bibfield[2]{#2}
\providecommand\bibinfo[2]{#2}
\providecommand\natexlab[1]{#1}
\providecommand\showeprint[2][]{arXiv:#2}

\bibitem[Akita et~al\mbox{.}(2020)]%
        {Kenta+2020}
\bibfield{author}{\bibinfo{person}{Kenta Akita}, \bibinfo{person}{Yuki
  Morimoto}, {and} \bibinfo{person}{Reiji Tsuruno}.}
  \bibinfo{year}{2020}\natexlab{}.
\newblock \showarticletitle{{Deep-Eyes: Fully Automatic Anime Character
  Colorization with Painting of Details on Empty Pupils}}. In
  \bibinfo{booktitle}{\emph{Eurographics 2020 - Short Papers}},
  \bibfield{editor}{\bibinfo{person}{Alexander Wilkie} {and}
  \bibinfo{person}{Francesco Banterle}} (Eds.). \bibinfo{publisher}{The
  Eurographics Association}.
\newblock
\showISBNx{978-3-03868-101-4}
\showISSN{1017-4656}
\urldef\tempurl%
\url{https://doi.org/10.2312/egs.20201023}
\showDOI{\tempurl}


\bibitem[Anonymous et~al\mbox{.}(2022)]%
        {danbooru2021}
\bibfield{author}{\bibinfo{person}{Anonymous}, \bibinfo{person}{Danbooru
  community}, {and} \bibinfo{person}{Gwern Branwen}.}
  \bibinfo{year}{2022}\natexlab{}.
\newblock \bibinfo{title}{Danbooru2021: A Large-Scale Crowdsourced and Tagged
  Anime Illustration Dataset}.
\newblock \bibinfo{howpublished}{\url{https://www.gwern.net/Danbooru2021}}.
\newblock
\urldef\tempurl%
\url{https://www.gwern.net/Danbooru2021}
\showURL{%
\tempurl}
\newblock
\shownote{Accessed: DATE}.


\bibitem[Augereau et~al\mbox{.}(2018)]%
        {Augereau+2018}
\bibfield{author}{\bibinfo{person}{Olivier Augereau}, \bibinfo{person}{Motoi
  Iwata}, {and} \bibinfo{person}{Koichi Kise}.}
  \bibinfo{year}{2018}\natexlab{}.
\newblock \showarticletitle{A Survey of Comics Research in Computer Science}.
\newblock \bibinfo{journal}{\emph{J. Imaging}} \bibinfo{volume}{4},
  \bibinfo{number}{7} (\bibinfo{year}{2018}), \bibinfo{pages}{87}.
\newblock
\urldef\tempurl%
\url{https://doi.org/10.3390/jimaging4070087}
\showDOI{\tempurl}


\bibitem[Balaban(2015)]%
        {balaban2015deep}
\bibfield{author}{\bibinfo{person}{Stephen Balaban}.}
  \bibinfo{year}{2015}\natexlab{}.
\newblock \showarticletitle{Deep learning and face recognition: the state of
  the art}.
\newblock \bibinfo{journal}{\emph{Biometric and surveillance technology for
  human and activity identification XII}}  \bibinfo{volume}{9457}
  (\bibinfo{year}{2015}), \bibinfo{pages}{68--75}.
\newblock


\bibitem[Balntas et~al\mbox{.}(2016)]%
        {balntas2016learning}
\bibfield{author}{\bibinfo{person}{Vassileios Balntas}, \bibinfo{person}{Edgar
  Riba}, \bibinfo{person}{Daniel Ponsa}, {and} \bibinfo{person}{Krystian
  Mikolajczyk}.} \bibinfo{year}{2016}\natexlab{}.
\newblock \showarticletitle{Learning local feature descriptors with triplets
  and shallow convolutional neural networks.}. In
  \bibinfo{booktitle}{\emph{Bmvc}}, Vol.~\bibinfo{volume}{1}.
  \bibinfo{pages}{3}.
\newblock


\bibitem[Boichi et~al\mbox{.}(2019)]%
        {dr-stone}
\bibfield{author}{\bibinfo{person}{Boichi}, \bibinfo{person}{{Shueisha}}, {and}
  \bibinfo{person}{{Dr.STONE Production Committee}}.}
  \bibinfo{year}{2019}\natexlab{}.
\newblock \bibinfo{title}{Dr.Stone}.
\newblock \bibinfo{howpublished}{\url{https://dr-stone.jp/}}.
\newblock
\newblock
\shownote{[Online; accessed 4-August-2023]}.


\bibitem[Ci et~al\mbox{.}(2018)]%
        {Yuanzheng+2018}
\bibfield{author}{\bibinfo{person}{Yuanzheng Ci}, \bibinfo{person}{Xinzhu Ma},
  \bibinfo{person}{Zhihui Wang}, \bibinfo{person}{Haojie Li}, {and}
  \bibinfo{person}{Zhongxuan Luo}.} \bibinfo{year}{2018}\natexlab{}.
\newblock \showarticletitle{User-Guided Deep Anime Line Art Colorization with
  Conditional Adversarial Networks}. In \bibinfo{booktitle}{\emph{Proceedings
  of the 26th ACM International Conference on Multimedia}} (Seoul, Republic of
  Korea) \emph{(\bibinfo{series}{MM '18})}. \bibinfo{publisher}{Association for
  Computing Machinery}, \bibinfo{address}{New York, NY, USA},
  \bibinfo{pages}{1536–1544}.
\newblock
\showISBNx{9781450356657}
\urldef\tempurl%
\url{https://doi.org/10.1145/3240508.3240661}
\showDOI{\tempurl}


\bibitem[Coolkyousinnjya and {Dragon Life Improvement Committee}(2017)]%
        {dragonmaid}
\bibfield{author}{\bibinfo{person}{Coolkyousinnjya} {and}
  \bibinfo{person}{{Dragon Life Improvement Committee}}.}
  \bibinfo{year}{2017}\natexlab{}.
\newblock \bibinfo{title}{Miss Kobayashi's Dragon Maid}.
\newblock
\newblock
\urldef\tempurl%
\url{https://maidragon.jp/1st/}
\showURL{%
\tempurl}
\newblock
\shownote{[Online; accessed 4-August-2023]}.


\bibitem[{Darling in the Franxx Production Committee}(2018)]%
        {franxx}
\bibfield{author}{\bibinfo{person}{{Darling in the Franxx Production
  Committee}}.} \bibinfo{year}{2018}\natexlab{}.
\newblock \bibinfo{title}{Darling in the Franxx}.
\newblock \bibinfo{howpublished}{\url{https://darli-fra.jp/staffcast/}}.
\newblock
\newblock
\shownote{[Online; accessed 4-August-2023]}.


\bibitem[David(1988)]%
        {DAVID88}
\bibfield{author}{\bibinfo{person}{H.~A. David}.}
  \bibinfo{year}{1988}\natexlab{}.
\newblock \bibinfo{booktitle}{\emph{The Method of Paired Comparisons, 2nd ed.}}
\newblock \bibinfo{publisher}{Oxford University Press}.
\newblock


\bibitem[{Eden of the East Production Committee}(2009)]%
        {eden2009}
\bibfield{author}{\bibinfo{person}{{Eden of the East Production Committee}}.}
  \bibinfo{year}{2009}\natexlab{}.
\newblock \bibinfo{title}{Eden of the East}.
\newblock \bibinfo{howpublished}{\url{https://juiz.jp/}}.
\newblock
\newblock
\shownote{[Online; accessed 4-August-2023]}.


\bibitem[{Furansujin Connection}(2016)]%
        {furansujin}
\bibfield{author}{\bibinfo{person}{{Furansujin Connection}}.}
  \bibinfo{year}{2016}\natexlab{}.
\newblock \bibinfo{title}{{Les étapes de fabrication}}.
\newblock
  \bibinfo{howpublished}{\url{https://web.archive.org/web/20220401075414/http://www.furansujinconnection.com/les-etapes-de-fabrication/.}}.
\newblock
\newblock
\shownote{[Online; accessed 2022-March-21]}.


\bibitem[Gatys et~al\mbox{.}(2016)]%
        {Gatys+2016}
\bibfield{author}{\bibinfo{person}{Leon~A. Gatys},
  \bibinfo{person}{Alexander~S. Ecker}, {and} \bibinfo{person}{Matthias
  Bethge}.} \bibinfo{year}{2016}\natexlab{}.
\newblock \showarticletitle{Image Style Transfer Using Convolutional Neural
  Networks}. In \bibinfo{booktitle}{\emph{2016 {IEEE} Conference on Computer
  Vision and Pattern Recognition, {CVPR} 2016, Las Vegas, NV, USA, June 27-30,
  2016}}. \bibinfo{publisher}{{IEEE} Computer Society},
  \bibinfo{pages}{2414--2423}.
\newblock
\urldef\tempurl%
\url{https://doi.org/10.1109/CVPR.2016.265}
\showDOI{\tempurl}


\bibitem[He et~al\mbox{.}(2016)]%
        {He+2016}
\bibfield{author}{\bibinfo{person}{Kaiming He}, \bibinfo{person}{Xiangyu
  Zhang}, \bibinfo{person}{Shaoqing Ren}, {and} \bibinfo{person}{Jian Sun}.}
  \bibinfo{year}{2016}\natexlab{}.
\newblock \showarticletitle{Deep Residual Learning for Image Recognition}. In
  \bibinfo{booktitle}{\emph{2016 IEEE Conference on Computer Vision and Pattern
  Recognition (CVPR)}}. \bibinfo{pages}{770--778}.
\newblock
\urldef\tempurl%
\url{https://doi.org/10.1109/CVPR.2016.90}
\showDOI{\tempurl}


\bibitem[Hermans et~al\mbox{.}(2017)]%
        {hermans2017defense}
\bibfield{author}{\bibinfo{person}{Alexander Hermans}, \bibinfo{person}{Lucas
  Beyer}, {and} \bibinfo{person}{Bastian Leibe}.}
  \bibinfo{year}{2017}\natexlab{}.
\newblock \showarticletitle{In defense of the triplet loss for person
  re-identification}.
\newblock \bibinfo{journal}{\emph{arXiv preprint arXiv:1703.07737}}
  (\bibinfo{year}{2017}).
\newblock


\bibitem[Huang and Belongie(2017)]%
        {Huang+2017}
\bibfield{author}{\bibinfo{person}{Xun Huang} {and} \bibinfo{person}{Serge~J.
  Belongie}.} \bibinfo{year}{2017}\natexlab{}.
\newblock \showarticletitle{Arbitrary Style Transfer in Real-Time with Adaptive
  Instance Normalization}. In \bibinfo{booktitle}{\emph{{IEEE} International
  Conference on Computer Vision, {ICCV} 2017, Venice, Italy, October 22-29,
  2017}}. \bibinfo{publisher}{{IEEE} Computer Society},
  \bibinfo{pages}{1510--1519}.
\newblock
\urldef\tempurl%
\url{https://doi.org/10.1109/ICCV.2017.167}
\showDOI{\tempurl}


\bibitem[Huang et~al\mbox{.}(2018)]%
        {Xun+2018}
\bibfield{author}{\bibinfo{person}{Xun Huang}, \bibinfo{person}{Ming-Yu Liu},
  \bibinfo{person}{Serge Belongie}, {and} \bibinfo{person}{Jan Kautz}.}
  \bibinfo{year}{2018}\natexlab{}.
\newblock \showarticletitle{Multimodal Unsupervised Image-to-image
  Translation}. In \bibinfo{booktitle}{\emph{Proceedings of the European
  Conference on Computer Vision (ECCV)}}.
\newblock


\bibitem[Isola et~al\mbox{.}(2017)]%
        {Isola+2017}
\bibfield{author}{\bibinfo{person}{Phillip Isola}, \bibinfo{person}{Jun{-}Yan
  Zhu}, \bibinfo{person}{Tinghui Zhou}, {and} \bibinfo{person}{Alexei~A.
  Efros}.} \bibinfo{year}{2017}\natexlab{}.
\newblock \showarticletitle{Image-to-Image Translation with Conditional
  Adversarial Networks}. In \bibinfo{booktitle}{\emph{2017 {IEEE} Conference on
  Computer Vision and Pattern Recognition, {CVPR} 2017, Honolulu, HI, USA, July
  21-26, 2017}}. \bibinfo{publisher}{{IEEE} Computer Society},
  \bibinfo{pages}{5967--5976}.
\newblock
\urldef\tempurl%
\url{https://doi.org/10.1109/CVPR.2017.632}
\showDOI{\tempurl}


\bibitem[Jin et~al\mbox{.}(2017)]%
        {Jin+2017}
\bibfield{author}{\bibinfo{person}{Yanghua Jin}, \bibinfo{person}{Jiakai
  Zhang}, \bibinfo{person}{Minjun Li}, \bibinfo{person}{Yingtao Tian},
  \bibinfo{person}{Huachun Zhu}, {and} \bibinfo{person}{Zhihao Fang}.}
  \bibinfo{year}{2017}\natexlab{}.
\newblock \showarticletitle{Towards the Automatic Anime Characters Creation
  with Generative Adversarial Networks}.
\newblock \bibinfo{journal}{\emph{CoRR}}  \bibinfo{volume}{abs/1708.05509}
  (\bibinfo{year}{2017}).
\newblock
\showeprint[arXiv]{1708.05509}
\urldef\tempurl%
\url{http://arxiv.org/abs/1708.05509}
\showURL{%
\tempurl}


\bibitem[Karras et~al\mbox{.}(2021a)]%
        {Karras+2021b}
\bibfield{author}{\bibinfo{person}{Tero Karras}, \bibinfo{person}{Miika
  Aittala}, \bibinfo{person}{Samuli Laine}, \bibinfo{person}{Erik
  H{\"{a}}rk{\"{o}}nen}, \bibinfo{person}{Janne Hellsten},
  \bibinfo{person}{Jaakko Lehtinen}, {and} \bibinfo{person}{Timo Aila}.}
  \bibinfo{year}{2021}\natexlab{a}.
\newblock \showarticletitle{Alias-Free Generative Adversarial Networks}. In
  \bibinfo{booktitle}{\emph{Advances in Neural Information Processing Systems
  34: Annual Conference on Neural Information Processing Systems 2021, NeurIPS
  2021, December 6-14, 2021, virtual}},
  \bibfield{editor}{\bibinfo{person}{Marc'Aurelio Ranzato},
  \bibinfo{person}{Alina Beygelzimer}, \bibinfo{person}{Yann~N. Dauphin},
  \bibinfo{person}{Percy Liang}, {and} \bibinfo{person}{Jennifer~Wortman
  Vaughan}} (Eds.). \bibinfo{pages}{852--863}.
\newblock
\urldef\tempurl%
\url{https://proceedings.neurips.cc/paper/2021/hash/076ccd93ad68be51f23707988e934906-Abstract.html}
\showURL{%
\tempurl}


\bibitem[Karras et~al\mbox{.}(2019)]%
        {Karras+2019}
\bibfield{author}{\bibinfo{person}{Tero Karras}, \bibinfo{person}{Samuli
  Laine}, {and} \bibinfo{person}{Timo Aila}.} \bibinfo{year}{2019}\natexlab{}.
\newblock \showarticletitle{A Style-Based Generator Architecture for Generative
  Adversarial Networks}. In \bibinfo{booktitle}{\emph{{IEEE} Conference on
  Computer Vision and Pattern Recognition, {CVPR} 2019, Long Beach, CA, USA,
  June 16-20, 2019}}. \bibinfo{publisher}{Computer Vision Foundation / {IEEE}},
  \bibinfo{pages}{4401--4410}.
\newblock
\urldef\tempurl%
\url{https://doi.org/10.1109/CVPR.2019.00453}
\showDOI{\tempurl}


\bibitem[Karras et~al\mbox{.}(2021b)]%
        {Karras+2021}
\bibfield{author}{\bibinfo{person}{Tero Karras}, \bibinfo{person}{Samuli
  Laine}, {and} \bibinfo{person}{Timo Aila}.} \bibinfo{year}{2021}\natexlab{b}.
\newblock \showarticletitle{A Style-Based Generator Architecture for Generative
  Adversarial Networks}.
\newblock \bibinfo{journal}{\emph{{IEEE} Trans. Pattern Anal. Mach. Intell.}}
  \bibinfo{volume}{43}, \bibinfo{number}{12} (\bibinfo{year}{2021}),
  \bibinfo{pages}{4217--4228}.
\newblock
\urldef\tempurl%
\url{https://doi.org/10.1109/TPAMI.2020.2970919}
\showDOI{\tempurl}


\bibitem[Karras et~al\mbox{.}(2020)]%
        {Karras+2020}
\bibfield{author}{\bibinfo{person}{Tero Karras}, \bibinfo{person}{Samuli
  Laine}, \bibinfo{person}{Miika Aittala}, \bibinfo{person}{Janne Hellsten},
  \bibinfo{person}{Jaakko Lehtinen}, {and} \bibinfo{person}{Timo Aila}.}
  \bibinfo{year}{2020}\natexlab{}.
\newblock \showarticletitle{Analyzing and Improving the Image Quality of
  StyleGAN}. In \bibinfo{booktitle}{\emph{2020 {IEEE/CVF} Conference on
  Computer Vision and Pattern Recognition, {CVPR} 2020, Seattle, WA, USA, June
  13-19, 2020}}. \bibinfo{publisher}{Computer Vision Foundation / {IEEE}},
  \bibinfo{pages}{8107--8116}.
\newblock
\urldef\tempurl%
\url{https://doi.org/10.1109/CVPR42600.2020.00813}
\showDOI{\tempurl}


\bibitem[Kawamori et~al\mbox{.}(2011)]%
        {aquarion}
\bibfield{author}{\bibinfo{person}{Shoji Kawamori},
  \bibinfo{person}{{SATELIGHT}}, {and} \bibinfo{person}{{Project AQUARION
  Evol}}.} \bibinfo{year}{2011}\natexlab{}.
\newblock \bibinfo{title}{Aquarion Evol}.
\newblock \bibinfo{howpublished}{\url{http://aqevol.com/staff_cast/index.php}}.
\newblock
\newblock
\shownote{[Online; accessed 4-August-2023]}.


\bibitem[Key and {Kyoto Animation}(2007)]%
        {clannad}
\bibfield{author}{\bibinfo{person}{Key} {and} \bibinfo{person}{{Kyoto
  Animation}}.} \bibinfo{year}{2007}\natexlab{}.
\newblock \bibinfo{title}{Clannad}.
\newblock
  \bibinfo{howpublished}{\url{https://www.tbs.co.jp/clannad/clannad1/02staffcast/staffcast.html}}.
\newblock
\newblock
\shownote{[Online; accessed 4-August-2023]}.


\bibitem[Kim et~al\mbox{.}(2019)]%
        {Kim+2019}
\bibfield{author}{\bibinfo{person}{Hyunsu Kim}, \bibinfo{person}{Ho~Young
  Jhoo}, \bibinfo{person}{Eunhyeok Park}, {and} \bibinfo{person}{Sungjoo Yoo}.}
  \bibinfo{year}{2019}\natexlab{}.
\newblock \showarticletitle{Tag2Pix: Line Art Colorization Using Text Tag With
  SECat and Changing Loss}. In \bibinfo{booktitle}{\emph{2019 {IEEE/CVF}
  International Conference on Computer Vision, {ICCV} 2019, Seoul, Korea
  (South), October 27 - November 2, 2019}}. \bibinfo{publisher}{{IEEE}},
  \bibinfo{pages}{9055--9064}.
\newblock
\urldef\tempurl%
\url{https://doi.org/10.1109/ICCV.2019.00915}
\showDOI{\tempurl}


\bibitem[Kim et~al\mbox{.}(2020)]%
        {Kim+2020}
\bibfield{author}{\bibinfo{person}{Junho Kim}, \bibinfo{person}{Minjae Kim},
  \bibinfo{person}{Hyeonwoo Kang}, {and} \bibinfo{person}{Kwang~Hee Lee}.}
  \bibinfo{year}{2020}\natexlab{}.
\newblock \showarticletitle{U-GAT-IT: Unsupervised Generative Attentional
  Networks with Adaptive Layer-Instance Normalization for Image-to-Image
  Translation}. In \bibinfo{booktitle}{\emph{International Conference on
  Learning Representations}}.
\newblock
\urldef\tempurl%
\url{https://openreview.net/forum?id=BJlZ5ySKPH}
\showURL{%
\tempurl}


\bibitem[LahIntheFutureland(2023)]%
        {sdlx}
\bibfield{author}{\bibinfo{person}{LahIntheFutureland}.}
  \bibinfo{year}{2023}\natexlab{}.
\newblock \bibinfo{title}{Mysteriousv4 SDXL 1.0}.
\newblock
  \bibinfo{howpublished}{\url{https://civitai.com/models/118441/lah-mysterious-or-sdxl}}.
\newblock
\newblock
\shownote{[Online; accessed 19-October-2023]}.


\bibitem[Lee et~al\mbox{.}(2019)]%
        {Lee+2019}
\bibfield{author}{\bibinfo{person}{Gayoung Lee}, \bibinfo{person}{Dohyun Kim},
  \bibinfo{person}{Youngjoon Yoo}, \bibinfo{person}{Dongyoon Han},
  \bibinfo{person}{Jung-Woo Ha}, {and} \bibinfo{person}{Jaehyuk Chang}.}
  \bibinfo{year}{2019}\natexlab{}.
\newblock \showarticletitle{Unpaired Sketch-to-Line Translation via Synthesis
  of Sketches}. In \bibinfo{booktitle}{\emph{SIGGRAPH Asia 2019 Technical
  Briefs}} (Brisbane, QLD, Australia) \emph{(\bibinfo{series}{SA '19})}.
  \bibinfo{publisher}{Association for Computing Machinery},
  \bibinfo{address}{New York, NY, USA}, \bibinfo{pages}{45–48}.
\newblock
\showISBNx{9781450369459}
\urldef\tempurl%
\url{https://doi.org/10.1145/3355088.3365163}
\showDOI{\tempurl}


\bibitem[Li et~al\mbox{.}(2021)]%
        {Li+2021}
\bibfield{author}{\bibinfo{person}{Bing Li}, \bibinfo{person}{Yuanlue Zhu},
  \bibinfo{person}{Yitong Wang}, \bibinfo{person}{Chia-Wen Lin},
  \bibinfo{person}{Bernard Ghanem}, {and} \bibinfo{person}{Linlin Shen}.}
  \bibinfo{year}{2021}\natexlab{}.
\newblock \showarticletitle{AniGAN: Style-Guided Generative Adversarial
  Networks for Unsupervised Anime Face Generation}.
\newblock \bibinfo{journal}{\emph{IEEE Transactions on Multimedia}}
  (\bibinfo{year}{2021}), \bibinfo{pages}{1--1}.
\newblock
\urldef\tempurl%
\url{https://doi.org/10.1109/TMM.2021.3113786}
\showDOI{\tempurl}


\bibitem[Li et~al\mbox{.}(2017)]%
        {Li+2017lines}
\bibfield{author}{\bibinfo{person}{Chengze Li}, \bibinfo{person}{Xueting Liu},
  {and} \bibinfo{person}{Tien-Tsin Wong}.} \bibinfo{year}{2017}\natexlab{}.
\newblock \showarticletitle{Deep Extraction of Manga Structural Lines}.
\newblock \bibinfo{journal}{\emph{ACM Trans. Graph.}} \bibinfo{volume}{36},
  \bibinfo{number}{4}, Article \bibinfo{articleno}{117} (\bibinfo{date}{7}
  \bibinfo{year}{2017}), \bibinfo{numpages}{12}~pages.
\newblock
\showISSN{0730-0301}
\urldef\tempurl%
\url{https://doi.org/10.1145/3072959.3073675}
\showDOI{\tempurl}


\bibitem[Liu et~al\mbox{.}(2019)]%
        {liu2019few}
\bibfield{author}{\bibinfo{person}{Ming-Yu Liu}, \bibinfo{person}{Xun Huang},
  \bibinfo{person}{Arun Mallya}, \bibinfo{person}{Tero Karras},
  \bibinfo{person}{Timo Aila}, \bibinfo{person}{Jaakko Lehtinen}, {and}
  \bibinfo{person}{Jan Kautz}.} \bibinfo{year}{2019}\natexlab{}.
\newblock \showarticletitle{Few-shot unsupervised image-to-image translation}.
  In \bibinfo{booktitle}{\emph{Proceedings of the IEEE/CVF International
  Conference on Computer Vision}}. \bibinfo{pages}{10551--10560}.
\newblock


\bibitem[Maejima et~al\mbox{.}(2021)]%
        {Akinobu+2021}
\bibfield{author}{\bibinfo{person}{Akinobu Maejima}, \bibinfo{person}{Hiroyuki
  Kubo}, \bibinfo{person}{Seitaro Shinagawa}, \bibinfo{person}{Takuya
  Funatomi}, \bibinfo{person}{Tatsuo Yotsukura}, \bibinfo{person}{Satoshi
  Nakamura}, {and} \bibinfo{person}{Yasuhiro Mukaigawa}.}
  \bibinfo{year}{2021}\natexlab{}.
\newblock \bibinfo{booktitle}{\emph{Anime Character Colorization Using Few-Shot
  Learning}}.
\newblock \bibinfo{publisher}{Association for Computing Machinery},
  \bibinfo{address}{New York, NY, USA}.
\newblock
\showISBNx{9781450390736}
\urldef\tempurl%
\url{https://doi.org/10.1145/3478512.3488604}
\showURL{%
\tempurl}


\bibitem[Masuda et~al\mbox{.}(2019)]%
        {Hiromichi2019}
\bibfield{author}{\bibinfo{person}{Hiromichi Masuda}, \bibinfo{person}{Tadashi
  Sudo}, \bibinfo{person}{Kazuo Rikukawa}, \bibinfo{person}{Yuji Mori},
  \bibinfo{person}{Naofumi Ito}, \bibinfo{person}{Yasuo Kameyama}, {and}
  \bibinfo{person}{Megumi Onouchi}.} \bibinfo{year}{2019}\natexlab{}.
\newblock \bibinfo{booktitle}{\emph{{Anime Industry Report}}}.
\newblock \bibinfo{type}{{T}echnical {R}eport}. \bibinfo{institution}{The
  Association of Japanese Animations}.
\newblock


\bibitem[Mescheder et~al\mbox{.}(2018)]%
        {mescheder2018training}
\bibfield{author}{\bibinfo{person}{Lars Mescheder}, \bibinfo{person}{Andreas
  Geiger}, {and} \bibinfo{person}{Sebastian Nowozin}.}
  \bibinfo{year}{2018}\natexlab{}.
\newblock \showarticletitle{Which training methods for GANs do actually
  converge?}. In \bibinfo{booktitle}{\emph{International conference on machine
  learning}}. PMLR, \bibinfo{pages}{3481--3490}.
\newblock


\bibitem[Nagatsuki et~al\mbox{.}(2016)]%
        {re-zero}
\bibfield{author}{\bibinfo{person}{Tappei Nagatsuki},
  \bibinfo{person}{{KADOKAWA}}, {and} \bibinfo{person}{{Re: Life in a Different
  World Starting from Zero 1 Production Committee}}.}
  \bibinfo{year}{2016}\natexlab{}.
\newblock \bibinfo{title}{Re: Life in a Different World Starting from Zero}.
\newblock \bibinfo{howpublished}{\url{http://re-zero-anime.jp/tv/}}.
\newblock
\newblock
\shownote{[Online; accessed 4-August-2023]}.


\bibitem[Nir et~al\mbox{.}(2022)]%
        {Nir+2022}
\bibfield{author}{\bibinfo{person}{Oron Nir}, \bibinfo{person}{Gal Rapoport},
  {and} \bibinfo{person}{Ariel Shamir}.} \bibinfo{year}{2022}\natexlab{}.
\newblock \showarticletitle{{CAST: Character labeling in Animation using
  Self-supervision by Tracking}}.
\newblock \bibinfo{journal}{\emph{Computer Graphics Forum}}
  (\bibinfo{year}{2022}).
\newblock
\showISSN{1467-8659}
\urldef\tempurl%
\url{https://doi.org/10.1111/cgf.14464}
\showDOI{\tempurl}


\bibitem[Nizan and Tal(2020)]%
        {Nizan+2020}
\bibfield{author}{\bibinfo{person}{Ori Nizan} {and} \bibinfo{person}{Ayellet
  Tal}.} \bibinfo{year}{2020}\natexlab{}.
\newblock \showarticletitle{Breaking the Cycle - Colleagues Are All You Need}.
  In \bibinfo{booktitle}{\emph{2020 {IEEE/CVF} Conference on Computer Vision
  and Pattern Recognition, {CVPR} 2020, Seattle, WA, USA, June 13-19, 2020}}.
  \bibinfo{publisher}{Computer Vision Foundation / {IEEE}},
  \bibinfo{pages}{7857--7866}.
\newblock
\urldef\tempurl%
\url{https://doi.org/10.1109/CVPR42600.2020.00788}
\showDOI{\tempurl}


\bibitem[Oda and {Toei Animation}(1997)]%
        {onepiece}
\bibfield{author}{\bibinfo{person}{Eiichiro Oda} {and} \bibinfo{person}{{Toei
  Animation}}.} \bibinfo{year}{1997}\natexlab{}.
\newblock \bibinfo{title}{{One Piece}}.
\newblock \bibinfo{howpublished}{\url{https://one-piece.com/anime/}}.
\newblock
\newblock
\shownote{[Online; accessed 4-August-2023]}.


\bibitem[Ojha et~al\mbox{.}(2021)]%
        {ojha2021few}
\bibfield{author}{\bibinfo{person}{Utkarsh Ojha}, \bibinfo{person}{Yijun Li},
  \bibinfo{person}{Jingwan Lu}, \bibinfo{person}{Alexei~A Efros},
  \bibinfo{person}{Yong~Jae Lee}, \bibinfo{person}{Eli Shechtman}, {and}
  \bibinfo{person}{Richard Zhang}.} \bibinfo{year}{2021}\natexlab{}.
\newblock \showarticletitle{Few-shot image generation via cross-domain
  correspondence}. In \bibinfo{booktitle}{\emph{Proceedings of the IEEE/CVF
  Conference on Computer Vision and Pattern Recognition}}.
  \bibinfo{pages}{10743--10752}.
\newblock


\bibitem[Oum and {Team RWBY Project}(2022)]%
        {rwby}
\bibfield{author}{\bibinfo{person}{Monyreak Oum} {and} \bibinfo{person}{{Team
  RWBY Project}}.} \bibinfo{year}{2022}\natexlab{}.
\newblock \bibinfo{title}{{RWBY: Ice Queendom}}.
\newblock
  \bibinfo{howpublished}{\url{https://anime.team-rwby-project.jp/staff/}}.
\newblock
\newblock
\shownote{[Online; accessed 4-August-2023]}.


\bibitem[P{\'e}rez et~al\mbox{.}(2003)]%
        {perez2003poisson}
\bibfield{author}{\bibinfo{person}{Patrick P{\'e}rez}, \bibinfo{person}{Michel
  Gangnet}, {and} \bibinfo{person}{Andrew Blake}.}
  \bibinfo{year}{2003}\natexlab{}.
\newblock \showarticletitle{Poisson image editing}.
\newblock In \bibinfo{booktitle}{\emph{ACM SIGGRAPH 2003 Papers}}.
  \bibinfo{pages}{313--318}.
\newblock


\bibitem[Reinhard et~al\mbox{.}(2001)]%
        {reinhard2001color}
\bibfield{author}{\bibinfo{person}{Erik Reinhard}, \bibinfo{person}{Michael
  Adhikhmin}, \bibinfo{person}{Bruce Gooch}, {and} \bibinfo{person}{Peter
  Shirley}.} \bibinfo{year}{2001}\natexlab{}.
\newblock \showarticletitle{Color transfer between images}.
\newblock \bibinfo{journal}{\emph{IEEE Computer graphics and applications}}
  \bibinfo{volume}{21}, \bibinfo{number}{5} (\bibinfo{year}{2001}),
  \bibinfo{pages}{34--41}.
\newblock


\bibitem[Ren et~al\mbox{.}(2015)]%
        {ren2015faster}
\bibfield{author}{\bibinfo{person}{Shaoqing Ren}, \bibinfo{person}{Kaiming He},
  \bibinfo{person}{Ross Girshick}, {and} \bibinfo{person}{Jian Sun}.}
  \bibinfo{year}{2015}\natexlab{}.
\newblock \showarticletitle{Faster R-CNN: Towards Real-Time Object Detection
  with Region Proposal Networks}. In \bibinfo{booktitle}{\emph{Proceedings of
  the 28th International Conference on Neural Information Processing Systems -
  Volume 1}} (Montreal, Canada) \emph{(\bibinfo{series}{NIPS'15})}.
  \bibinfo{publisher}{MIT Press}, \bibinfo{address}{Cambridge, MA, USA},
  \bibinfo{pages}{91–99}.
\newblock


\bibitem[Saito et~al\mbox{.}(2020)]%
        {Saito+2020}
\bibfield{author}{\bibinfo{person}{Kuniaki Saito}, \bibinfo{person}{Kate
  Saenko}, {and} \bibinfo{person}{Ming{-}Yu Liu}.}
  \bibinfo{year}{2020}\natexlab{}.
\newblock \showarticletitle{{COCO-FUNIT:} Few-Shot Unsupervised Image
  Translation with a Content Conditioned Style Encoder}. In
  \bibinfo{booktitle}{\emph{Computer Vision - {ECCV} 2020 - 16th European
  Conference, Glasgow, UK, August 23-28, 2020, Proceedings, Part {III}}}
  \emph{(\bibinfo{series}{Lecture Notes in Computer Science},
  Vol.~\bibinfo{volume}{12348})}, \bibfield{editor}{\bibinfo{person}{Andrea
  Vedaldi}, \bibinfo{person}{Horst Bischof}, \bibinfo{person}{Thomas Brox},
  {and} \bibinfo{person}{Jan{-}Michael Frahm}} (Eds.).
  \bibinfo{publisher}{Springer}, \bibinfo{pages}{382--398}.
\newblock
\urldef\tempurl%
\url{https://doi.org/10.1007/978-3-030-58580-8\_23}
\showDOI{\tempurl}


\bibitem[Sanada et~al\mbox{.}(2018)]%
        {angelsofdeath}
\bibfield{author}{\bibinfo{person}{Makoto Sanada},
  \bibinfo{person}{{Kadokawa}}, {and} \bibinfo{person}{{Angels of Death
  Production Committee}}.} \bibinfo{year}{2018}\natexlab{}.
\newblock \bibinfo{title}{Angels of Death}.
\newblock
  \bibinfo{howpublished}{\url{http://www.jcstaff.co.jp/sakuhin/nenpyo/2018/06_satsuriku/satsuriku.htm}}.
\newblock
\newblock
\shownote{[Online; accessed 4-August-2023]}.


\bibitem[Schroff et~al\mbox{.}(2015)]%
        {schroff2015facenet}
\bibfield{author}{\bibinfo{person}{Florian Schroff}, \bibinfo{person}{Dmitry
  Kalenichenko}, {and} \bibinfo{person}{James Philbin}.}
  \bibinfo{year}{2015}\natexlab{}.
\newblock \showarticletitle{Facenet: A unified embedding for face recognition
  and clustering}. In \bibinfo{booktitle}{\emph{Proceedings of the IEEE
  conference on computer vision and pattern recognition}}.
  \bibinfo{pages}{815--823}.
\newblock


\bibitem[Shimizu et~al\mbox{.}(2021)]%
        {Shimizu+2021}
\bibfield{author}{\bibinfo{person}{Yugo Shimizu}, \bibinfo{person}{Ryosuke
  Furuta}, \bibinfo{person}{Delong Ouyang}, \bibinfo{person}{Yukinobu
  Taniguchi}, \bibinfo{person}{Ryota Hinami}, {and} \bibinfo{person}{Shonosuke
  Ishiwatari}.} \bibinfo{year}{2021}\natexlab{}.
\newblock \showarticletitle{Painting Style-Aware Manga Colorization Based On
  Generative Adversarial Networks}. In \bibinfo{booktitle}{\emph{2021 {IEEE}
  International Conference on Image Processing, {ICIP} 2021, Anchorage, AK,
  USA, September 19-22, 2021}}. \bibinfo{publisher}{{IEEE}},
  \bibinfo{pages}{1739--1743}.
\newblock
\urldef\tempurl%
\url{https://doi.org/10.1109/ICIP42928.2021.9506254}
\showDOI{\tempurl}


\bibitem[Siegel and Castellan(1988)]%
        {SIEGEL88}
\bibfield{author}{\bibinfo{person}{Sidney Siegel} {and}
  \bibinfo{person}{N.~John Castellan}.} \bibinfo{year}{1988}\natexlab{}.
\newblock \bibinfo{booktitle}{\emph{Nonparametric Statistics for the Behavioral
  Sciences}}.
\newblock \bibinfo{publisher}{McGrall-Hill International}.
\newblock


\bibitem[Silva et~al\mbox{.}(2019)]%
        {Silva+2019}
\bibfield{author}{\bibinfo{person}{Felipe~Coelho Silva}, \bibinfo{person}{Paulo
  Andr{\'{e}}~Lima de Castro}, \bibinfo{person}{H{\'{e}}lio~Ricardo
  J{\'{u}}nior}, {and} \bibinfo{person}{Ernesto~Cordeiro Marujo}.}
  \bibinfo{year}{2019}\natexlab{}.
\newblock \showarticletitle{Mangan: Assisting Colorization Of Manga Characters
  Concept Art Using Conditional {GAN}}. In \bibinfo{booktitle}{\emph{2019
  {IEEE} International Conference on Image Processing, {ICIP} 2019, Taipei,
  Taiwan, September 22-25, 2019}}. \bibinfo{publisher}{{IEEE}},
  \bibinfo{pages}{3257--3261}.
\newblock
\urldef\tempurl%
\url{https://doi.org/10.1109/ICIP.2019.8803667}
\showDOI{\tempurl}


\bibitem[Simo-Serra et~al\mbox{.}(2018a)]%
        {Serra+2018}
\bibfield{author}{\bibinfo{person}{Edgar Simo-Serra}, \bibinfo{person}{Satoshi
  Iizuka}, {and} \bibinfo{person}{Hiroshi Ishikawa}.}
  \bibinfo{year}{2018}\natexlab{a}.
\newblock \showarticletitle{Mastering Sketching: Adversarial Augmentation for
  Structured Prediction}.
\newblock \bibinfo{journal}{\emph{ACM Trans. Graph.}} \bibinfo{volume}{37},
  \bibinfo{number}{1}, Article \bibinfo{articleno}{11} (\bibinfo{date}{1}
  \bibinfo{year}{2018}), \bibinfo{numpages}{13}~pages.
\newblock
\showISSN{0730-0301}
\urldef\tempurl%
\url{https://doi.org/10.1145/3132703}
\showDOI{\tempurl}


\bibitem[Simo-Serra et~al\mbox{.}(2018b)]%
        {Serra+2018b}
\bibfield{author}{\bibinfo{person}{Edgar Simo-Serra}, \bibinfo{person}{Satoshi
  Iizuka}, {and} \bibinfo{person}{Hiroshi Ishikawa}.}
  \bibinfo{year}{2018}\natexlab{b}.
\newblock \showarticletitle{Real-Time Data-Driven Interactive Rough Sketch
  Inking}.
\newblock \bibinfo{journal}{\emph{ACM Trans. Graph.}} \bibinfo{volume}{37},
  \bibinfo{number}{4}, Article \bibinfo{articleno}{98} (\bibinfo{date}{7}
  \bibinfo{year}{2018}), \bibinfo{numpages}{14}~pages.
\newblock
\showISSN{0730-0301}
\urldef\tempurl%
\url{https://doi.org/10.1145/3197517.3201370}
\showDOI{\tempurl}


\bibitem[Simo-Serra et~al\mbox{.}(2016)]%
        {Serra+2016}
\bibfield{author}{\bibinfo{person}{Edgar Simo-Serra}, \bibinfo{person}{Satoshi
  Iizuka}, \bibinfo{person}{Kazuma Sasaki}, {and} \bibinfo{person}{Hiroshi
  Ishikawa}.} \bibinfo{year}{2016}\natexlab{}.
\newblock \showarticletitle{Learning to Simplify: Fully Convolutional Networks
  for Rough Sketch Cleanup}.
\newblock \bibinfo{journal}{\emph{ACM Trans. Graph.}} \bibinfo{volume}{35},
  \bibinfo{number}{4}, Article \bibinfo{articleno}{121} (\bibinfo{date}{7}
  \bibinfo{year}{2016}), \bibinfo{numpages}{11}~pages.
\newblock
\showISSN{0730-0301}
\urldef\tempurl%
\url{https://doi.org/10.1145/2897824.2925972}
\showDOI{\tempurl}


\bibitem[Su et~al\mbox{.}(2021)]%
        {Su+2021}
\bibfield{author}{\bibinfo{person}{Hao Su}, \bibinfo{person}{Jianwei Niu},
  \bibinfo{person}{Xuefeng Liu}, \bibinfo{person}{Qingfeng Li},
  \bibinfo{person}{Jiahe Cui}, {and} \bibinfo{person}{Ji Wan}.}
  \bibinfo{year}{2021}\natexlab{}.
\newblock \showarticletitle{MangaGAN: Unpaired Photo-to-Manga Translation Based
  on The Methodology of Manga Drawing}. In
  \bibinfo{booktitle}{\emph{Thirty-Fifth {AAAI} Conference on Artificial
  Intelligence, {AAAI} 2021, Thirty-Third Conference on Innovative Applications
  of Artificial Intelligence, {IAAI} 2021, The Eleventh Symposium on
  Educational Advances in Artificial Intelligence, {EAAI} 2021, Virtual Event,
  February 2-9, 2021}}. \bibinfo{publisher}{{AAAI} Press},
  \bibinfo{pages}{2611--2619}.
\newblock
\urldef\tempurl%
\url{https://ojs.aaai.org/index.php/AAAI/article/view/16364}
\showURL{%
\tempurl}


\bibitem[Van~der Maaten and Hinton(2008)]%
        {van2008visualizing}
\bibfield{author}{\bibinfo{person}{Laurens Van~der Maaten} {and}
  \bibinfo{person}{Geoffrey Hinton}.} \bibinfo{year}{2008}\natexlab{}.
\newblock \showarticletitle{Visualizing data using t-SNE.}
\newblock \bibinfo{journal}{\emph{Journal of machine learning research}}
  \bibinfo{volume}{9}, \bibinfo{number}{11} (\bibinfo{year}{2008}).
\newblock


\bibitem[Wang and Yu(2020)]%
        {Wang+2020}
\bibfield{author}{\bibinfo{person}{Xinrui Wang} {and} \bibinfo{person}{Jinze
  Yu}.} \bibinfo{year}{2020}\natexlab{}.
\newblock \showarticletitle{Learning to Cartoonize Using White-Box Cartoon
  Representations}. In \bibinfo{booktitle}{\emph{Proceedings of the IEEE/CVF
  Conference on Computer Vision and Pattern Recognition (CVPR)}}.
\newblock


\bibitem[Yamazaki et~al\mbox{.}(2017)]%
        {magusbride}
\bibfield{author}{\bibinfo{person}{Kore Yamazaki}, \bibinfo{person}{{Mag
  Garden}}, {and} \bibinfo{person}{{The Ancient Magus' Bride Production
  Committee}}.} \bibinfo{year}{2017}\natexlab{}.
\newblock \bibinfo{title}{The Ancient Magus Bride}.
\newblock \bibinfo{howpublished}{\url{https://mahoyome.jp/staffcast/}}.
\newblock
\newblock
\shownote{[Online; accessed 4-August-2023]}.


\bibitem[Yarbus(1967)]%
        {yarbus1967}
\bibfield{author}{\bibinfo{person}{Alfred~L. Yarbus}.}
  \bibinfo{year}{1967}\natexlab{}.
\newblock \bibinfo{booktitle}{\emph{Eye Movements and Vision}}.
\newblock \bibinfo{publisher}{Plenum Press}, \bibinfo{address}{New York}.
\newblock
\newblock
\shownote{Translated from Russian}.


\bibitem[Yatate and {Sunrise Bandai Visual}(2007)]%
        {xenoglossia}
\bibfield{author}{\bibinfo{person}{Hajime Yatate} {and}
  \bibinfo{person}{{Sunrise Bandai Visual}}.} \bibinfo{year}{2007}\natexlab{}.
\newblock \bibinfo{title}{Idolmaster: Xenoglossia}.
\newblock
  \bibinfo{howpublished}{\url{https://www.sunrise-inc.co.jp/idolmaster/}}.
\newblock
\newblock
\shownote{[Online; accessed 4-August-2023]}.


\bibitem[You et~al\mbox{.}(2019)]%
        {You+2019}
\bibfield{author}{\bibinfo{person}{Sheng You}, \bibinfo{person}{Ning You},
  {and} \bibinfo{person}{Minxue Pan}.} \bibinfo{year}{2019}\natexlab{}.
\newblock \showarticletitle{{PI-REC:} Progressive Image Reconstruction Network
  With Edge and Color Domain}.
\newblock \bibinfo{journal}{\emph{CoRR}}  \bibinfo{volume}{abs/1903.10146}
  (\bibinfo{year}{2019}).
\newblock
\showeprint[arXiv]{1903.10146}
\urldef\tempurl%
\url{http://arxiv.org/abs/1903.10146}
\showURL{%
\tempurl}


\bibitem[Zhang et~al\mbox{.}(2020)]%
        {Zhang+2020b}
\bibfield{author}{\bibinfo{person}{Lvmin Zhang}, \bibinfo{person}{Yi JI}, {and}
  \bibinfo{person}{Chunping Liu}.} \bibinfo{year}{2020}\natexlab{}.
\newblock \showarticletitle{DanbooRegion: An Illustration Region Dataset}. In
  \bibinfo{booktitle}{\emph{European Conference on Computer Vision (ECCV)}}.
\newblock


\bibitem[Zhang et~al\mbox{.}(2021)]%
        {Zhang+2021}
\bibfield{author}{\bibinfo{person}{Lvmin Zhang}, \bibinfo{person}{Chengze Li},
  \bibinfo{person}{Edgar Simo{-}Serra}, \bibinfo{person}{Yi Ji},
  \bibinfo{person}{Tien{-}Tsin Wong}, {and} \bibinfo{person}{Chunping Liu}.}
  \bibinfo{year}{2021}\natexlab{}.
\newblock \showarticletitle{User-Guided Line Art Flat Filling With Split
  Filling Mechanism}. In \bibinfo{booktitle}{\emph{{IEEE} Conference on
  Computer Vision and Pattern Recognition, {CVPR} 2021, virtual, June 19-25,
  2021}}. \bibinfo{publisher}{Computer Vision Foundation / {IEEE}},
  \bibinfo{pages}{9889--9898}.
\newblock
\urldef\tempurl%
\url{https://openaccess.thecvf.com/content/CVPR2021/html/Zhang\_User-Guided\_Line\_Art\_Flat\_Filling\_With\_Split\_Filling\_Mechanism\_CVPR\_2021\_paper.html}
\showURL{%
\tempurl}


\bibitem[Zhang et~al\mbox{.}(2018)]%
        {Zhang+2018}
\bibfield{author}{\bibinfo{person}{Lvmin Zhang}, \bibinfo{person}{Chengze Li},
  \bibinfo{person}{Tien-Tsin Wong}, \bibinfo{person}{Yi Ji}, {and}
  \bibinfo{person}{Chunping Liu}.} \bibinfo{year}{2018}\natexlab{}.
\newblock \showarticletitle{Two-Stage Sketch Colorization}.
\newblock \bibinfo{journal}{\emph{ACM Trans. Graph.}} \bibinfo{volume}{37},
  \bibinfo{number}{6}, Article \bibinfo{articleno}{261} (\bibinfo{date}{12}
  \bibinfo{year}{2018}), \bibinfo{numpages}{14}~pages.
\newblock
\showISSN{0730-0301}
\urldef\tempurl%
\url{https://doi.org/10.1145/3272127.3275090}
\showDOI{\tempurl}


\bibitem[Zhu et~al\mbox{.}(2017)]%
        {Zhu+2017}
\bibfield{author}{\bibinfo{person}{Jun{-}Yan Zhu}, \bibinfo{person}{Taesung
  Park}, \bibinfo{person}{Phillip Isola}, {and} \bibinfo{person}{Alexei~A.
  Efros}.} \bibinfo{year}{2017}\natexlab{}.
\newblock \showarticletitle{Unpaired Image-to-Image Translation Using
  Cycle-Consistent Adversarial Networks}. In \bibinfo{booktitle}{\emph{{IEEE}
  International Conference on Computer Vision, {ICCV} 2017, Venice, Italy,
  October 22-29, 2017}}. \bibinfo{publisher}{{IEEE} Computer Society},
  \bibinfo{pages}{2242--2251}.
\newblock
\urldef\tempurl%
\url{https://doi.org/10.1109/ICCV.2017.244}
\showDOI{\tempurl}


\end{thebibliography}

    \opensource{
        \clearpage \appendix
	    \begin{center}
    \vspace*{\fill} 
    \fontsize{15}{23}\selectfont
    \textbf{To Reviewers:} it is impossible to properly demonstrate the presented case study without copyrighted works. As the journal does not allow any copyrighted material, despite using it for the purpose of exemplification being Fair-Use, we include all the copyrighted materials in additional materials only for reviewers to see. These materials will then be made available in the author's paper website.
  \vspace*{\fill}
\end{center}

\begin{figure}[b]
    \centering
    \includegraphics[width=\columnwidth]{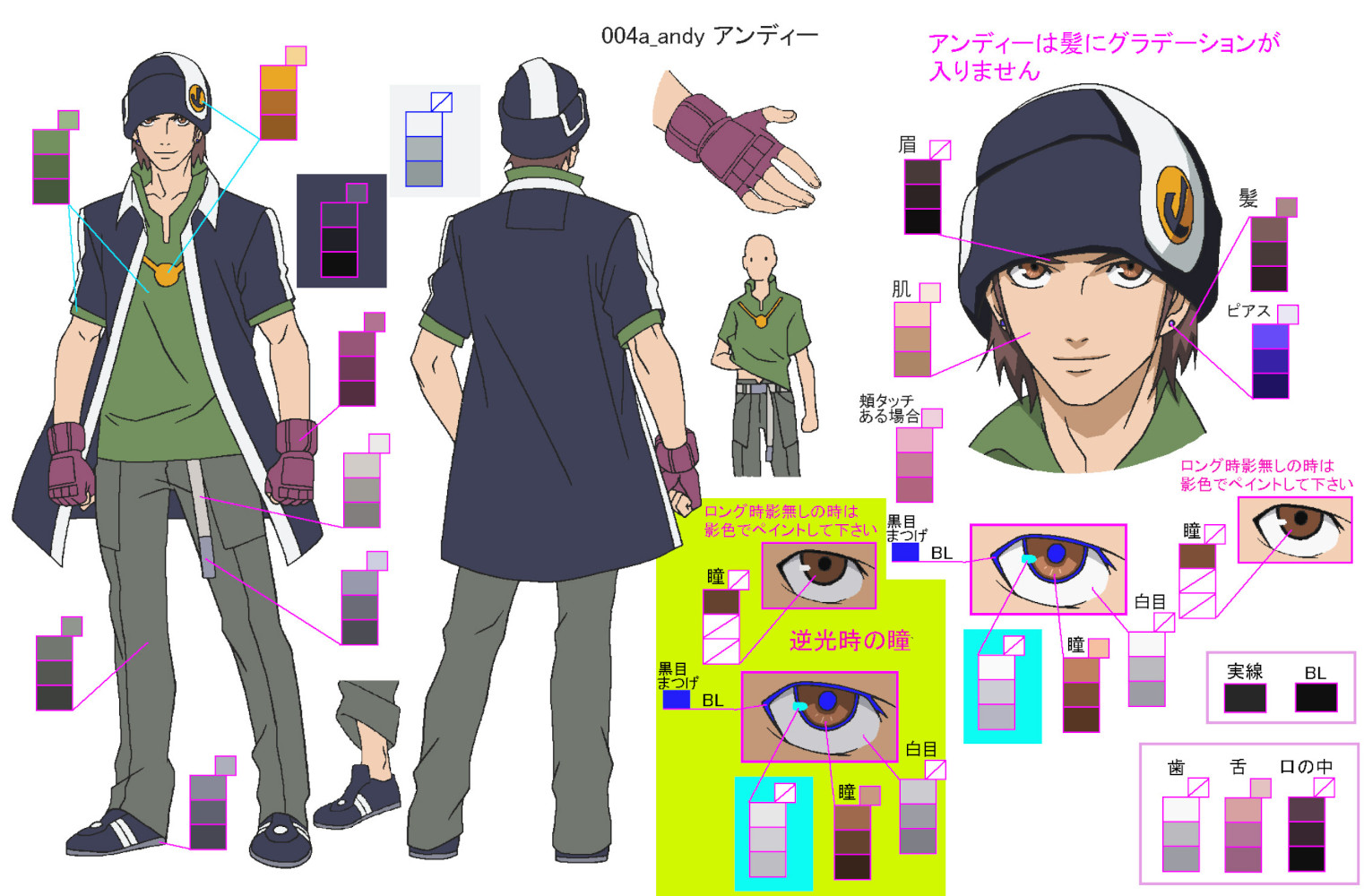}
    \caption[Color Guide]{\textbf{Color Guide}. Also known as a model sheet, this type of document depicts all the information an artist is expected to need to know how to draw a character in accordance with the production direction and style. Example from \textit{Aquarion Evol} \cite{aquarion}}.
    \label{color-guide}
\end{figure}

\begin{figure}[b]
    \centering
    \centering
    \begin{subfigure}{0.12\columnwidth}
        \centering
        \includegraphics[width=\textwidth]{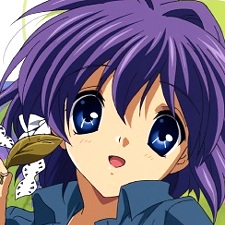}
    \end{subfigure}
    \begin{subfigure}{0.12\columnwidth}
        \centering
        \includegraphics[width=\textwidth]{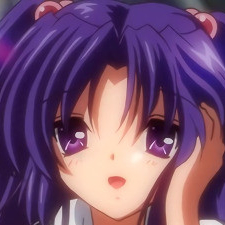}
    \end{subfigure}
    \begin{subfigure}{0.12\columnwidth}
        \centering
        \includegraphics[width=\textwidth]{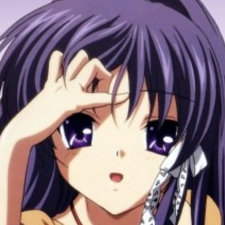}
    \end{subfigure}
    \hspace{2em}
    \begin{subfigure}{0.12\columnwidth}
        \centering
        \includegraphics[width=\textwidth]{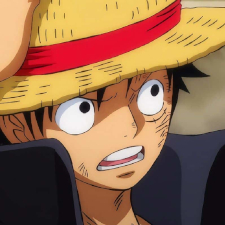}
    \end{subfigure}
    \begin{subfigure}{0.12\columnwidth}
        \centering
        \includegraphics[width=\textwidth]{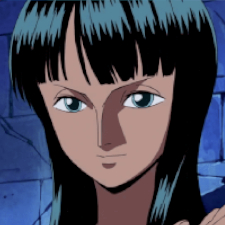}
    \end{subfigure}
    \begin{subfigure}{0.12\columnwidth}
        \centering
        \includegraphics[width=\textwidth]{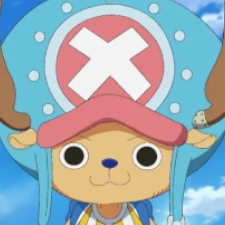}
    \end{subfigure}
    \caption{\textbf{Design Variety.} Characters in some productions have very similar structures (left \shortcite{clannad}), and yet others may present a high variety of designs (right \shortcite{onepiece}). This hinders traditional facial recognition.}
    \label{figure:character-structure}
\end{figure}

\begin{figure*}
    \centering
    \begin{subfigure}{.121\linewidth}
         \centering
         \includegraphics[width=\textwidth]{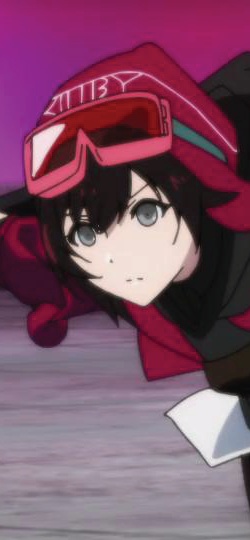}
         \caption{Input}
    \end{subfigure}
    \begin{subfigure}{.121\linewidth}
         \centering
         \includegraphics[width=\textwidth]{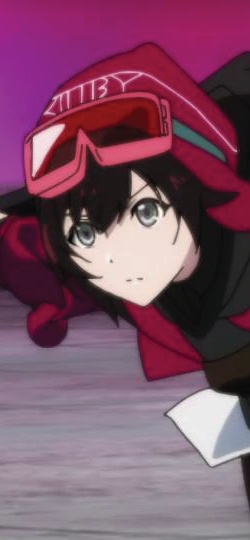}
         \caption{Ours}
    \end{subfigure}
    \begin{subfigure}{.121\linewidth}
         \centering
         \includegraphics[width=\textwidth]{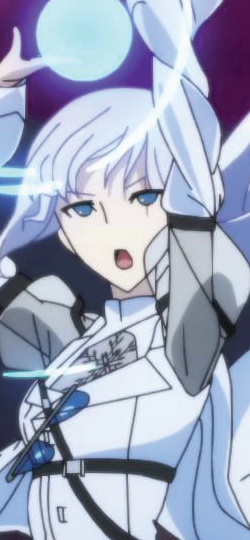}
         \caption{Input}
    \end{subfigure}
    \begin{subfigure}{.121\linewidth}
         \centering
         \includegraphics[width=\textwidth]{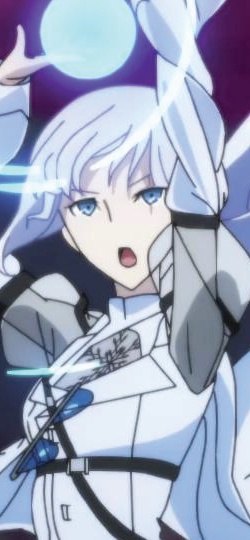}
         \caption{Ours}
    \end{subfigure}
    \begin{subfigure}{.121\linewidth}
         \centering
         \includegraphics[width=\textwidth]{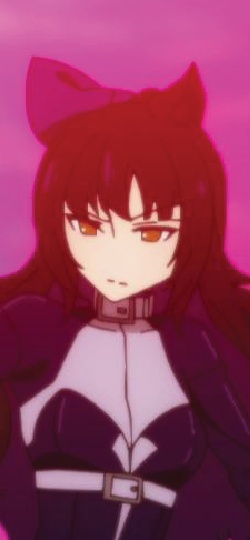}
         \caption{Input}
    \end{subfigure}
    \begin{subfigure}{.121\linewidth}
         \centering
         \includegraphics[width=\textwidth]{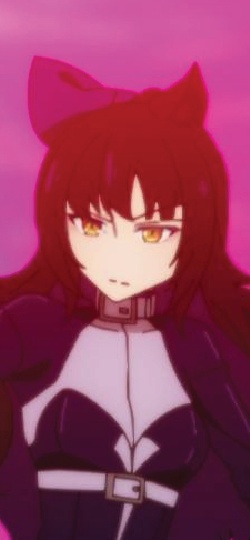}
         \caption{Ours}
    \end{subfigure}
    \begin{subfigure}{.121\linewidth}
         \centering
         \includegraphics[width=\textwidth]{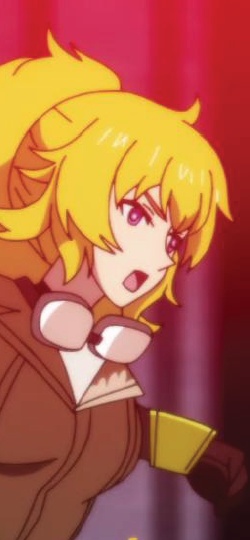}
         \caption{Input}
    \end{subfigure}
    \begin{subfigure}{.121\linewidth}
         \centering
         \includegraphics[width=\textwidth]{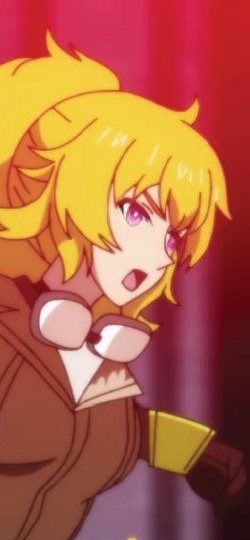}
         \caption{Ours}
    \end{subfigure}
    \vspace{-.8em}
    \caption{\textbf{Proposed Use-Case.} Context-aware translation can be used to automatically redraw the eyes of any given character according to any provided color-guide, without fine-tuning. These characters and their target designs (from \textit{RWBY: Ice Queendom} \cite{rwby}) were not part of the training set.}
    \label{fig:rwby}
\end{figure*}

\begin{figure*}
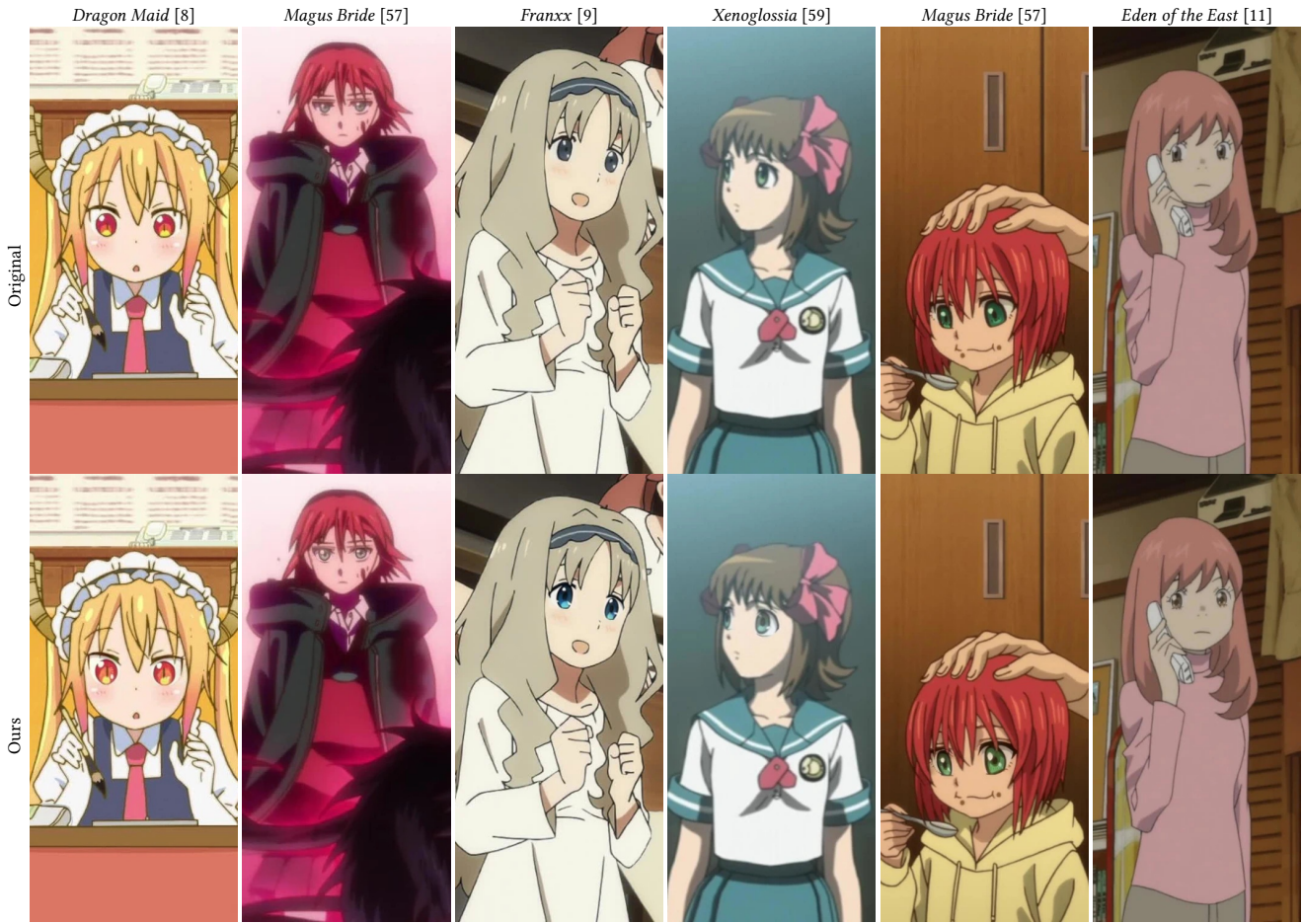

  \setlength{\tabcolsep}{1pt}
  \footnotesize{\begin{tabular}{lcccccc}
   & \textit{Dragon Maid} \cite{dragonmaid} & \textit{Magus Bride} \cite{magusbride} & \textit{Franxx} \cite{franxx} & \textit{Xenoglossia} \cite{xenoglossia} & \textit{Magus Bride} \cite{magusbride} & \textit{Eden of the East} \cite{eden2009} \\
    \raisebox{-.75cm}{\rotatebox{90}{Original}}
        & \includegraphics[valign=c, width=0.157\textwidth]{img/results/G.jpg}
        & \includegraphics[valign=c, width=0.157\textwidth]{img/results/I.jpg}
        & \includegraphics[valign=c, width=0.157\textwidth]{img/results/M.jpg}
        & \includegraphics[valign=c, width=0.157\textwidth]{img/results/C.jpg}
        & \includegraphics[valign=c, width=0.157\textwidth]{img/results/E.jpg}
        & \includegraphics[valign=c, width=0.157\textwidth]{img/results/K.jpg} \\
    \raisebox{-.75cm}{\rotatebox{90}{Ours}}
        & \includegraphics[valign=c, width=0.157\textwidth]{img/results/H.jpg}
        & \includegraphics[valign=c, width=0.157\textwidth]{img/results/J.jpg}
        & \includegraphics[valign=c, width=0.157\textwidth]{img/results/N.jpg}
        & \includegraphics[valign=c, width=0.157\textwidth]{img/results/D.jpg}
        & \includegraphics[valign=c, width=0.157\textwidth]{img/results/F.jpg}
        & \includegraphics[valign=c, width=0.157\textwidth]{img/results/L.jpg}
  \end{tabular}}
  \caption[Increasing Detail]{\textbf{Increasing Detail}. The method behaves as intended when prompted to increase the amount of detail in character eyes, despite challenges such as unusual face proportions, adverse color balance due to lighting effects, head tilts, minor hair occlusions and character designs from productions that have not been seen during training.}
  \label{detail}
\end{figure*}

\begin{figure*}
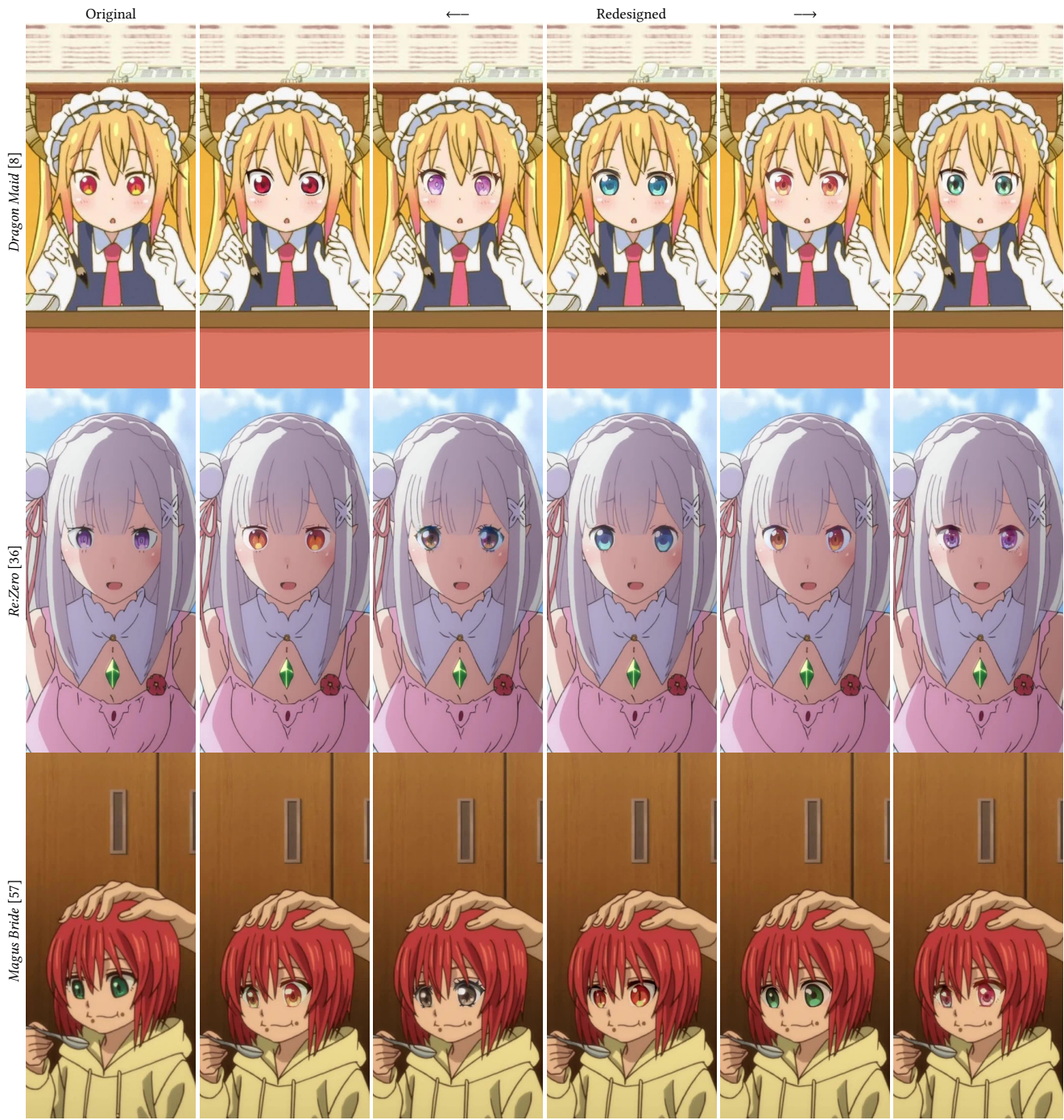

  \setlength{\tabcolsep}{1pt}
  \footnotesize{\begin{tabular}{lcccccc}
     & Original & & $\longleftarrow$  & Redesigned & $\longrightarrow$ & \\
    \raisebox{-.75cm}{\rotatebox{90}{\textit{Dragon Maid} \cite{dragonmaid}}}
        & \includegraphics[valign=c, width=0.157\textwidth]{img/results/G.jpg}
        & \includegraphics[valign=c, width=0.157\textwidth]{img/results/H1.jpg}
        & \includegraphics[valign=c, width=0.157\textwidth]{img/results/H2.jpg}
        & \includegraphics[valign=c, width=0.157\textwidth]{img/results/H3.jpg}
        & \includegraphics[valign=c, width=0.157\textwidth]{img/results/H4.jpg}
        & \includegraphics[valign=c, width=0.157\textwidth]{img/results/H5.jpg} \\
    \raisebox{-.75cm}{\rotatebox{90}{\textit{Re:Zero} \cite{re-zero}}}
        & \includegraphics[valign=c, width=0.157\textwidth]{img/results/S.jpg}
        & \includegraphics[valign=c, width=0.157\textwidth]{img/results/S1.jpg}
        & \includegraphics[valign=c, width=0.157\textwidth]{img/results/S2.jpg}
        & \includegraphics[valign=c, width=0.157\textwidth]{img/results/S3.jpg}
        & \includegraphics[valign=c, width=0.157\textwidth]{img/results/S4.jpg}
        & \includegraphics[valign=c, width=0.157\textwidth]{img/results/S5.jpg} \\
    \raisebox{-.75cm}{\rotatebox{90}{\textit{Magus Bride} \cite{magusbride}}}
        & \includegraphics[valign=c, width=0.157\textwidth]{img/results/E.jpg}
        & \includegraphics[valign=c, width=0.157\textwidth]{img/results/F1.jpg}
        & \includegraphics[valign=c, width=0.157\textwidth]{img/results/F2.jpg}
        & \includegraphics[valign=c, width=0.157\textwidth]{img/results/F3.jpg}
        & \includegraphics[valign=c, width=0.157\textwidth]{img/results/F4.jpg}
        & \includegraphics[valign=c, width=0.157\textwidth]{img/results/F5.jpg}
  \end{tabular}}
  \caption[Redesign Results]{\textbf{Redesign Results}. A side-effect of how our proposed networks are trained is that our method is also capable of applying entirely different designs to characters. While not the focus of our work, it demonstrates the versatility of our method and the robustness of the learned model, even when applied to high-resolution imagery.}
  \label{redesign1}
\end{figure*}

\begin{figure*}
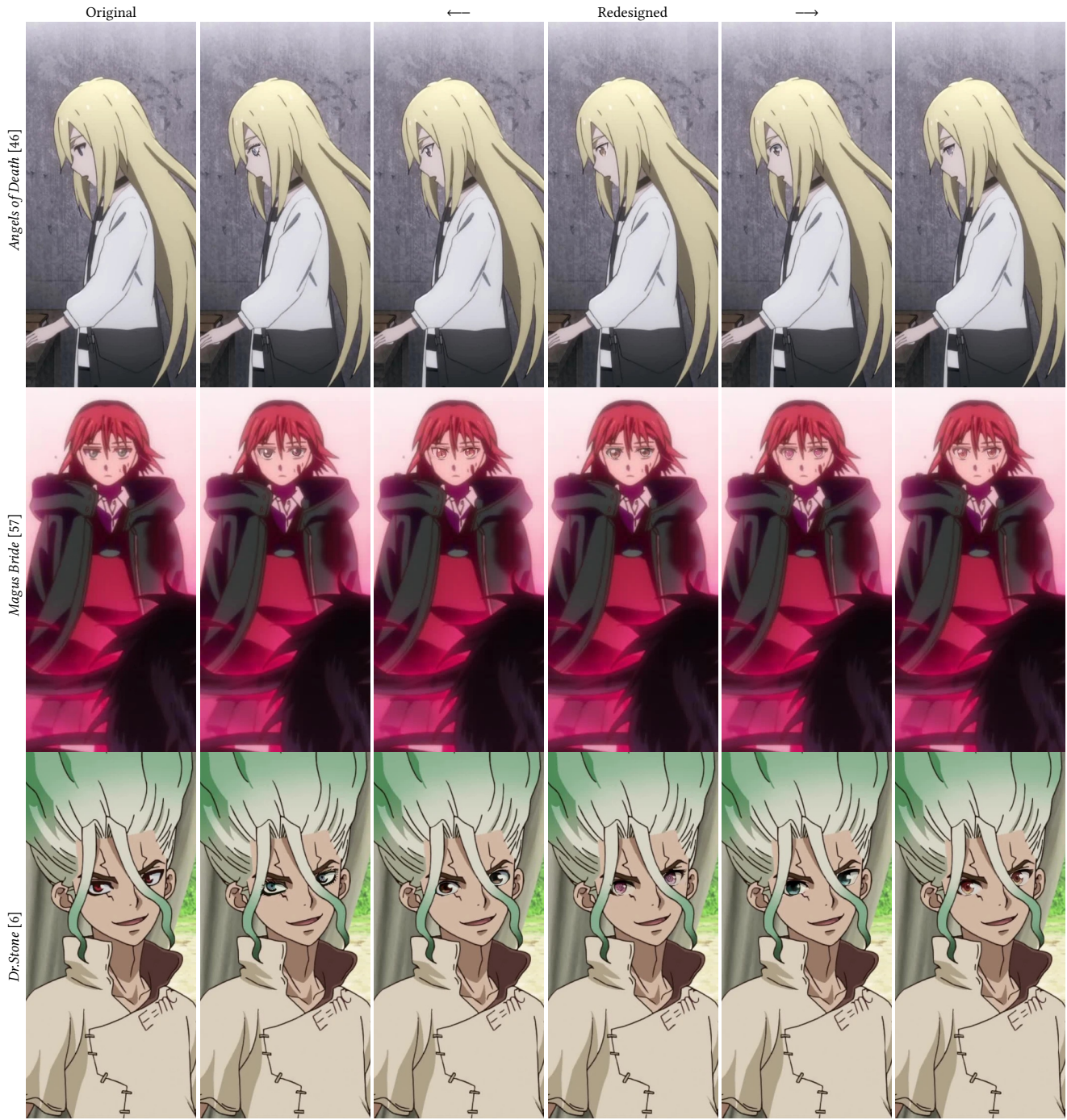

  \setlength{\tabcolsep}{1pt}
  \footnotesize{\begin{tabular}{lcccccc}
   & Original & & $\longleftarrow$ & Redesigned & $\longrightarrow$ & \\
    \raisebox{-.75cm}{\rotatebox{90}{\textit{Angels of Death} \cite{angelsofdeath}}}
        & \includegraphics[valign=c, width=0.157\textwidth]{img/results/R.jpg}
        & \includegraphics[valign=c, width=0.157\textwidth]{img/results/R1.jpg}
        & \includegraphics[valign=c, width=0.157\textwidth]{img/results/R2.jpg}
        & \includegraphics[valign=c, width=0.157\textwidth]{img/results/R3.jpg}
        & \includegraphics[valign=c, width=0.157\textwidth]{img/results/R4.jpg}
        & \includegraphics[valign=c, width=0.157\textwidth]{img/results/R5.jpg} \\
    \raisebox{-.75cm}{\rotatebox{90}{\textit{Magus Bride} \cite{magusbride}}}
        & \includegraphics[valign=c, width=0.157\textwidth]{img/results/I.jpg}
        & \includegraphics[valign=c, width=0.157\textwidth]{img/results/J1.jpg}
        & \includegraphics[valign=c, width=0.157\textwidth]{img/results/J2.jpg}
        & \includegraphics[valign=c, width=0.157\textwidth]{img/results/J3.jpg}
        & \includegraphics[valign=c, width=0.157\textwidth]{img/results/J4.jpg}
        & \includegraphics[valign=c, width=0.157\textwidth]{img/results/J5.jpg} \\
    \raisebox{-.75cm}{\rotatebox{90}{\textit{Dr.Stone} \cite{dr-stone}}}
        & \includegraphics[valign=c, width=0.157\textwidth]{img/results/T.jpg}
        & \includegraphics[valign=c, width=0.157\textwidth]{img/results/T1.jpg}
        & \includegraphics[valign=c, width=0.157\textwidth]{img/results/T2.jpg}
        & \includegraphics[valign=c, width=0.157\textwidth]{img/results/T3.jpg}
        & \includegraphics[valign=c, width=0.157\textwidth]{img/results/T4.jpg}
        & \includegraphics[valign=c, width=0.157\textwidth]{img/results/T5.jpg} \\
  \end{tabular}}
  \caption[Redesign Results]{\textbf{Redesign Results}. Further examples of character redesign which demonstrate that our method is robust enough to handle characters in a variety of challenging scenarios. This includes dealing with oblique camera angles, irregular lighting conditions, head tilts and occlusions caused by hair.}
  \label{redesign2}
\end{figure*}

\begin{figure*}
    \centering
    \begin{subfigure}{0.49\textwidth}
        \centering
        \includegraphics[width=\textwidth]{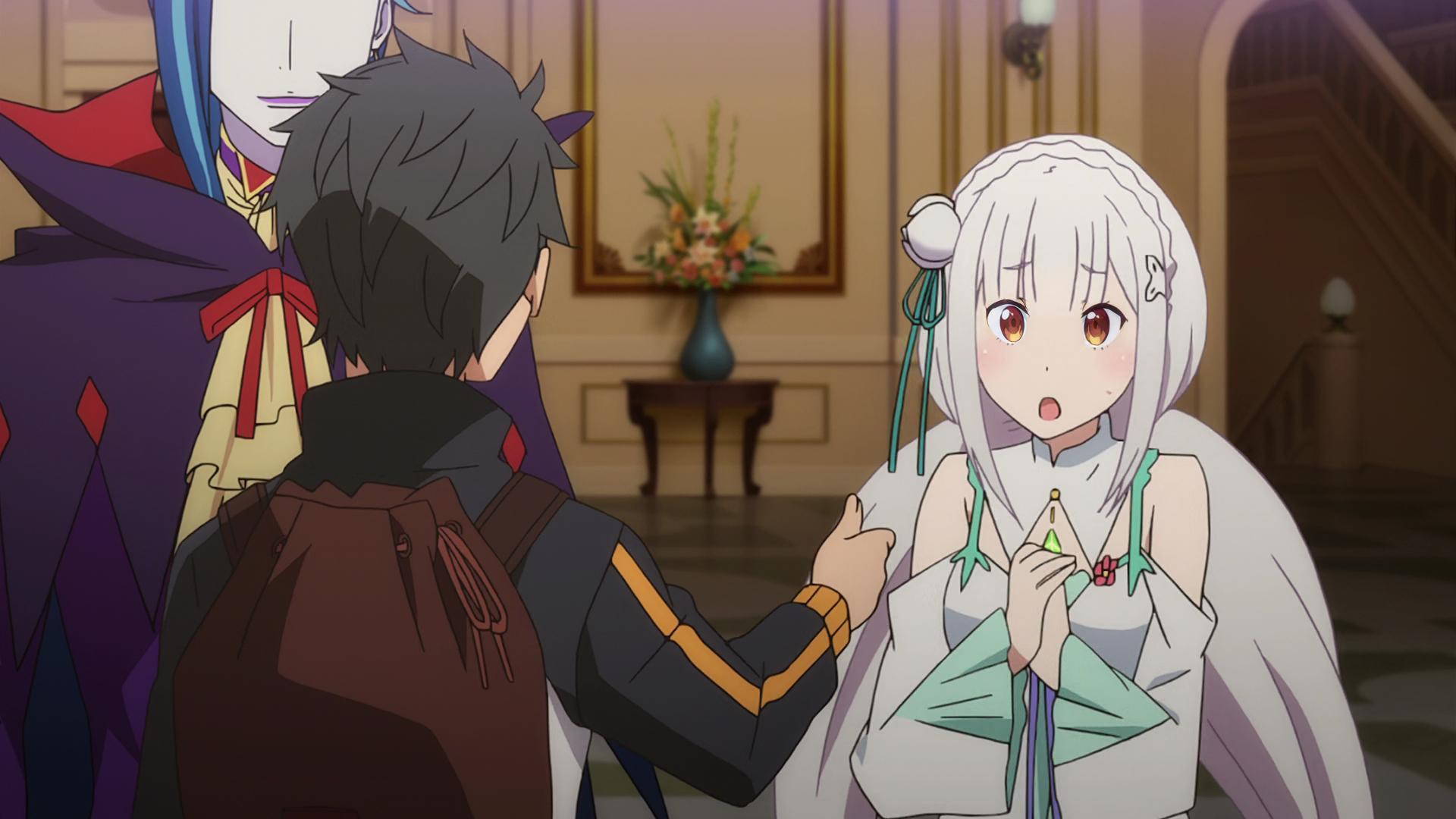}
    \end{subfigure}
    \begin{subfigure}{0.49\textwidth}
        \centering
        \includegraphics[width=\textwidth]{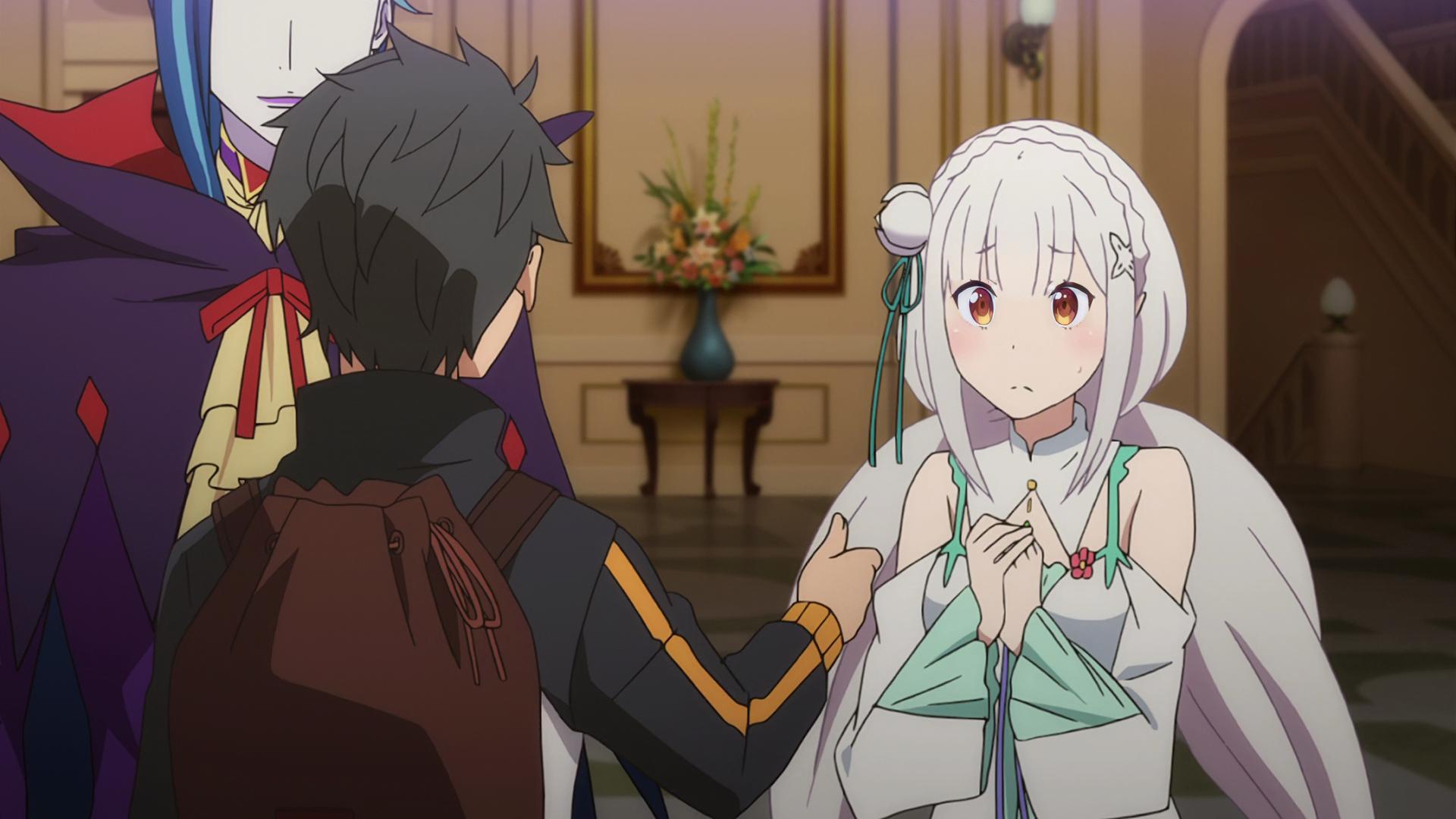}
    \end{subfigure}
    \begin{subfigure}{0.49\textwidth}
        \centering
        \includegraphics[width=\textwidth]{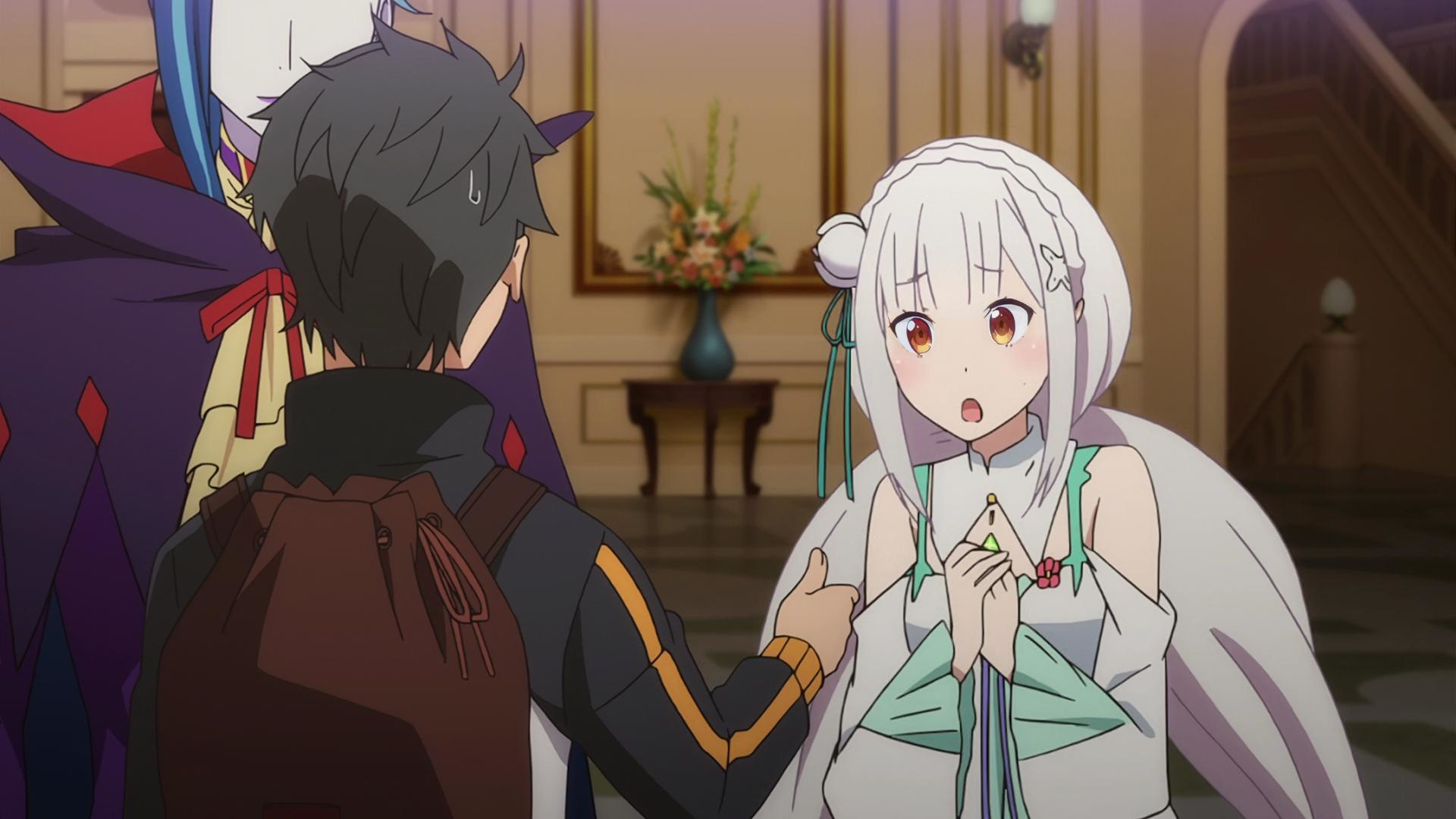}
    \end{subfigure}
    \begin{subfigure}{0.49\textwidth}
        \centering
        \includegraphics[width=\textwidth]{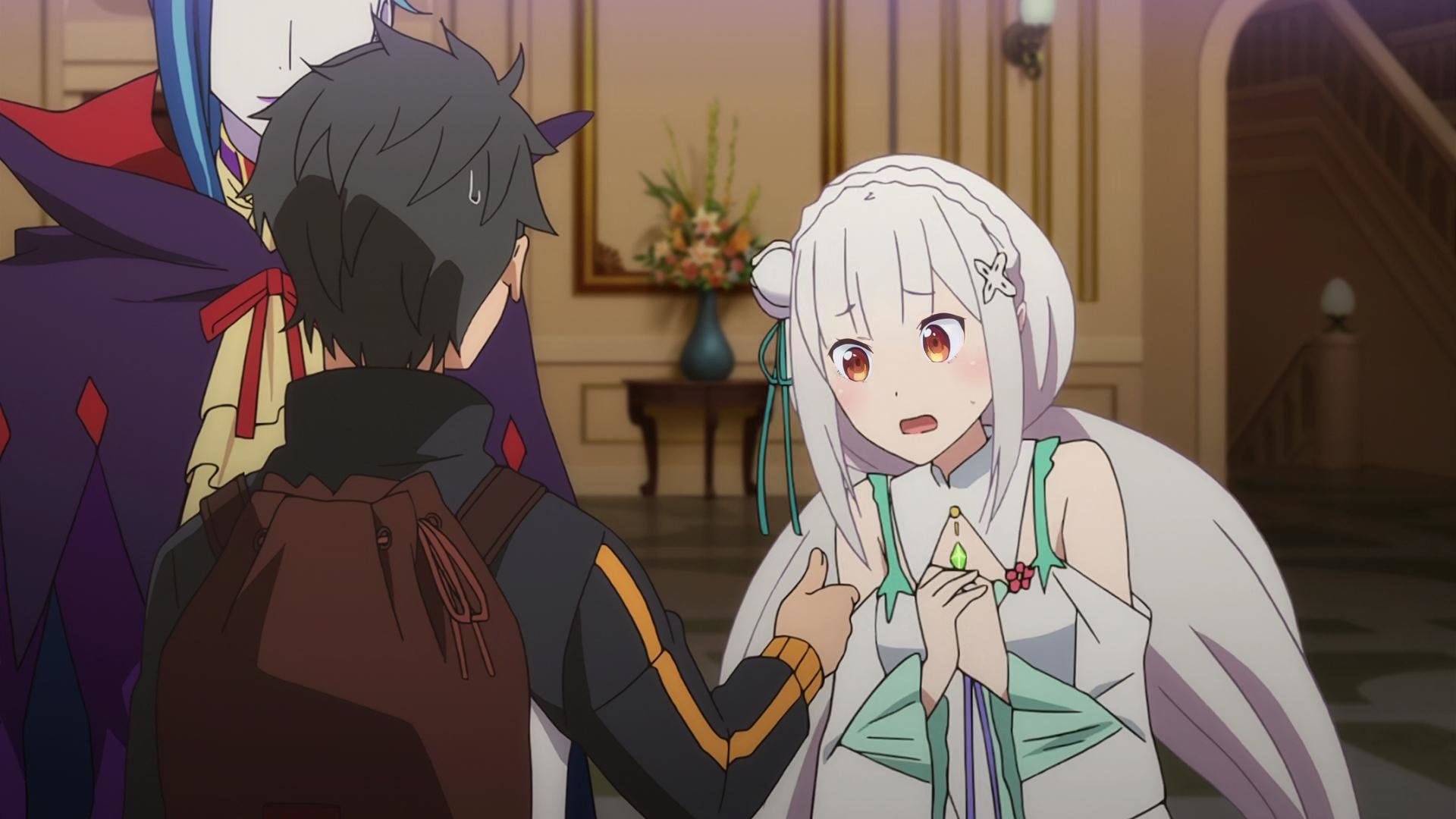}
    \end{subfigure}
    \begin{subfigure}{0.49\textwidth}
        \centering
        \includegraphics[width=\textwidth]{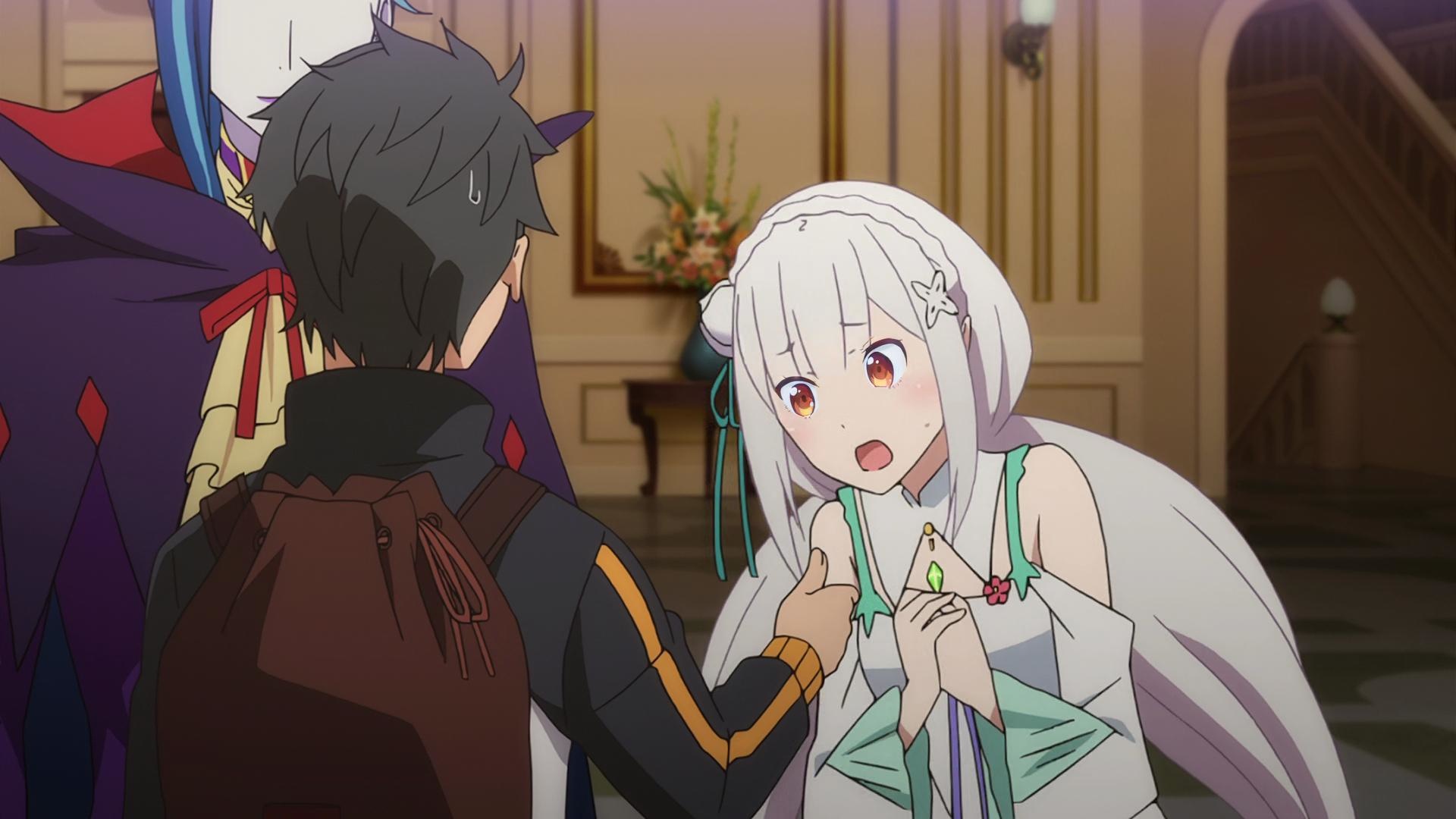}
    \end{subfigure}
    \begin{subfigure}{0.49\textwidth}
        \centering
        \includegraphics[width=\textwidth]{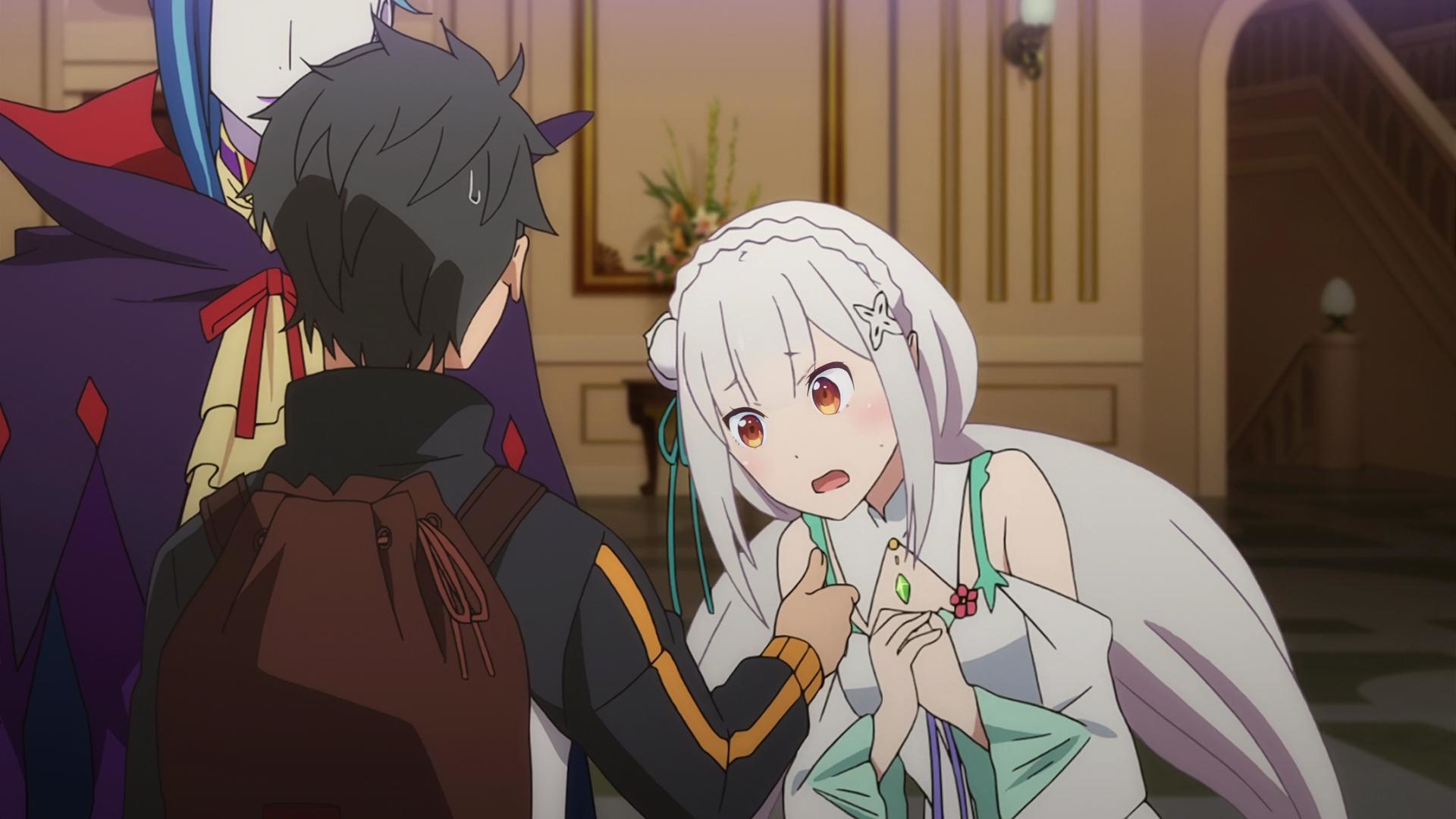}
    \end{subfigure}
    \begin{subfigure}{0.49\textwidth}
        \centering
        \includegraphics[width=\textwidth]{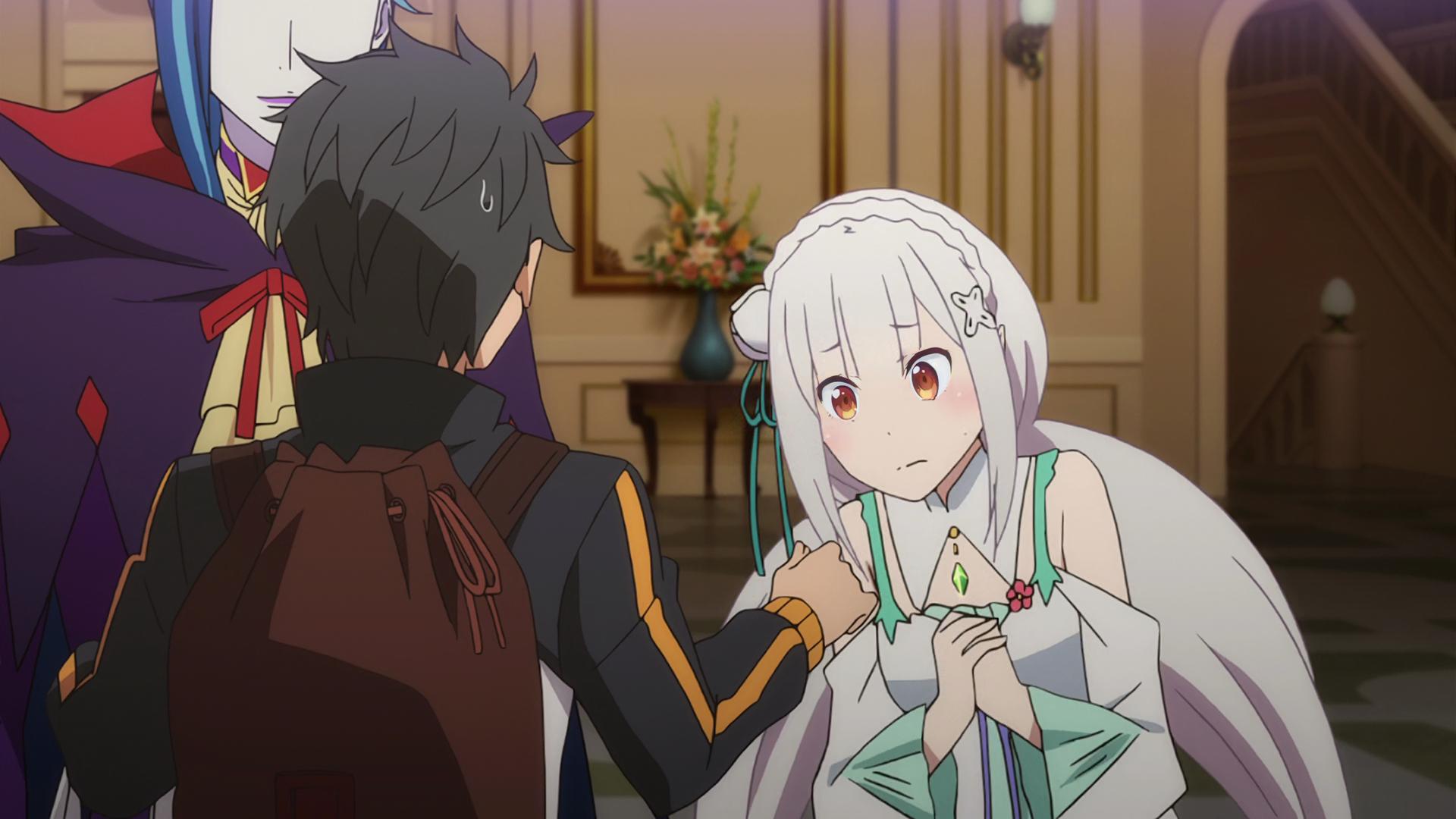}
    \end{subfigure}
    \begin{subfigure}{0.49\textwidth}
        \centering
        \includegraphics[width=\textwidth]{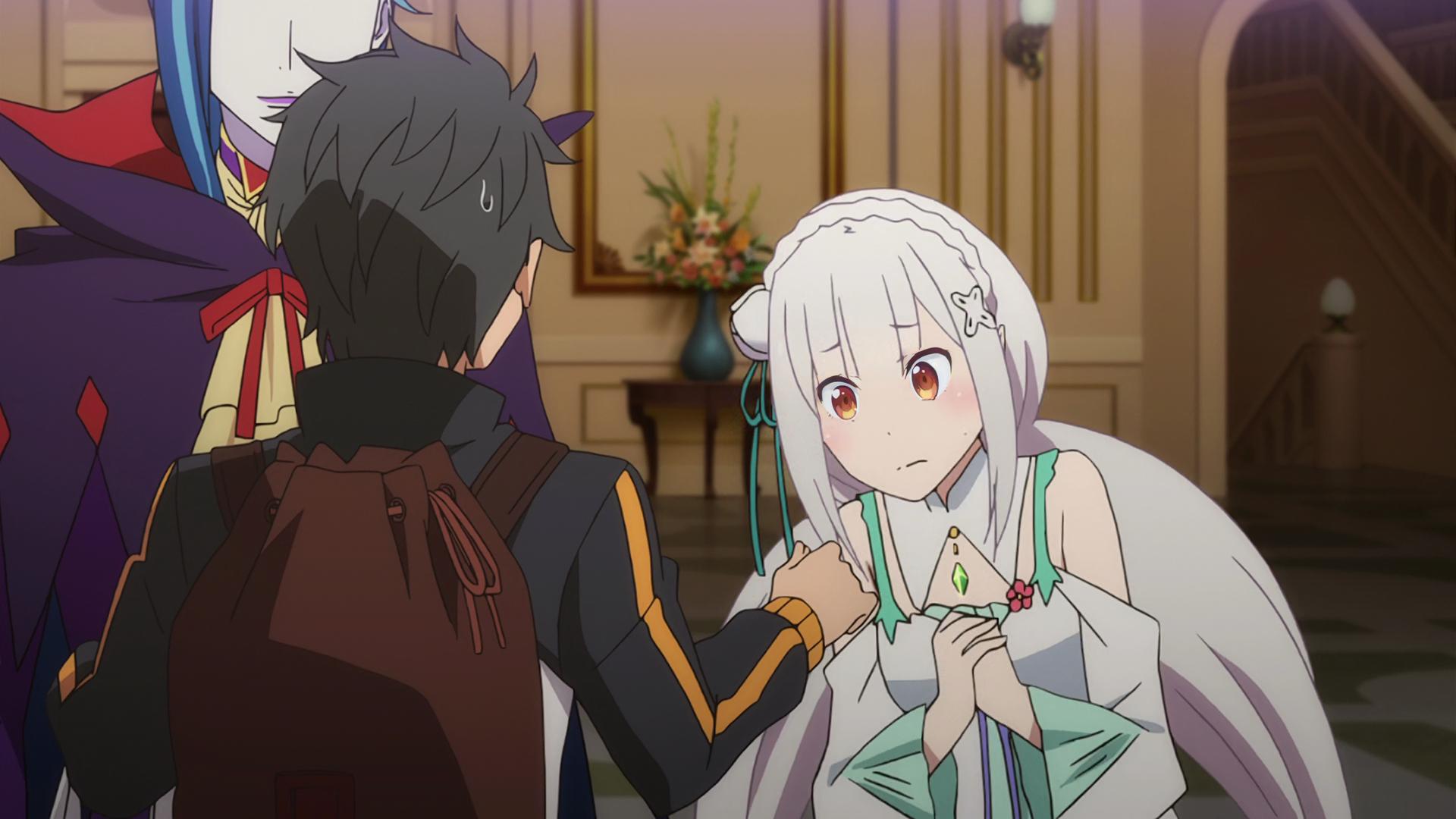}
    \end{subfigure}
    \caption[Temporal Coherence]{\textbf{Temporal Coherence}. Example of redesigning the eyes of a character from \textit{Re:Zero} \cite{re-zero} in a particularly lengthy shot. Output is temporally consistent.}
    \label{temporal-coherence}
\end{figure*}

    }
\end{document}